\documentclass[opendfm]{xlance/xlance}

\usepackage{rotating}
\usepackage{bbding}
\usepackage{makecell}
\usepackage{CJKutf8}
\usepackage{threeparttable}
\usepackage{wrapfig}
\usepackage{needspace}
\usepackage{algorithm}
\usepackage{algpseudocode}
\usepackage{arydshln}

\renewcommand{\arraystretch}{1}

\Crefformat{section}{#2\S~#1#3}
\Crefformat{subsection}{#2\S~#1#3}
\Crefname{appendix}{Appendix}{Appendices}
\Crefname{failurecase}{Case}{Cases}

\protected\def\correspondingauthor#1{#1\textsuperscript{\dag}}

\newcommand{\ours}{\textsc{SynthDCB}}

\newcommand{\Logo}{\raisebox{-0.3\height}{\includegraphics[width=7em]{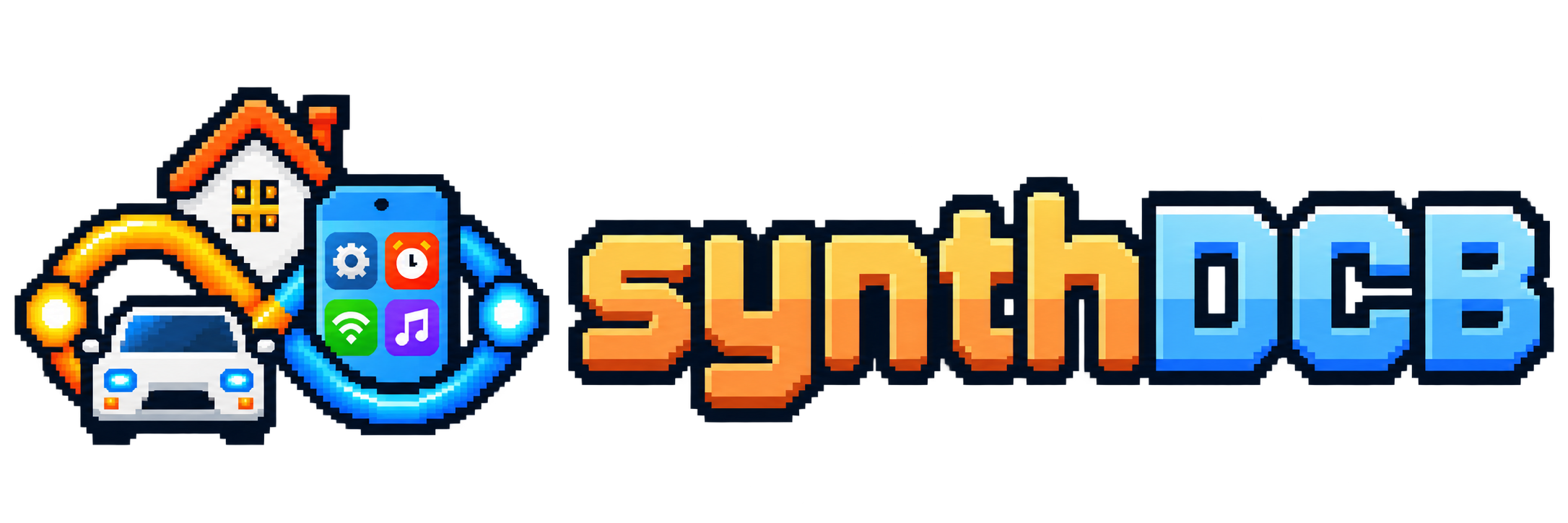}}}

\definecolor{LG}{RGB}{198,239,206} %
\definecolor{MG}{RGB}{149,215,165} %
\definecolor{DG}{RGB}{99,190,123}  %
\definecolor{DDG}{RGB}{49,165,90}  %
\definecolor{LR}{RGB}{255,199,206} %
\definecolor{MR}{RGB}{251,152,156} %
\definecolor{DR}{RGB}{248,105,107} %
\definecolor{DDR}{RGB}{244,58,58} %
\definecolor{MO}{HTML}{D98E04}

\newcommand{\V}{\smash{\Checkmark}}

\algrenewcommand\algorithmicrequire{\textbf{Input:}}
\algrenewcommand\algorithmicensure{\textbf{Output:}}

\newcommand{\para}[1]{%
 \textbf{#1}\quad
}

\lstdefinestyle{prompt}{%
  basicstyle=\normalfont\footnotesize\color{black},
  frame=none,framerule=0pt,framesep=0pt,
  xleftmargin=0pt,xrightmargin=0pt,
  aboveskip=0pt,belowskip=0pt,
  backgroundcolor=\color{boxcolor@bg},
  numbers=none,columns=fullflexible,breaklines=true,breakatwhitespace=false,
  keepspaces=true,showstringspaces=false,
  showspaces=false,showtabs=false,tabsize=4,
}
\makeatletter
\newcommand{\promptsoftwrap}{%
  \lst@lAddTo\lsthk@PostOutput{%
    \ifdim\lst@currlwidth>\linewidth
      \global\lst@currlwidth\linewidth
    \fi}%
}
\makeatother
\newcommand{\inputprompt}[3]{%
  \begingroup
  \setboxscheme{#1}%
  \promptsoftwrap
  \begin{promptbox}[\texttt{\detokenize{#2}}]
    \lstinputlisting[style=prompt]{#3}%
  \end{promptbox}%
  \endgroup
}
\setcitestyle{authoryear,round,citesep={;},aysep={,},yysep={;}}
\fontencoding{T1}\selectfont

\title{\Logo: World Distillation for Scalable Device Control Benchmarking}

\small{
\author[1,3]{Zichen Zhu}
\author[1]{Dingye Liu}
\author[1]{Nan Jiang}
\author[1]{Danyang Zhang}
\author[1]{Hang Zheng}
\author[1]{Shijie Zheng}
\author[1]{Situo Zhang}
\author[1]{Kunyao Lan}
\author[5]{Yingzi Ma}
\author[6]{Sheng Jiang}
\author[1,2,4]{\correspondingauthor{Lu Chen}}
\author[1,4]{\correspondingauthor{Kai Yu}}
}

\small{
\affiliation[1]{X-LANCE Lab, School of Computer Science, Shanghai Jiao Tong University}
\affiliation[2]{Shanghai Innovation Institution}
\affiliation[3]{Nanyang Technological University}
\affiliation[4]{Jiangsu Key Lab of Language Computing}
\affiliation[5]{University of Wisconsin–Madison}
\affiliation[6]{AISpeech Co., Ltd.}
}

\abstract{
Scaling device-control evaluation requires scaling both interactive worlds and the tasks agents must solve. We introduce \textbf{\ours}, a benchmark for environment awareness and device control built through \textbf{world distillation} (WD). World distillation transforms knowledge of device behaviors, constraints, and environmental effects into modular executable components. A unified capability representation connects tool discovery, world execution, and task synthesis. \ours~instantiates Home, Car, and Phone with 114 device classes, 1,166 functions, and 95 events, supporting five task families with 32 subtypes. Explicit states and tool-call traces support task validation and agent evaluation. We evaluate 22 models on 2,560 test tasks; the best achieves 88.8\% mean Pass@1.
Construction experiments demonstrate token efficiency and expansion of task coverage. At the richest knowledge level, WD fully covers 22/23 properties, 19/29 functionalities, and 30/39 event requirements specified in a real RobotVacuum protocol. Furthermore, the 1,000 sampled tasks semantically cover 81.6\% of the 854 reference instructions in the real-world corpus. \ours~connects an expanding space of real-to-sim executable capabilities to verifiable agent evaluation.
}

\metadatainfo{
    \small\makebox[\linewidth][c]{
        \href{https://x-lance.github.io/SynthDCB/}{\web~\textbf{Website}}
        \qquad
        \href{https://huggingface.co/datasets/JamesZhutheThird/SynthDCB}{\huggingface~\textbf{Dataset}}
        \qquad
        \href{https://github.com/X-LANCE/SynthDCB}{\github~\textbf{Code}}
    }
}

\begin{document}
\publishdate{2026.10.14}
\revisedate{2026.11.19}
\maketitle

\raggedbottom
\addtocontents{toc}{\protect\setcounter{tocdepth}{-1}}

\section{Introduction}
\label{sec:introduction}

Tool-using agents are advancing rapidly and integrated into voice assistants in home, car, or mobile phones, accompanied by a growing collection of benchmarks and executable environments for evaluating their capabilities \citep{taubench,toolsandbox,simuhome,smhbench}. At the same time, device manufacturers and system developers continually introduce new devices and expand their functionality. When simulation environments do not keep pace, the capabilities covered during agent training and evaluation can lag behind those required in deployment. This creates a need for benchmarks that can be rapidly extended as new device capabilities emerge, both to evaluate agents against evolving requirements and to support iterative agent training.

Meeting this need requires addressing two connected challenges. On the construction side, adding a device capability requires consistent tool interfaces, behavior implementations, world configurations, task requirements, and success criteria; generating more instructions over an unchanged environment does not expand what the benchmark can execute. On the evaluation side, agents must understand and operate the environment presented to them, rather than merely invoke controls in a fixed configuration. This requires discovering available devices and operations, interpreting the current world state, and determining which parts of a user request can be satisfied. An extensible benchmark should therefore support both the introduction of new executable capabilities and the assessment of environment understanding and device control across different worlds.

Consider the request: ``Cool the bedroom to $22\,^{\circ}\mathrm{C}$ before 22:00, and turn on the bedside lamp at 20\% brightness.'' Changing an air conditioner's setpoint is not the same as bringing the room to the requested temperature: the latter depends on environmental dynamics, and meeting the deadline also depends on when actions are taken. If the cooling requirement is infeasible in the current world, the lighting requirement may still be achievable. Evaluating these distinctions requires explicit action effects, temporal evolution, and success conditions for individual requirements. We therefore ask: \emph{How can we rapidly extend executable device environments and verifiable tasks to evaluate agents' ability to understand and control unfamiliar worlds?}

We introduce the \emph{world distillation} (WD) pipeline that transforms behavioral knowledge into reusable executable components for constructing worlds and their tasks. Typed capability specifications guide component implementation and validation, define discoverable tool interfaces, and provide the operations used by task builders. World composition assembles these components into domain-specific layouts or module structures. Execution-grounded task synthesis derives requirements from available capabilities and checks candidate tasks through reference-execution replay. Once validated, a component can be reused across world configurations and task instances without reimplementing its behavior again. This supports on-demand device expansion, while configurable worlds enable evaluation of how agents explore and use the capabilities of the devices instantiated for each task.

\begin{figure}[t]
    \centering
    \includegraphics[width=\linewidth]{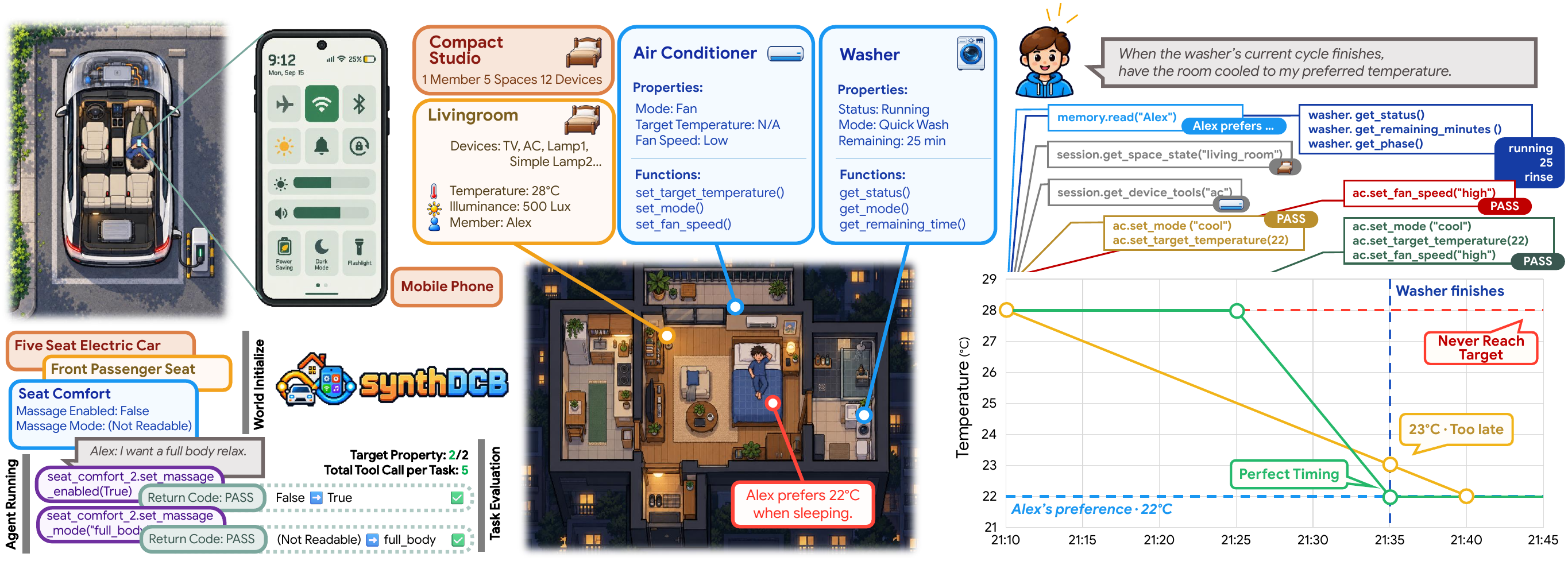}
    \caption{Overview of \ours.
    \ours{} builds configurable executable worlds from reusable device capabilities across Home, Car, and Phone.
    Agents interact with these worlds through device tools and are evaluated on tasks involving environment states, user preferences, and temporal dynamics.
    The example shows coordinating an air conditioner with a washer cycle to reach Alex's preferred temperature at the required time.}
    \label{fig:overview}
\end{figure}

We instantiate this approach as \ours, covering Home, Car, and Phone with 114 device classes, 1,166 functions, and 95 events. The benchmark supports five task families with 32 subtypes, combining state queries, device control, environmental goals, and coordinated execution. Evaluation of 22 models on 2,560 test tasks yields a best mean Pass@1 of 88.8\%. Model rankings vary across domains. Larger models perform well on direct device control but struggle primarily with interpreting complex user intent and coordinating subtasks to fulfill the complete request. Smaller models face more basic challenges in following interaction protocols and selecting appropriate tools to complete individual atomic tasks. %

Construction experiments with Codex using the same GPT-5.6 Sol demonstrate the scalability and efficiency of world distillation. Building all three domains consumes 19.16M tokens at an estimated cost of \$120.09 before cache discounts within a couple of hours, automating specification, implementation, testing, and repair that could otherwise require over a month of manual work. In the RobotVacuum comparison, WD achieves broader complete capability coverage than Codex low at comparable token budgets under richer knowledge inputs, and uses 51.5-86.7\% fewer tokens per completely covered requirement than Codex xhigh across three knowledge levels. With full-protocol input, 1,000 generated RobotVacuum tasks cover 81.6\% of 854 reference instructions and 69.8\% of 172 categorized intents. These results connect efficient capability construction to broader task coverage, demonstrating the scalability of world distillation. %

We summarize our contributions as follows:
\begin{itemize}[leftmargin=1.2em, itemsep=0.5pt, topsep=0.5pt]
    \item \textbf{An extensible environment and world distillation pipeline.}
    We design a stateful simulator with explicit device behaviors, shared environmental effects, and temporal dynamics, together with a pipeline that transforms device knowledge into executable components and validated tasks.

    \item \textbf{A benchmark and evaluation of 22 models.}
    We introduce \ours~across Home, Car, and Phone, with five task families covering 32 subtypes. Evaluation on 2,560 test tasks and detailed failure analysis expose distinct gaps in environment awareness and reliable device control.

    \item \textbf{An analysis of construction scalability and efficiency.}
    We quantify domain construction costs, stage execution, and capability coverage under different knowledge inputs. Comparisons with Codex demonstrate efficient expansion of executable behaviors, providing a foundation for broader evaluation and capability transfer.
\end{itemize}

\section{Related Work}
\label{sec:related}

\para{Tool-use benchmarking.}
Tool-use research evaluates and trains agents to complete tasks through external APIs. Call-oriented benchmarks assess tool selection and argument construction~\citep{api-bank,bfcl}, while synthesis methods generate tool-use supervision~\citep{toollm,apigenmt,toolace}. Stateful benchmarks evaluate execution consequences through policy-constrained interactions~\citep{taubench,tau2bench}, state dependencies and trajectory milestones~\citep{toolsandbox}, or state checks~\citep{appworld}. Device control further requires modeling how multiple operations affect shared environment states over time.

\para{Device-control agents and environments.}
Device-control benchmarks examine instruction understanding, environment discovery, and coordinated execution. Existing approaches emphasize command validity, grounded control, and personalized assistance~\citep{homebench,sage,personalhomebench}. SimuHome and SMH-Bench introduce home simulators for temporal workflows and contextual reasoning~\citep{simuhome,smhbench}, while VehicleWorld and DevicesWorld extend evaluation to in-car interaction and heterogeneous device workflows~\citep{vehicleworld,devicesworld}. As device capabilities evolve, these environments must expand their executable behaviors together with the tasks that exercise them. \Cref{tab:benchmark_comparison} situates \ours~among related benchmarks. \ours~combines coverage across 3 domains with extensible device behaviors and verifiable tasks, allowing new capabilities to enter evaluation through a shared world distillation pipeline.

\para{Scalable executable environments and tasks.}
Environment and task synthesis expands agent training and evaluation while aligning tasks with executable outcomes. Environment-centered approaches generate environments alongside tasks or verifiers~\citep{agentgen,envscaler,agentworld,ebench}; other methods organize function dependencies or combine procedural homes with executable goals~\citep{funcbenchgen,homeflow}. Scaling device worlds additionally requires reusable behaviors that interact through shared state and simulated time. \ours~addresses this through world distillation: device knowledge becomes executable components, world composition instantiates their interactions, and task synthesis derives verifiable requirements. Each new component can thus expand both the environment and its task inventory.

\begin{table}[H]
\centering
\begin{threeparttable}

\setlength{\tabcolsep}{3.28pt}
\renewcommand{\arraystretch}{1}
\scriptsize

\begin{tabular}{@{}llccccccccccc@{}}
\toprule
\textbf{Benchmark} & \textbf{Domains} & \textbf{\# Tasks} & \textbf{\# Types} & \textbf{\# Dev.} & \textbf{\# Func.} & \textbf{\# Inst.} & \textbf{E-Dyn.} & \textbf{Time} & \textbf{Mem.} & \textbf{Infeas.} & \textbf{E-Exp.} & \textbf{T-Exp.} \\
\midrule
ToolSandbox & Personal Assistant & 1,032 & 7 &  & 34 &  &  &  &  & \V &  & Manual \\
$\tau$-Bench & Retail, Airline & 165 &  &  & 28 &  &  &  &  &  & Manual & Manual \\
$\tau^2$-Bench & Telecom & 114 & 15 &  & 43 &  &  &  &  &  & Semi & Semi \\
DevicesWorld$^\dagger$ & Home, Phone, Desktop & 6,140 & 9 & 11 &  &  &  & \V &  & \V &  & Semi \\
VehicleWorld & Car & 1,291 & 4 & 30 & 250 &  &  &  &  &  &  & Semi \\
SimuHome & Home & 600 & 12 & 17 & 17 &  & \V & \V &  & \V &  & Semi \\
SMH-Bench$^\dagger$ & Home & 1,100 & 22 & 17 &  & 3 &  & \V & \V &  &  & Semi \\
\midrule
\textbf{\ours} & \textbf{Home, Car, Phone} & \textbf{2,560 + 314} & \textbf{32} & \textbf{114} & \textbf{1,166} & \textbf{8} & \V & \V & \V & \V & \textbf{Auto} & \textbf{Auto} \\
\bottomrule
\end{tabular}

\begin{tablenotes}[flushleft]
\scriptsize
\item[] Types: fine-grained task types; Dev.: devices; Func.: functions; Inst.: instruction styles; E-Dyn.: environmental dynamics; Mem.: memory; Infeas.: infeasible tasks; E-Exp.: environment expansion; T-Exp.: task expansion. Auto: fully automated, with no human effort required; Semi: automated with human intervention; Manual: fully human-authored.
\item[$\dagger$] As of this revision (November 19, 2026), we could not locate a public release of the full task set for either benchmark. Statistics and feature annotations are drawn from the respective papers.
\end{tablenotes}

\end{threeparttable}
\caption{Comparison of daily agent benchmarks. }
\label{tab:benchmark_comparison}
\end{table}

\section{Method}
\label{sec:method}

\ours~connects executable capabilities to verifiable tasks through a shared simulator and two generation stages: \emph{world distillation} builds reusable device models and composes world instances; \emph{task construction} derives requests, private targets, and reference executions from their available capabilities. This design lets a new device model support many worlds and user requests.

\begin{figure}[t]
 \centering
 \includegraphics[width=\linewidth]{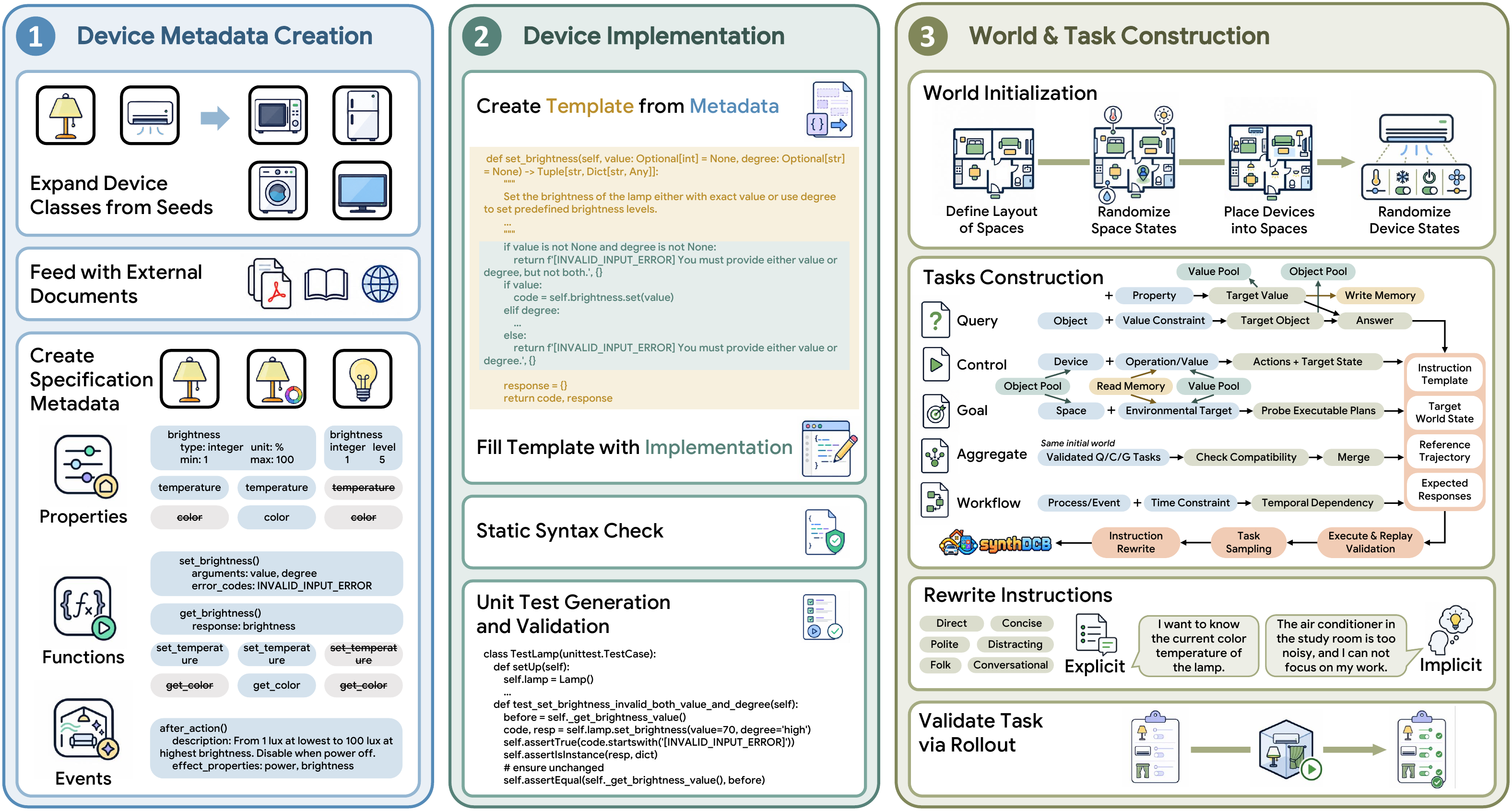}
 \caption{The system pipeline of world distillation and task construction. World distillation connects behavioral knowledge to executable capabilities, composed worlds, and validated tasks. A shared capability specification links discovery, execution, and task construction. Component checks drive bounded repair; reference replay validates task construction.}
 \label{fig:pipeline}
\end{figure}

\subsection{Simulator Design}
\label{sec:representation}

\para{Devices, spaces, and worlds.}
A \emph{device} class $\mathcal D$ combines a specification $C^D=(\mathcal P^D,\mathcal T^D,\mathcal E^D)$ of properties, tool-call functions, and events defining environmental effects with a Python implementation $\mathcal F^D$. Instantiating device $d$ assigns its property vector $\mathbf p^d_{t_0}=(p^d_{k,t_0})_{k\in\mathcal P^D}$, giving the instance state $d_{t_0}=(\mathcal D,\mathbf p^d_{t_0})$.

A physical or logical \emph{space} class $\mathcal V$ specifies shared properties $\mathcal P^V$ and metadata. Instantiating space $v$ assigns its property vector $\mathbf p^v_{t_0}=(p^v_{k,t_0})_{k\in\mathcal P^V}$ and device instances indexed by $\mathcal I_v^D$, giving $v_{t_0}=(\mathcal V,\mathbf p^v_{t_0},\{d_{t_0}\}_{d\in\mathcal I_v^D})$. The states in the same space at the same moment are considered consistent and change synchronously.

A \emph{world} specification $\mathcal W$ defines the layout, connections, and device placement rules. Instantiation binds spaces indexed by $\mathcal I_W^V$ to this layout, resolves relations and member locations, and initializes runtime state $r_{t_0}$ and clock $c_{t_0}$. The initial world is $W_{t_0}=(\mathcal W,\{v_{t_0}\}_{v\in\mathcal I_W^V},r_{t_0},c_{t_0})$. Task memory, when provided, is also bound to its runtime state.

\para{Runtime simulation.}
For interaction round at moment $t$, the agent submits an action sequence $A_t$ whose tool calls are issued together and executed in the supplied order. Each query reads the state at its execution point; successful control calls immediately update device state, so later queries can observe earlier changes within the same round. The collected tool responses form the observations for the next round, as agent is required to explore the environment by itself. Let $E_t$ denote the collection of currently active environmental effects, each specifying its source, target property, and update rule. The round-level transition of runtime state is
\begin{equation}
 A_t=(a_{1},\cdots,a_{m_t}),
 \qquad (W_{t+1},E_{t+1},O_{t+1})=\text{Simulator}(W_t,E_t,A_t),
 \label{eq:world}
\end{equation}
where observation $O_{t+1}$ contains responses of $m_t$ tool calls. The effect collection is refreshed within the round as device states change and time advances. After-action events update immediate contributions, such as illumination; at every one-minute tick, gradual effects such as cooling are recomputed from current device and space states. The runtime combines the applicable effects to update shared environmental properties and advances ongoing device processes. Thus, $E_{t+1}$ captures the active effects after the round. Time advances according to configured action durations, explicit \texttt{sleep\_time} calls, and the round-end time policy. %

\subsection{World Distillation}
\label{sec:distillation}

\para{Device lists and specification metadata.}
World distillation converts behavioral knowledge into reusable executable components under a generation and validation harness. We iteratively expand a domain's device list with descriptions of its functionality and environmental effects, registering base and inherited classes. %
For each class, we generate specification metadata defining the properties, functions, constraints, and events above. Schema and capability-consistency checks, followed by semantic review, provide feedback for revision. %

\para{Code template, implementation, and tests.}
Deterministic templates instantiate class structure, property wrappers, and common interfaces using metadata; an LLM completes the remaining function and event bodies and generates unit tests. Static checks, specification--implementation checks, effect validation, and isolated test execution return failures for repair. A dependency-based job planner runs independent jobs in parallel, blocks dependent jobs on failure, and supports reruns of prerequisite stages. We provide scalability and efficiency analysis of world distillation in \Cref{sec:scaling}.

Inherited classes reuse their parent's specification and implementation through \texttt{update} and \texttt{disabled} declarations. For example, a cooling-only air conditioner removes heating modes and auxiliary heating and overrides effects related to heating. Stage-specific retrieval supplies domain documents, i.e., user manuals or product list of shopping app, and expert specification, code, and test examples; accepted artifacts are also included in the reference library to support later generation.

\para{World composition.}
\label{sec:composition}
The validated device library $\mathcal M=\{(C_i^D,\mathcal F_i^D)\}_i$ is composed with space classes $\{\mathcal V_j\}_j$ under the world specification $\mathcal W$:
\begin{equation}
 W_{t_0}=\operatorname{Compose}(\mathcal M,\{\mathcal V_j\}_j,\mathcal W;z),
 \label{eq:composition}
\end{equation}
where $i$ and $j$ index device and space classes, and $z$ fixes sampling choices. The layout and placement rules in $\mathcal W$ determine space instances, device counts, and assignments. Initialization assigns legal property vectors $\mathbf p^d_{t_0}$ and $\mathbf p^v_{t_0}$ while preserving coupled-state constraints, resolves connections and member locations, and initializes runtime state $r_{t_0}$ and clock $c_{t_0}$. The saved configuration fixes $W_{t_0}$ for task construction and replay.

\subsection{Task Construction}
\label{sec:synthesis}

\para{Capability-grounded sampling.}
Each instantiated world supplies a shared capability inventory of queryable properties, device operations, processes, and candidate environmental goals. Builders efficiently reuse this inventory and cached execution previews to produce batches of tasks from the same world or valid initial-state variants. \Cref{tab:task_types} summarizes the five task families. Target devices or spaces can be selected directly, through environment exploration, or through filtering and comparison queries from object pool; target values can come from legal samples, relative changes, queried values from value pool, or memory. Queries that determine later targets or values become explicit prerequisite atom tasks. Goal builders probe executable space change plans, while Aggregate builders combine validated components from the same initial state after checking read/write conflicts and target compatibility.

\begin{table}[H]
\centering
\scriptsize
\setlength{\tabcolsep}{5pt}
\renewcommand{\arraystretch}{1}

\begin{tabular}{@{}l >{\raggedright\arraybackslash}p{0.20\linewidth} >{\raggedright\arraybackslash}p{0.68\linewidth}@{}}
\toprule
\textbf{Family} & \makecell[t{>{\raggedright\arraybackslash}p{\linewidth}}]{\textbf{Construction target}} & \makecell[t{>{\raggedright\arraybackslash}p{\linewidth}}]{\textbf{Subtypes}} \\
\midrule
Query & \makecell[t{>{\raggedright\arraybackslash}p{\linewidth}}]{Information or entity selection} & \makecell[t{>{\raggedright\arraybackslash}p{\linewidth}}]{property.device (Q1), property.space (Q2), space.person\_location (Q3), space.occupants (Q4), target.quantity (Q5), target.enum (Q6), target.comparison.top1 (Q7), target.comparison.topk (Q8)} \\
\cmidrule{1-3}
Control & \makecell[t{>{\raggedright\arraybackslash}p{\linewidth}}]{Device state or operation} & \makecell[t{>{\raggedright\arraybackslash}p{\linewidth}}]{single (C1), multi\_property (C2), multi\_target (C3), complex (C4)} \\
\cmidrule{1-3}
Goal & \makecell[t{>{\raggedright\arraybackslash}p{\linewidth}}]{Environmental state} & \makecell[t{>{\raggedright\arraybackslash}p{\linewidth}}]{single (G1), multi\_property (G2), multi\_target (G3), complex (G4)} \\
\cmidrule{1-3}
Aggregate & \makecell[t{>{\raggedright\arraybackslash}p{\linewidth}}]{Compatible combined requests} & \makecell[t{>{\raggedright\arraybackslash}p{\linewidth}}]{query (A1), control (A2), goal (A3), control\_query (A4), control\_goal (A5), query\_goal (A6), control\_query\_goal (A7)} \\
\cmidrule{1-3}
Workflow & \makecell[t{>{\raggedright\arraybackslash}p{\linewidth}}]{Process or event timing} & \makecell[t{>{\raggedright\arraybackslash}p{\linewidth}}]{single (W1), condition\_follow (W2), condition\_delay (W3), condition\_window (W4), synchronize (W5), at (W6), before (W7), environment\_deadline (W8), environment\_before\_process (W9)} \\
\bottomrule
\end{tabular}
\caption{Five families and 32 subtypes; domain coverage depends on instantiated capabilities.}
\label{tab:task_types}
\end{table}

\para{Memory and feasibility.}
Memory adds user-specific or global preferences: property queries can create or update records, and supported Control and Goal tasks retrieve their target values from saved preferences. LLM-based rewriting makes memory content more natural while introducing semantically similar distractor memories. Feasibility variants introduce missing devices or spaces, unsupported capabilities, invalid values, and goals outside established reachable bounds. The task builder can also sample aggregate tasks combining both executable and infeasible atom tasks; a failed prerequisite blocks its dependent atom tasks while independent atom tasks remain active and should be completed by agent.

\para{Joint construction and validation.}
A builder produces a structured request $q$ with atomic requirements and dependencies, a private target $g$, a template instruction $u_0$, and a reference plan $\pi^{\mathrm{ref}}$:
\begin{equation}
 (q,u_0,g,\pi^{\mathrm{ref}})=\operatorname{Build}_{\phi}(W_0;\zeta),
 \qquad b=(W_0,u_0,g,e),
 \label{eq:task-construction}
\end{equation}
where $\phi$ is the task family and $\zeta$ fixes sampling choices. Executing and replaying the reference plan checks the task structure, instruction anchors, and expected outcomes, including reports for failed requests. The accepted bundle stores this construction evidence in $e$. %
Interaction protocols and evaluation criteria are described in \Cref{sec:benchmark}.

\para{Instruction rewrite.}
The instruction of each atom task is sampled through the combination of templates to keep diversity. We further use an optional LLM to rewrite template instructions under seven style policies (\Cref{tab:instruction_styles}). For basic styles, LLM-rewrite variants are more natural and smooth while preserving necessary information. Distracting variants extend conversational phrasing with irrelevant asides. Implicit variants reduce specificity to test contextual interpretation. The policies thus vary wording, information explicitness, and distracting content.

\begin{table}[H]
\centering
\scriptsize
\setlength{\tabcolsep}{5pt}
\renewcommand{\arraystretch}{1}

\begin{tabular}{@{}l p{0.28\linewidth} p{0.535\linewidth}@{}}
\toprule
\textbf{Style} & \makecell[t{>{\raggedright\arraybackslash}p{\linewidth}}]{\textbf{Expression policy}} & \makecell[t{>{\raggedright\arraybackslash}p{\linewidth}}]{\textbf{Example}} \\
\midrule
Template (S1) & \makecell[t{>{\raggedright\arraybackslash}p{\linewidth}}]{Rule-generated instruction} & \makecell[t{>{\raggedright\arraybackslash}p{\linewidth}}]{\textit{For Bedroom Lamp 1, set brightness to 20\%.}} \\
\cmidrule{1-3}
Direct (S2) & \makecell[t{>{\raggedright\arraybackslash}p{\linewidth}}]{Explicit model terminology} & \makecell[t{>{\raggedright\arraybackslash}p{\linewidth}}]{\textit{Set the bedroom lamp's brightness to 20\%.}} \\
Concise (S3) & \makecell[t{>{\raggedright\arraybackslash}p{\linewidth}}]{Compact, explicit wording} & \makecell[t{>{\raggedright\arraybackslash}p{\linewidth}}]{\textit{Bedroom lamp: brightness 20\%.}} \\
Polite (S4) & \makecell[t{>{\raggedright\arraybackslash}p{\linewidth}}]{Courtesy, explicit wording} & \makecell[t{>{\raggedright\arraybackslash}p{\linewidth}}]{\textit{Please set the brightness of bedroom lamp to 20\%.}} \\
Conversational (S5) & \makecell[t{>{\raggedright\arraybackslash}p{\linewidth}}]{Natural everyday phrasing} & \makecell[t{>{\raggedright\arraybackslash}p{\linewidth}}]{\textit{Could you dim the bedroom lamp to 20\%?}} \\
Folk (S6) & \makecell[t{>{\raggedright\arraybackslash}p{\linewidth}}]{Colloquial phrasing} & \makecell[t{>{\raggedright\arraybackslash}p{\linewidth}}]{\textit{Turn that bedroom light down to 20\%, will ya?}} \\
Distracting (S7) & \makecell[t{>{\raggedright\arraybackslash}p{\linewidth}}]{Conversation with irrelevant asides} & \makecell[t{>{\raggedright\arraybackslash}p{\linewidth}}]{\textit{That bedroom lamp looks old-fashioned. Could you dim it to 20\%?}} \\
\cmidrule{1-3}
Implicit (S8) & \makecell[t{>{\raggedright\arraybackslash}p{\linewidth}}]{Broad intent, omitted details} & \makecell[t{>{\raggedright\arraybackslash}p{\linewidth}}]{\textit{Make the bedroom light a little dimmer.}} \\
\bottomrule
\end{tabular}
\caption{Examples of instruction style variants.}
\label{tab:instruction_styles}
\end{table}

\section{The \ours~Benchmark}
\label{sec:benchmark}

\subsection{Dataset and Splits}
\label{sec:coverage}

\ours~covers Home, Car, and Phone with 114 device classes and 1,166 tools. Its current evaluation splits contain \textbf{314 validation tasks} (150 Home, 105 Car, and 59 Phone) and \textbf{2,560 test tasks} (1,281 Home, 768 Car, and 511 Phone). Both splits are sampled disjointly from a shared pool of 30,309 tasks constructed from 130 worlds. We use the validation set to compare 8 instruction styles across 9 models, and evaluate 22 models on the test set with conversational style to reflect most frequent everyday requests. The test set includes 2,289 feasible, 170 infeasible, and 101 partial tasks, with memory requirements in 200 tasks. \Cref{app:benchmark} reports detailed statistics of \ours.%

\subsection{Agent and Interaction Protocol}
\label{sec:interaction}

All evaluated models\footnote{Temperature is 0 and top-$p$ is 1. Response limits are 1,024 tokens for local-deployed models, and 16,384 for API served models, while all other parameters remain at their default values.} use the same ReAct-style agent~\citep{yao2023react} and interact with \ours~through iterative observations and actions.
Agents use system-level tools to discover environmental rules, available devices, and function schemas, then use this information to select target devices and construct function calls in subsequent turns.

Query answers use \texttt{session.submit\_answer}, one per question in instruction order with required JSON types.
Each infeasible or blocked atom task needs a \texttt{session.submit\_report} with status, reason code, and explanation; partial tasks also require completing all independently executable atom tasks.

Simulated time advances through configured action/turn costs (inactivated by default) and \texttt{session.sleep\_time}. Agent interaction ends when the agent calls \texttt{session.finish} or reaches the 20-turn limit. For tasks containing Goal or Workflow atoms, background simulation continues without further agent actions to the task's fixed evaluation horizon. Other tasks are evaluated immediately when interaction ends. We provide further analysis of interaction budgets in \Cref{sec:budget-analysis}.

\subsection{Evaluation Metrics}
\label{sec:metrics}

We use mean Pass@1(\%) as main metric, where a successful attempt must pass all applicable state, answer, report, memory, and temporal/workflow checks. For partial/infeasible tasks we also report reason-code accuracy (RC) and semantic report pass rate (Sem.). For memory-related tasks we report separate success rates for memory-conditioned (Read) Control/Goal and memory-write (Write) atoms.
Eff-A and Eff-T count satisfied state assertions and exactly matched query answers per tool call and interaction turn, respectively, for further tool call efficiency inspection.
Full metrics are defined in \Cref{app:metrics}.

\begin{table}[H]

\centering
\begin{threeparttable}

\scriptsize
\setlength{\tabcolsep}{2pt}
\renewcommand{\arraystretch}{1}
\begin{tabular}{lccccccccccccc}
\toprule
\multirow{2.5}{*}{\textbf{Model}} & \multirow{2.5}{*}{\textbf{Home}$^\dagger$}& \multirow{2.5}{*}{\textbf{Car}$^\dagger$}& \multirow{2.5}{*}{\textbf{Phone}$^\dagger$} & \multirow{2.5}{*}{\textbf{Overall}$^\dagger$}& \multirow{2.5}{*}{\textbf{Eff-A}}& \multirow{2.5}{*}{\textbf{Eff-T}} & \multicolumn{4}{c}{\textbf{Partial/Infeasible}} & \multicolumn{3}{c}{\textbf{Memory}} \\
\cmidrule(lr){8-11}\cmidrule(l){12-14}
 &  &   &  &   &   &  & Inf.$^\dagger$ & Part.$^\dagger$ & RC & Sem. & Mem.$^\dagger$ & Read & Write \\
\midrule
Claude Haiku 4.5\tnote{*}~\citep{claudehaiku45} & 63.5 & 43.6 & 33.7 & \cellcolor{blue!10}51.6 & 0.13 & 0.32 & 45.3 & 59.4 & 50.9 & 50.6 & 43.0 & 69.3 & 40.9 \\
DeepSeek V4.1 Flash\tnote{*}~\citep{deepseekv41flash} & 78.3 & 74.1 & 91.4 & \cellcolor{blue!10}79.6 & 0.19 & 0.54 & 75.3 & 82.2 & 77.9 & 76.0 & 65.5 & 93.9 & 58.0 \\
Gemini 2.5 Flash-Lite\tnote{*}~\citep{gemini25flashlite} & 72.7 & 69.0 & 85.3 & \cellcolor{blue!10}74.1 & 0.22 & 0.40 & 78.8 & 83.2 & 80.4 & 80.4 & 57.5 & 91.2 & 53.4 \\
Gemini 3.8 Flash\tnote{*}~\citep{gemini38flash} & \textbf{86.5} & \textbf{87.0} & 97.3 & \cellcolor{blue!10}\textbf{88.8} & 0.22 & 0.50 & 92.4 & 96.0 & 93.9 & 93.5 & 75.5 & 96.5 & 68.2 \\
GLM-5.3 Flash\tnote{*}~\citep{glm53flash} & 71.3 & 65.8 & 87.9 & \cellcolor{blue!10}72.9 & 0.21 & 0.59 & 74.7 & 76.2 & 75.6 & 73.4 & 60.0 & 96.5 & 55.7 \\
GPT-5.6 Luna~\citep{gpt56family} & 72.4 & 70.4 & 84.1 & \cellcolor{blue!10}74.1 & 0.22 & 0.69 & 66.3 & 69.3 & 68.3 & 67.8 & 51.3 & 83.6 & 54.5 \\
GPT-5.6 Terra~\citep{gpt56family} & 81.4 & 78.6 & 91.7 & \cellcolor{blue!10}82.6 & 0.23 & 0.66 & 63.5 & 75.6 & 68.0 & 67.8 & 60.7 & 94.7 & 63.3 \\
GPT-5.6 Sol~\citep{gpt56family} & 76.9 & 73.0 & 93.7 & \cellcolor{blue!10}79.1 & 0.23 & 0.68 & 69.4 & 80.9 & 73.9 & 73.9 & 63.3 & 97.1 & 48.1 \\
GPT-6 Luna~\citep{gpt6luna} & 56.0 & 57.1 & 48.3 & \cellcolor{blue!10}54.8 & 0.21 & 0.64 & 68.4 & 39.9 & 71.0 & 70.0 & 24.8 & 76.3 & 62.1 \\
GPT-6 Sol~\citep{gpt6sol} & 75.9 & 76.1 & 96.9 & \cellcolor{blue!10}80.2 & 0.23 & 0.63 & 85.1 & 93.1 & 88.4 & 88.4 & 69.8 & 97.7 & 62.9 \\
GPT-6.1 Terra~\citep{gpt61terra} & 75.6 & 74.3 & 97.1 & \cellcolor{blue!10}79.5 & 0.24 & 0.78 & 90.2 & 99.0 & 93.8 & 93.8 & 64.0 & 98.5 & 48.9 \\
GPT-6.1 Sol~\citep{gpt61sol} & 78.7 & 76.0 & 98.0 & \cellcolor{blue!10}81.7 & 0.25 & 0.63 & 87.5 & 94.7 & 90.3 & 90.3 & 70.8 & 98.8 & 62.1 \\
GPT-6 Astra~\citep{gpt6astra} & 79.2 & 75.8 & \textbf{98.2} & \cellcolor{blue!10}82.0 & 0.26 & 0.63 & 89.2 & 97.0 & 92.5 & 92.5 & 72.3 & 99.4 & 62.5 \\
\midrule
Qwen3.5-0.8B~\citep{qwen35} & 1.1 & 2.5 & 3.7 & \cellcolor{blue!10}2.0 & 0.03 & 0.03 & 0.4 & 0.0 & 0.3 & 0.3 & 0.0 & 0.0 & 1.5 \\
Qwen3.5-2B~\citep{qwen35} & 0.6 & 2.1 & 3.5 & \cellcolor{blue!10}1.6 & 0.00 & 0.02 & 0.0 & 0.0 & 0.0 & 0.0 & 0.0 & 0.0 & 0.0 \\
Qwen3.5-4B~\citep{qwen35} & 35.4 & 36.7 & 29.7 & \cellcolor{blue!10}34.7 & 0.15 & 0.17 & 45.5 & 18.8 & 41.5 & 40.4 & 7.8 & 26.0 & 34.1 \\
Qwen3.5-9B~\citep{qwen35} & 38.1 & 41.2 & 47.2 & \cellcolor{blue!10}40.8 & 0.18 & 0.20 & 60.6 & 41.3 & 54.7 & 53.7 & 13.2 & 35.4 & 44.7 \\
Qwen3.5-27B~\citep{qwen35} & 53.8 & 51.1 & 65.0 & \cellcolor{blue!10}55.3 & 0.18 & 0.20 & 64.9 & 69.0 & 66.4 & 65.9 & 34.8 & 43.6 & 51.9 \\
MiniCPM5-1B~\citep{minicpm5_1b} & 0.6 & 2.1 & 3.5 & \cellcolor{blue!10}1.6 & 0.01 & 0.02 & 0.0 & 0.0 & 0.0 & 0.0 & 0.0 & 0.0 & 0.0 \\
MiniCPM5-2B~\citep{minicpm5_2b} & 5.4 & 11.1 & 13.5 & \cellcolor{blue!10}8.7 & 0.05 & 0.10 & 3.7 & 1.3 & 4.0 & 3.7 & 1.0 & 7.0 & 10.2 \\
Gemma-4-E2B-it~\citep{gemma4} & 0.8 & 2.2 & 3.1 & \cellcolor{blue!10}1.7 & 0.03 & 0.02 & 1.0 & 0.0 & 1.2 & 0.5 & 0.0 & 0.0 & 0.0 \\
Gemma-4-E4B-it~\citep{gemma4} & 3.9 & 10.9 & 11.9 & \cellcolor{blue!10}7.6 & 0.05 & 0.04 & 14.1 & 0.7 & 9.2 & 8.5 & 2.0 & 3.2 & 9.1 \\
\bottomrule
\end{tabular}
\begin{tablenotes}[flushleft]
\item[*] One sample per task for these models due to our budget constraints; all other models use three samples per task.
\item[$\dagger$] Mean Pass@1 percentages. Inf./Part./Mem.: 170 infeasible, 101 partial, and 200 memory-related tasks.
\end{tablenotes}
\end{threeparttable}
\caption{Main results on \ours. Higher is better.}
\label{tab:main}
\end{table}

\subsection{Main Results}
\label{sec:main-results}

We present the main results in \Cref{tab:main}. Recent closed-source models demonstrate strong performance: Gemini 3.8 Flash leads overall at 88.8\%, followed by GPT-5.6 Terra (82.6\%), GPT-6 Astra (82.0\%), and GPT-6.1 Sol (81.7\%). Rankings vary across domains, with Gemini leading Home and Car at 86.5\% and 87.0\%, while Astra leads Phone at 98.2\%. Small open-weight models struggle with task completion: the evaluated MiniCPM5 and Gemma-4 variants achieve only 1.6-8.7\% mean Pass@1, revealing a substantial gap in reliable control on edge devices.

\Needspace{0.34\textheight}
\begin{wrapfigure}{r}{0.35\textwidth}
\centering
\includegraphics[width=\linewidth]{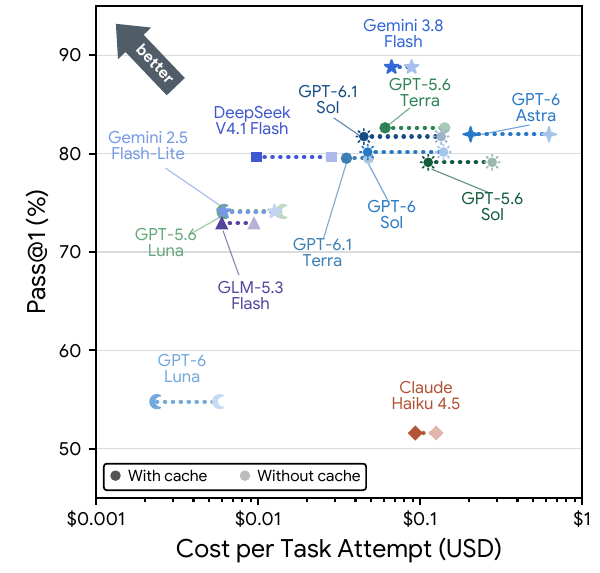}
\caption{Cost-performance trade-off across API models.}
\label{fig:api-cost-performance}
\end{wrapfigure}
\para{Similar performance conceals distinct strengths and failure modes.}
Contemporary GPT models across capability tiers achieve similar overall success rates, with Astra performing better on feasibility assessment, coordination of partially executable requests, and memory use. GPT-5.6 Terra, however, has substantial advantages on counting (Q5) and enumeration (Q6), achieving 90.6\% and 85.5\%, compared with Astra's 37.5\% and 4.3\% (\Cref{fig:subtype_results}). The case study in \Cref{app:failure-cases}, \Cref{case:astra-enumeration}, connects Astra's low enumeration score to its interpretation of the output requirements: 129 of its 132 failed attempts contain the correct target list together with an extra baseline answer. The agent treats an intermediate lookup as an additional query to answer. This pattern suggests that stronger reasoning does not necessarily translate into higher task success when output requirements are misinterpreted.

\para{Efficiency and cost at similar performance.}
Beyond these task-specific differences, we examine interaction efficiency, measured by Eff-A and Eff-T, alongside inference cost in \Cref{fig:api-cost-performance}. GPT-6.1 Terra achieves a similar overall success rate to GPT-5.6 Sol (79.5\% vs.\ 79.1\%), with higher Eff-A and Eff-T at only about 31\% of Sol's cache-discounted inference cost. \Cref{app:agent-cost} details the cost calculation. These results show that similar overall success rates can be achieved with greater interaction efficiency and substantially lower inference cost.

\Needspace{0.36\textheight}
\subsection{Instruction-Style Ablation}
\label{sec:style_ablation}

\begin{wrapfigure}{r}{0.35\textwidth}
\centering
\includegraphics[width=\linewidth]{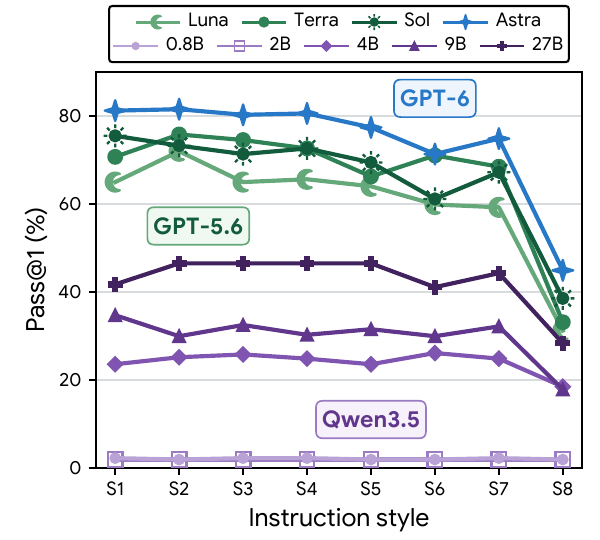}
\caption{Mean Pass@1 across instruction styles on validation set.}
\label{fig:style_ablation}
\end{wrapfigure}
We evaluate 9 models across 8 instruction styles on validation set (\Cref{fig:style_ablation}). Direct instructions (S2) achieve the highest average Pass@1, improving GPT-5.6 Luna by 7.0 percentage points over Template (S1). This pattern favors clear, explicit wording that reduces semantic ambiguity and the need to infer task requirements. Average performance declines with colloquial Folk instructions (S6) and falls sharply under Implicit instructions (S8). S6 preserves concrete requirements despite informal phrasing, whereas S8 deliberately leaves more details unstated. Distracting instructions (S7) remain broadly comparable to Conversational instructions (S5), suggesting relative robustness to irrelevant asides for most of the models. \Cref{tab:ap_styles} provides the complete results.

\para{Implicit instructions still permit partial achievement of task requirements.}
From S1 to S8, Astra's Pass@1 drops by 36.3 pp (-44.7\%), whereas device-property accuracy, answer accuracy, and list-answer recall decline by 22.8 pp (-23.2\%), 14.4 pp (-15.4\%), and 13.1 pp (-13.3\%), respectively (\Cref{tab:ap_implicit_metrics}). These smaller declines indicate that the model still achieves parts of the user intents even when it cannot satisfy the original targets whose details have been omitted.

\subsection{Goal Achievement and Interaction Budget}
\label{sec:budget-analysis}

\begin{figure}[t]
\centering
\includegraphics[width=\linewidth]{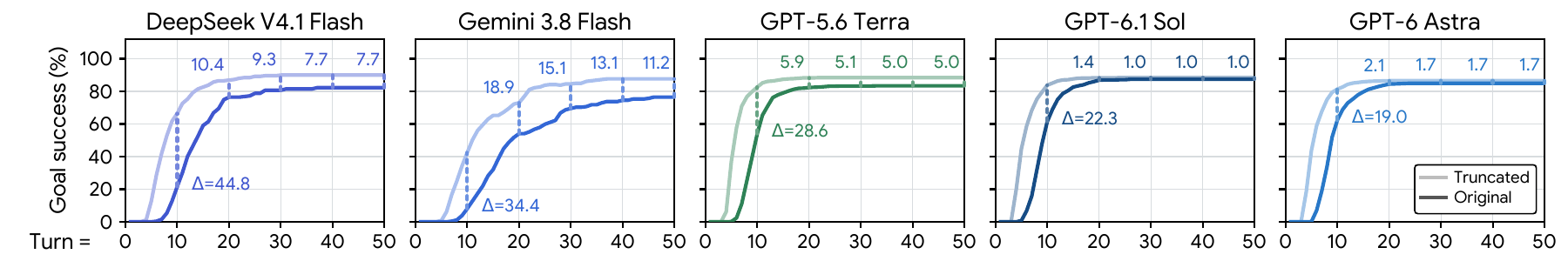}
\caption{Goal success across turn budgets. Original counts goals in trajectories voluntarily completed by turn $k$; Truncated stops interaction at $k$ and evaluates after simulation to the fixed horizon.}
\label{fig:budget-analysis}
\end{figure}

We assess long-horizon interaction for 5 top-performance models on 180 tasks containing 259 feasible Goal atoms. We replay and extend 83 truncated runs to at most 50 turns. \Cref{fig:budget-analysis} reveals two patterns. First, Truncated exceeds Original by 19.0-44.8 pp at turn 10, indicating that agents often continue interacting when their existing actions already suffice for eventual goal attainment. This suggests \textbf{conservative stopping behavior}. Second, most models' Truncated Goal-success curves largely plateau by turn 20. Extending the cutoff to 50 turns improves each GPT model by only 0.13 pp and DeepSeek by 3.1 pp, while Gemini gains 14.7 pp. The 20-turn budget thus \textbf{balances performance and interaction cost} for most selected models, emphasizing \textbf{agents' ability to complete tasks efficiently within limited turns}. \Cref{app:budget-method} details the protocol.

\begin{figure}[h]
  \centering
  \includegraphics[width=\linewidth]{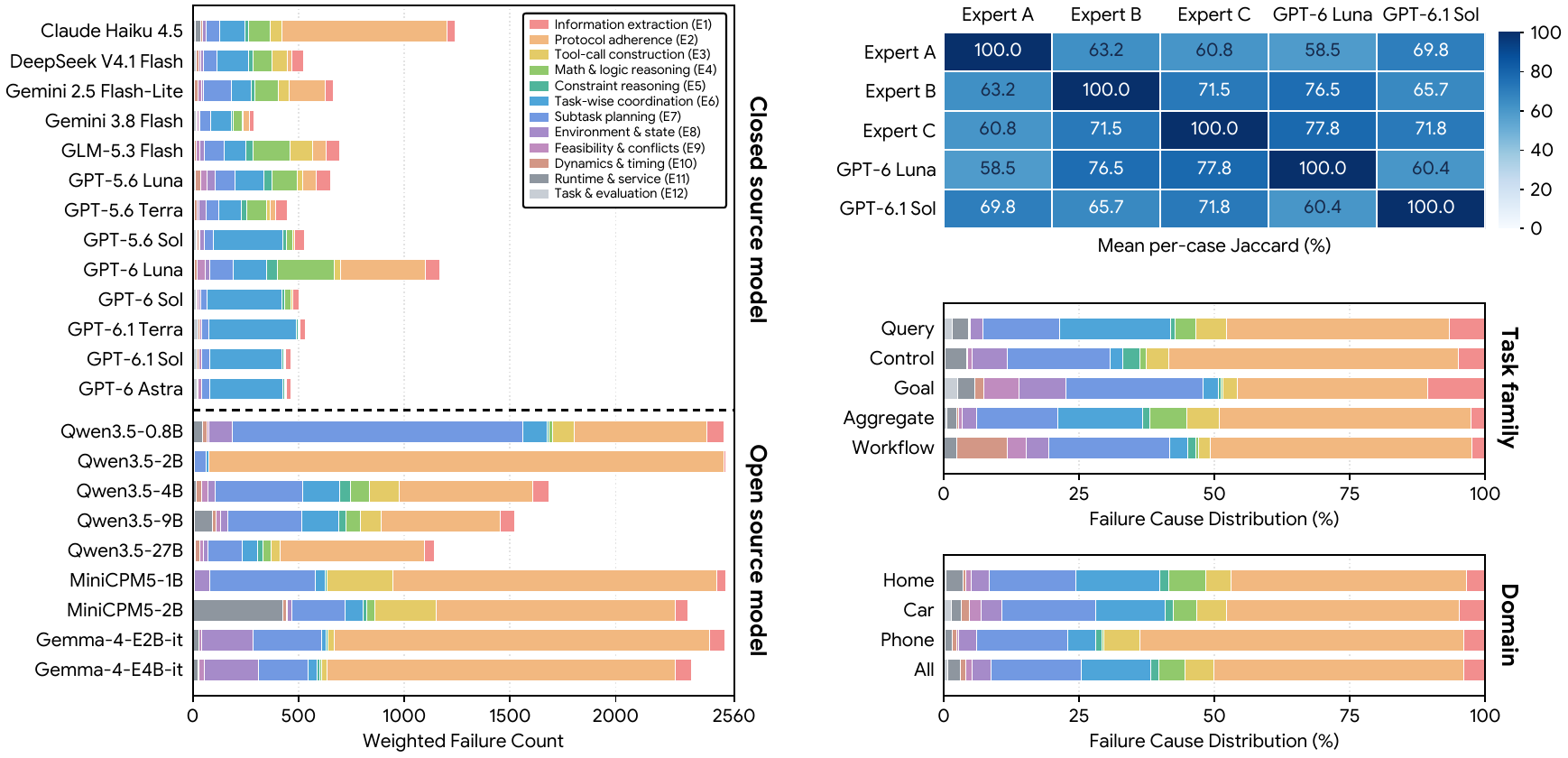}
  \caption{Failure cause statistics. \textit{Left}: weighted counts; \textit{bottom right}: normalized contribution within failures; \textit{upper right}: case-wise agreement in a separate 200-case annotation audit.}
  \label{fig:errors_combined}
\end{figure}

\subsection{Error Analysis}
\label{sec:error-analysis}

We analyze failed first attempts across all 22 models using 12 categories in four groups (\Cref{tab:failure-taxonomy}), with each attempt contributing total weight one across its ordered causes. Consistent with the main results, stronger models execute individual device operations reliably but struggle with interpreting complex requests and coordinating subtasks. Task understanding and coordination (E6) account for 73.5\% and 75.2\% of weighted failures for GPT-6.1 Sol and GPT-6 Astra, respectively. Dynamic and temporal reasoning (E10) also accounts for 9.3\% of weighted failures within Workflow tasks, highlighting the demands of timing and synchronization. Smaller models face more basic barriers in following interaction protocols (E2) and constructing valid tool calls (E3), which can prevent even individual subtasks from being completed. E2 alone accounts for 96.8\% of Qwen3.5-2B's weighted failures. \Cref{app:failure-method} provides the attribution protocol and representative case studies.

\section{World Distillation Scalability and Efficiency Analysis}
\label{sec:scaling}

\para{Capability coverage and token efficiency.}
To quantify the efficiency of world distillation, we compare World Distillation (WD) and Codex on constructing a RobotVacuum class from scratch with three levels of external knowledge: a brief description (L1, 0.06k chars), a functional overview (L2, 0.77k chars), and complete protocol knowledge (L3, 89.8k chars). We also summarized capability requirements from the full protocol.
Both methods use GPT-5.6 Sol, with Codex evaluated at low and xhigh reasoning effort. Experiment settings are detailed in \Cref{app:onboarding}.

As shown in \Cref{tab:construction}, WD uses \textbf{0.9-66.0\% and 46.2-92.5\% fewer tokens} than Codex low and xhigh across the three knowledge levels. At comparable token budgets, WD achieves higher complete coverage than Codex low across properties, functions, and effect/event requirements at both L2 and L3. Compared with Codex xhigh, WD uses \textbf{51.5-86.7\% fewer tokens per completely covered requirement}. At L1, the brief description provides little protocol detail, and all methods achieve limited coverage. WD's coverage advantage emerges with richer knowledge at L2 and L3.

\begin{table}[H]
\centering

\begin{minipage}[t]{0.54\textwidth}
\centering
\vspace{0pt}
\begin{threeparttable}

\scriptsize
\setlength{\tabcolsep}{1pt}

\begin{tabular}{lcccccc}
\toprule
\textbf{Method} & \textbf{\#Devices} & \textbf{\#Properties}$^\dagger$ &
\textbf{\#Functions}$^\dagger$ & \textbf{\#Events}$^\dagger$ &
\textbf{Tokens} & \textbf{Min.} \\
\midrule
L1-Codex low   & 1 & 7 (6/1)   & 13 (1/7)  & 1 (6/5)   & 1.48M & 10.09 \\
L1-Codex xhigh & 1 & 12 (6/1)  & 21 (1/7)  & 2 (9/3)   & 6.74M & 12.48 \\
L1-WD          & 1 & 7 (3/1)   & 12 (0/5)  & 2 (6/3)   & 0.50M & 11.45 \\
\midrule
L2-Codex low   & 1 & 23 (0/19) & 39 (6/23) & 0 (14/15) & 1.50M & 10.16 \\
L2-Codex xhigh & 1 & 23 (18/1) & 39 (6/23) & 1 (14/16) & 5.46M & 11.71 \\
L2-WD          & 1 & 25 (18/0) & 49 (11/17)& 2 (30/8)  & 1.15M & 24.31 \\
\midrule
L3-Codex low   & 1 & 19 (9/14) & 29 (11/18)& 1 (12/18) & 2.77M & 14.40 \\
L3-Codex xhigh & 1 & 18 (23/0) & 29 (22/7) & 2 (19/16) & 5.09M & 13.64 \\
L3-WD          & 1 & 25 (22/0) & 57 (19/10)& 2 (30/7)  & 2.74M & 26.95 \\
\midrule
Home & 54 & 330 & 616 & 49 & 11.57M & 48.09 \\
Car & 38 & 177 & 333 & 29 & 4.83M & 33.45 \\
Phone & 22 & 123 & 217 & 17 & 2.76M & 32.33 \\
\bottomrule
\end{tabular}

\begin{tablenotes}[flushleft]
\scriptsize
\item[$\dagger$] Parentheses report complete/partial coverage of
23 property/query dimensions, 29 protocol functionalities, and
39 effect/event requirements, respectively.
\end{tablenotes}
\end{threeparttable}
\caption{Comparison of device model construction.}
\label{tab:construction}
\end{minipage}
\hfill
\begin{minipage}[t]{0.44\textwidth}
\centering
\vspace{0pt}
\includegraphics[width=0.9\linewidth]{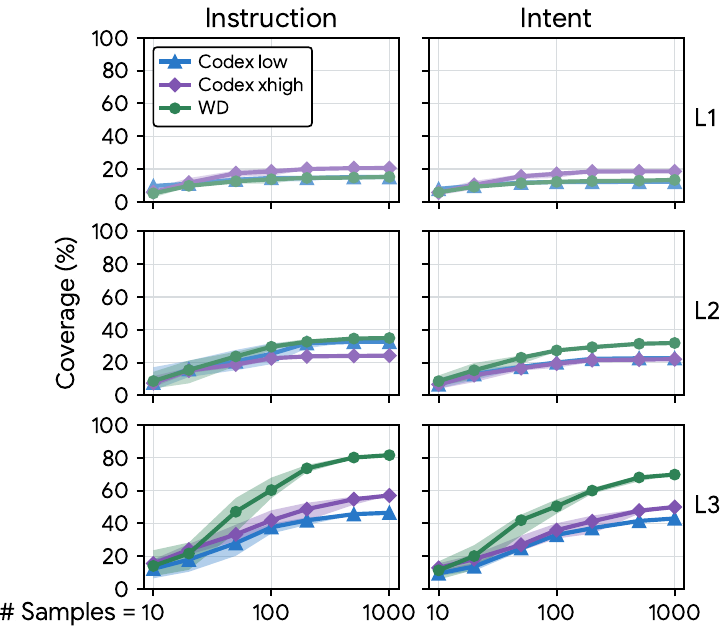}
\captionof{figure}{Coverage of instructions and intents as task sample size grows.} %
\label{fig:robot-instructions}
\end{minipage}

\end{table}

\para{Coverage of reference instructions and intents.}
We extend the comparison to task coverage using 854 instructions drawn from industrial test cases that share the same source as our reference knowledge. We categorize these instructions into 172 intents. For each setting, we use the resulting device model to generate 1,000 validated tasks, then repeatedly sample subsets of varying sizes to estimate instruction and intent coverage (\Cref{fig:robot-instructions}).
With complete protocol information (L3), WD achieves the broadest coverage, reaching 81.6\% of reference instructions and 69.8\% of intents, supporting its use in industrial settings with detailed device specifications. Under limited information (L1), Codex xhigh covers 20.6\% of instructions compared with WD's 15.2\%, as it draws more effectively on pretrained knowledge to supplement sparse external input.
\Cref{app:robot-coverage-protocol} details the coverage evaluation.

\begin{figure}[H]
  \centering
  \includegraphics[width=\linewidth]{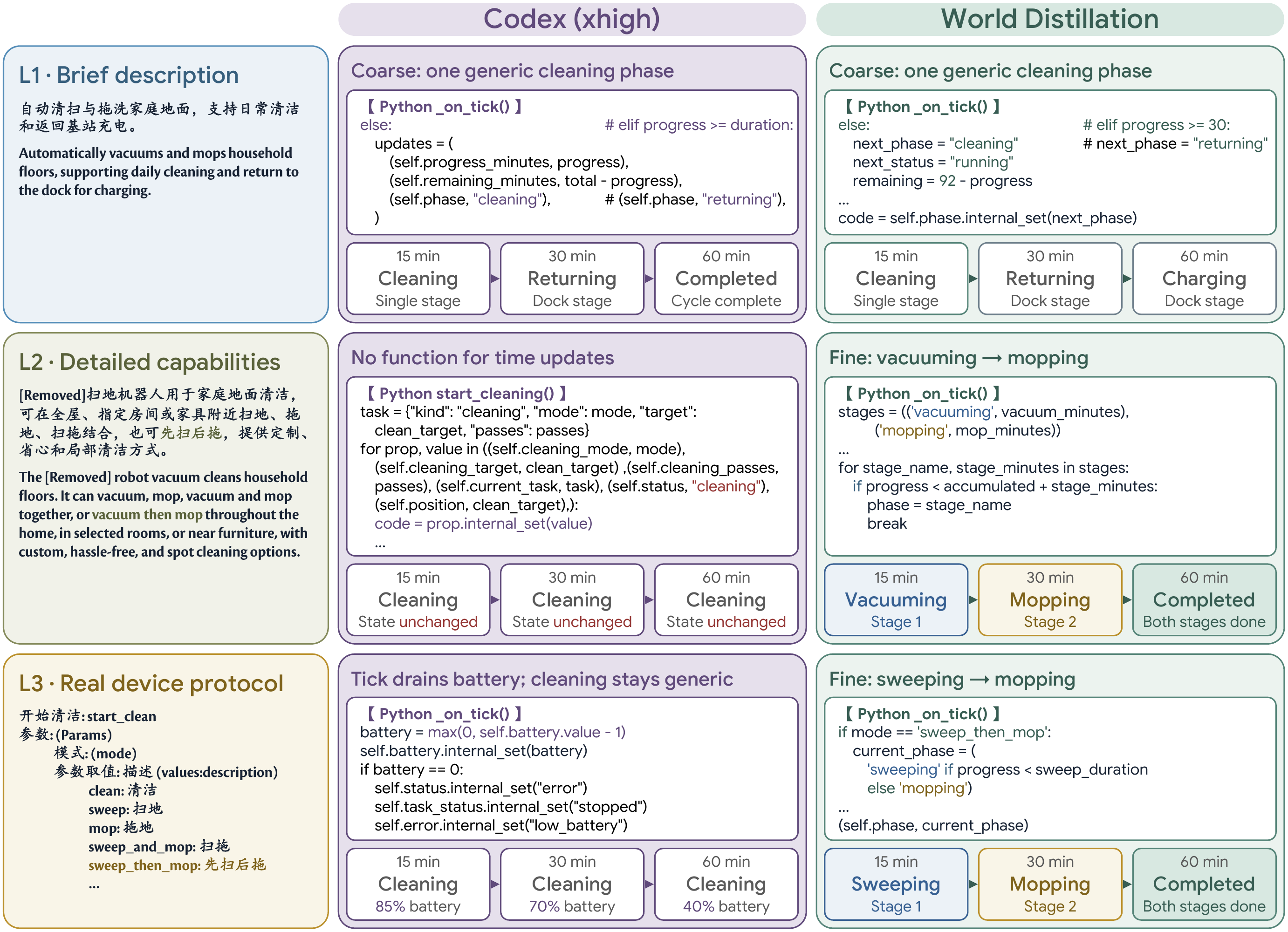}
  \caption{RobotVacuum external knowledge, implementation code, and replayed cleaning states across L1-L3.}
  \label{fig:robotvacuum_case_study}
\end{figure}

\para{Executable dynamics across knowledge levels.}
\Cref{fig:robotvacuum_case_study} compares Codex xhigh and WD under the same frozen World runtime. At L1, the description does not specify sequential sweeping and mopping, and both methods implement a single \texttt{cleaning} phase before returning to the dock. At L2, WD correctly implements the newly specified vacuum-then-mop behavior through explicit vacuuming and mopping phases. Codex records \texttt{vacuum\_then\_mop} as a task parameter but lacks an \texttt{\_on\_tick} handler to advance the process, while \texttt{start\_cleaning} keeps the state fixed at \texttt{cleaning}. At L3, WD adopts the protocol-defined \texttt{sweep\_then\_mop} mode and implements the corresponding sweeping and mopping phases, while Codex updates battery consumption but retains a generic cleaning state. WD switches to mopping at 30 minutes and completes cleaning at 60 minutes under its default simulation settings. These durations are simulation parameters, while the phase ordering follows the supplied knowledge. This case demonstrates how WD translates richer behavioral descriptions and protocol definitions into executable process transitions.

\Needspace{0.28\textheight}
\begin{wraptable}{r}{0.35\textwidth}
\centering
\begin{threeparttable}

\scriptsize
\setlength{\tabcolsep}{1pt}
\begin{tabular}{c@{\hspace{1pt}}ccccc}
\toprule
\textbf{\#Attempts}  & \textbf{Spec.} & \textbf{Temp.} & \textbf{Impl.} & \textbf{UniT.} & \textbf{Total} \\
\midrule
1 & 54 & 109 & 92 & 72 & 327 \\
2 & 27 & 5 & 16 & 31 & 79 \\
3 & 18 & 0 & 2 & 7 & 27 \\
4+ & 15 & 0 & 4 & 4 & 23 \\
\midrule
Total & 251 & 119 & 154 & 179 & 703 \\
\;\;\#Calls & 493 & 0 & 134 & 202 & 829 \\
\;\;Tokens & 10.71M & 0 & 3.96M & 4.50M & 19.16M \\
\bottomrule
\end{tabular}

\begin{tablenotes}[flushleft]
\scriptsize
\item[] Spec./Temp./Impl./UniT.: specification, template, implementation, and tests.
\end{tablenotes}
\end{threeparttable}
\caption{Construction stages.}
\label{tab:construction_stages}
\end{wraptable}

\para{Domain construction cost.}
We construct device models for all three domains using a job planner with 16 workers per domain, with a combined construction time of \textbf{113.87 minutes}. The full pipeline consumes \textbf{19.16M tokens}, costing approximately \textbf{\$120.09} before cache discounts. By automating specification, implementation, and validation, WD largely reduces the manual effort required to expand executable environments. Our construction prioritizes efficient coverage of device capabilities for agent testing and capability transfer, rather than production-ready device integration. Stage-level analysis shows that most classes require only one recorded execution for template generation, implementation, and testing, while specification generation requires the most repeated executions (\Cref{tab:construction_stages}). These results demonstrate a practical path to expanding behavioral coverage across domains and supporting the evaluation and development of device-control agents.

\section{Conclusion and Future Work}
\label{sec:conclusion}

We introduced \ours, a scalable benchmark for environment awareness and device control across Home, Car, and Phone. World distillation converts device knowledge into reusable executable components for world and task construction (\Cref{sec:distillation}, \Cref{sec:synthesis}). The benchmark combines diverse tasks, a shared interaction protocol, and state and process checks (\Cref{sec:coverage}, \Cref{sec:interaction}, \Cref{sec:metrics}). Evaluation of 22 models yields a best mean Pass@1 of 88.8\%, while similar overall scores conceal differences in capabilities, efficiency, and cost (\Cref{sec:main-results}). Instruction-style analysis favors explicit wording and reveals partial task achievement under implicit requests (\Cref{sec:style_ablation}). Failure analysis distinguishes protocol and tool-use barriers in smaller models from complex intent interpretation and coordination failures in stronger models (\Cref{sec:error-analysis}). The budget analysis shows that agents often continue interacting after their actions suffice for eventual Goal attainment, with limited gains beyond 20 turns for most selected models (\Cref{sec:budget-analysis}). Construction experiments demonstrate efficient capability expansion and broader instruction and intent coverage with detailed protocol knowledge (\Cref{sec:scaling}). Future work will explore targeted training with generated tasks to improve the reliability and efficiency of smaller models for on-device deployment.

\clearpage
\subsection*{AI use statement}
Generative AI is used in the benchmark construction pipeline for capability specifications, behavior-code completion, test generation, repair, and instruction rewriting, and in configured semantic evaluation. AI coding and writing assistants like Codex, Claude Code, ChatGPT are also used to inspect source code and existing results, draft and revise manuscript text, prepare illustrative scripts and figures, and check document compilation. \Cref{app:prompts} provides the prompts used for construction, evaluation, and analysis.

\subsection*{Reproducibility statement}
\Cref{app:method,app:benchmark} describe world distillation, temporal semantics, task taxonomy, interaction protocols, and scoring rules. \Cref{app:results,app:failure-method,app:agent-cost} provide disaggregated results, failure attribution, and cost accounting, respectively. \Cref{app:prompts} documents the prompts used for construction, evaluation, and analysis. The benchmark code, synthetic data, and prompts are available at \url{https://github.com/X-LANCE/SynthDCB}.

\subsection*{Ethics Statement}

\para{Human Involvement}
All human-involved checks, including inspection of generated artifacts, were conducted by the authors. No crowdworkers or students participated in this work.

\para{Privacy}
\ours~uses synthetic simulated device worlds. Scenario object identifiers, initial states, target states, and user instructions are artificial and are not linked to real people, households, or devices. The auxiliary RobotVacuum coverage analysis uses reference instructions from the external knowledge. These materials are distinct from the synthetic benchmark tasks and from personal user interaction logs. Generated artifacts undergo schema validation, replay validation, and author inspection; any personally identifying content encountered is excluded.

\para{Intended Use and Potential Risks}
\ours~provides benchmark code, generated device-class specifications, simulated scenarios, tool schemas, reference trajectories, supervised tool-call records, and evaluation scripts for benchmark evaluation and controlled agent-training research. Generation is constrained to daily device-control operations over predefined device properties and functions, and invalid or offensive generated content is excluded if encountered during validation and inspection. Results on \ours~measure tool-use competence in simulated environments and do not establish readiness for real-world autonomous device control. Direct transfer to physical systems, especially appliances, vehicles, locks, heating systems, and other safety-critical devices, may introduce safety risks. Real-world deployment requires independent safety validation, permission management, human oversight, rollback mechanisms, and domain-specific constraints on high-risk actions.

\para{Licensing}
Publicly released artifacts will be accompanied by a license or terms-of-use file. External models and software artifacts are used under their respective licenses or provider terms, and proprietary models are accessed through API services.
The RobotVacuum reference materials, including protocol definitions, functional descriptions, and the instruction corpus, contain proprietary industrial information and will not be publicly released due to intellectual property restrictions.

\newpage

\bibliography{cite}
\bibliographystyle{plainnat}

\newpage

\appendix

\addtocontents{toc}{\protect\setcounter{tocdepth}{2}}
\begingroup
\edef\savedtocdepth{\number\value{tocdepth}}
\setcounter{tocdepth}{-1}
\renewcommand{\contentsname}{}
\small
\setlength{\parskip}{0pt}
\tableofcontents
\setcounter{tocdepth}{\savedtocdepth}
\endgroup
\clearpage

\clearpage
\section{World Distillation}
\label[appendix]{app:method}

\subsection{Capability Specification and Interfaces}

A device specification contains metadata, properties, functions, and events, with optional inheritance and process declarations.
Properties declare their type, default, access mode, units, and admissible values.
The implementation supports booleans, integers, numbers, strings, and structured collections.
Numeric declarations can specify bounds, step origins, and precision; collection declarations specify their structure.
Read-only, read/write, and disabled properties are distinguished explicitly.
Inheritance resolves updates and disabled members before the effective specification is used for code generation or loading.

Functions declare a JSON parameter schema, mutation status, read and write sets, and semantics such as observation, assignment, collection mutation, or process control.
Preconditions and argument sources connect those operations to the state they require.
The loader checks that referenced properties exist and that the declaration agrees with the implementation.
For an instance exposing a subset of properties, a function is exposed only if all its declared property dependencies are present.
This rule also determines which actions task construction can use.

Events describe continuous effects, effects following an action, or responses to declared dependencies.
Home space effects operate on temperature, humidity, illuminance, noise, and air quality; each domain catalog determines its own legal targets and units.
Phone modules update local state and declared dependencies without Home space effects.
Optional process specifications describe modes, phases, phase durations, and persisted progress variables.
Public methods return a status and response; event hooks return a status and a list of effects.

\subsection{Component Generation and Repair}

Generation is organized around artifacts with explicit dependencies.
The type inventory and placement constraints identify candidate components.
The metadata stage combines the request, domain property catalog, inherited specification, schema, and retrieved references.
Template assembly constructs the class and property wrappers, after which code completion implements public actions and event hooks.
Test generation uses the specification and implementation to exercise valid calls, invalid inputs, boundary conditions, power behavior, and effects.

The production gate performs deterministic schema and declaration checks, inspects generated source against its execution restrictions, checks specification--implementation consistency, and executes generated unit tests in an isolated subprocess.
Errors are returned to the earliest relevant stage, preserving the requested capabilities and bounded repair settings.
Attempt records retain failures, auxiliary generation calls, usage, and repair lineage.

\subsection{Runtime and Temporal Semantics}
\label[appendix]{app:runtime}

The runtime executes commands against a validated world and maintains a separate audit of command results, state changes, and time ticks.
Device actions update their own state and may emit space effects.
During a minute tick, hooks observe a common pre-tick snapshot, and the reducer combines their effects before committing environmental updates.
Contributions replace the previous contribution from the same source; additive deltas are accumulated; approach effects move a baseline toward declared targets at bounded rates.
The domain contract validates allowed combinations and units, then clamps and rounds values using the property's specification.
These rules define the simulator's behavior abstraction and are not a calibrated physical model.

Command execution and time advancement use state checkpoints so rejected or failing operations do not leave partial committed updates.
The simulator advances in integer minutes.
Waiting, declared action durations, and configured turn costs determine elapsed time; model latency does not advance the simulation clock.
Independent calls can share a turn, while calls requiring a new observation must occur in a subsequent turn.
Device-tool discovery is also required in a preceding turn under the benchmark configuration.

The agent's finish signal ends interaction immediately.
When a private target defines an evaluation horizon, background simulation continues to that fixed horizon without further agent actions.
Goal construction derives a horizon from the reference duration $d$:
\begin{equation}
 t_{\mathrm{eval}}=t_0+\max\!\left(60,\;60\left\lceil\frac{2d}{60}\right\rceil\right).
\end{equation}
Workflow targets supply their own horizon and transition relations.
The evaluation horizon remains fixed when the agent waits or finishes. The public information supplied to the agent is retained in each turn's context.
Without a horizon, final-state checks use the interaction-end state.
Previously submitted answers retain their submission-time interpretation.

Workflow scoring records the first qualifying transition for each required event.
Ordered relations use minute and audit sequence to disambiguate events in the same minute; other relations check bounded delays, synchronization, specified times, or strict ``before'' deadlines.
The legacy target-level \texttt{by}/\texttt{within} constraint uses an inclusive deadline.
This distinction is represented in the target and instruction.

\subsection{World and Task Construction Algorithms}

The world constructor samples a supported layout, assigns compatible components, samples legal property values, applies initialization operations, and loads the resulting world through the shared contract.
A task may derive a perturbed initial state from a source world; its effective state is saved with the task.
The following algorithm summarizes the active task-building path.

\begin{enumerate}
\item Load a world and build an inventory of exposed properties, operations, processes, and environmental effects.
\item Sample a task family and candidate requirements from that inventory. Preview operations to bind their resulting states, answers, and relevant time transitions.
\item Construct the private target and a reference plan. For a negative requirement, retain its construction reason; for a partial task, retain executable requirements as well.
\item Replay the plan from a fresh copy of the effective initial world. Run schema, public/private, instruction-alignment, termination, and evaluator checks.
\item Accept the complete bundle when its reference satisfies the target. Preserve unsuccessful candidates in the generation records.
\item Produce instruction styles from the accepted requirements, retaining the same base-task identity and paired assets.
\end{enumerate}

Goal probing has a finite search budget.
Exhausting that budget records a construction failure; an unreachable-goal label requires separate construction evidence.
Aggregate construction merges compatible requirements, targets, memories, and reference operations from the same effective world, then validates the merged task.

\subsection{Construction Experiment Setup}
\label[appendix]{app:onboarding}

We compare WD and Codex on RobotVacuum construction using external knowledge of varying levels of richness: a short summary (L1), a detailed functional description (L2), and the complete device protocol (L3).
Both use GPT-5.6 Sol with identical device descriptions and framework references; Codex uses low or xhigh reasoning, while WD uses the provider default.
Source inspection and behavioral checks assess complete and partial coverage against a common checklist summarized from the protocol.
The generated RobotVacuum device is not included in the final Home domain of \ours{} and is used solely for analysis purposes.

\para{Codex setting.}
The experiments use Codex version 0.165.1. For each experiment, we create a copy containing the complete \ours{} source code, expert references and domain metadata for the Home domain, and a brief task instruction. We place the corresponding external knowledge in \texttt{DEVICE.md} in the root directory.

We launch the task in \texttt{/goal} mode and enable ``Approve for me'' as the permission protocol.

The experiments are conducted on the same local server and under the same network conditions.

\para{Cost estimation.}
For the Codex baselines, we measure the elapsed time from the start of goal execution to completion, along with the token usage reported by Codex. For WD, we measure the elapsed time from launching the job planner to completing the entire pipeline and count all token usage during this period, including all recorded generation, retry, review, repair, and failure-analysis calls. Token costs are calculated using details provided in \Cref{app:agent-cost}, without cache discounts.

\subsection{Generated-Task Coverage}
\label[appendix]{app:robot-coverage-protocol}

The external knowledge references include an instruction corpus with 861 records and 854 distinct instruction texts. We retain all 854 texts, including spatial references, gestures, following requests, clarification, and slot-filling interactions. A frozen protocol-assisted taxonomy groups them into 172 intents, merging synonymous expressions and concrete parameter values while retaining actions, directions, query targets, modes, ordering, recurrence, and compound or conversational requests.

Each method supplies 1,000 validated tasks at each knowledge level. At sample sizes 10, 20, 50, 100, 200, 500, and 1,000, five samples without replacement yield the mean coverage and minimum-maximum bands ploted in \Cref{fig:robot-instructions}.

We use GPT-6.1 Sol to match the original task templates and execution evidence against the reference instructions, preserving action and direction while abstracting concrete parameter values. Instruction coverage counts the union of matched texts. Intent coverage counts intents with at least one matched text.

\subsection{Additional Construction Results}
The source-capability checklist contains 23 property/query dimensions, 29 protocol functions, and 39 effect/event requirements. At L2, WD fully covers 18/23 property dimensions and 14/72 parameters, compared with 0/23 and 5/72 for Codex low. At L3, the corresponding results are 22/23 and 50/72 for WD, and 9/23 and 27/72 for Codex low. WD covers 23/29 action effects at both levels, compared with 14/29 and 12/29 for Codex low. It also fully covers five of eight temporal effects and both required events.

The execution probes show timed cleaning, pause/resume, scheduled starts, and progressive charging in WD-L2. WD-L3 supports ten process modes with their required arguments. Remaining gaps include incompatible optional-argument combinations and missing schedule triggers. The original Codex-L2 constructor prevents World restoration; the task-coverage experiment uses its repaired copy. Codex-L3 updates battery use during cleaning but lacks the explicit sweep-then-mop transitions shown in \Cref{fig:robotvacuum_case_study}.

\clearpage
\section{Benchmark Specification and Evaluation}
\label[appendix]{app:benchmark}

\subsection{Device and Capability Inventory}
\label[appendix]{app:inventory}

\Cref{tab:ap_domain_stat,tab:ap_devices} report the per-domain component and dataset statistics and the device-class inventory, respectively.
The validation and test splits draw from the shared source-world pool; the split unit is the base task. Tasks from the two sets may share the same world instance.
An atom (task) is an entry in the constructed task graph, including prerequisite observations, memory writes, and infeasible or blocked requirements.
Repeated occurrences across tasks are counted separately, while instruction variants and evaluation attempts do not add atoms.

The base/derived class counts are 36/18 for Home, 30/8 for Car, and 21/1 for Phone.
Across the test-task configurations, Home covers 17 space types, with 5--22 spaces and 10--125 device instances per task world.
Car configurations have 11--13 spaces and 40--52 devices; Phone has two spaces and 21 module instances.
Task-local initialization specifies each task's starting conditions.
Component-class counts describe the reusable library, while instance counts describe composed worlds.

\begin{table}[H]
\centering
\begin{threeparttable}

\scriptsize
\setlength{\tabcolsep}{2.5pt}
\begin{tabular}{lcccccccc}
\toprule
\textbf{Domain} & \textbf{Classes} & \textbf{Functions} & \makecell[c]{\textbf{Properties}\\\textbf{(Writable)}} & \textbf{Events} & \makecell[c]{\textbf{Source}\\\textbf{worlds}} & \makecell[c]{\textbf{Pool}\\\textbf{tasks}} & \makecell[c]{\textbf{Valid}\\\textbf{tasks / atoms}}& \makecell[c]{\textbf{Test}\\\textbf{tasks / atoms}} \\
\midrule
Home & 54 & 616 & 330 (229) & 49 & 60 & 15,877 & 150 / 2,189 & 1,281 / 17,211 \\
Car & 38 & 333 & 177 (83) & 29 & 40 & 9,918 & 105 / 1,087 & 768 / 7,023 \\
Phone & 22 & 217 & 123 (41) & 17 & 30 & 4,514 & 59 / 316 & 511 / 2,730 \\
\midrule
Total & 114 & 1,166 & 630 (353) & 95 & 130 & 30,309 & 314 / 3,592 & 2,560 / 26,964 \\
\bottomrule
\end{tabular}
\end{threeparttable}
\caption{Component inventory and dataset size.}
\label{tab:ap_domain_stat}
\end{table}

\begin{table}[H]
\centering
\begin{threeparttable}

\scriptsize
\begin{tabular}{@{}l>{\raggedright\arraybackslash}p{\dimexpr\linewidth-32pt-2\tabcolsep\relax}@{}}
\toprule
\textbf{Domain} & \makecell[t{>{\raggedright\arraybackslash}p{\linewidth}}]{\textbf{Component classes}} \\
\midrule
Home & \makecell[t{>{\raggedright\arraybackslash}p{\linewidth}}]{Air Conditioner, Air Exchanger, Air Fryer, Air Purifier, Bathroom Heater, Bathroom Mirror, Bluetooth Speaker, Cat Litter Box, Ceiling Fan, Coffee Maker, Color Lamp, Cooling Only Air Conditioner, Curtain, Dehumidifier, Dishwasher, Door Lock, Dryer, Drying Rack, Dual Layer Curtain, Electric Water Heater, Evaporative Humidifier, Fan, Floor Washer, Fragrance Diffuser, Gas Water Heater, Hat Lamp Bathroom Heater, Heater, Humidifier, Induction Cooker, Kettle, Lamp, Microwave, Monitor, Oven, Pet Feeder, Pet Water Dispenser, Projector, Range Hood, Refrigerator, Rice Cooker, Robot Vacuum, Simple Lamp, Speaker, Surveillance Camera, Television, Toilet, Ultrasonic Humidifier, Warm Air Bathroom Heater, Washer Dryer, Washing Machine, Water Dispenser, Water Heater, Water Purifier, Window Opener} \\
\midrule
Car & \makecell[t{>{\raggedright\arraybackslash}p{\linewidth}}]{Adaptive Damping, Air Suspension, Ambient Light, Cabin Light, Car Display, Car Refrigerator, Charging, Climate Control, Cooling Heating Car Refrigerator, Dashcam, Door, Drive Settings, Driver Assistance, Dynamic Ambient Light, Electrified Drive Settings, Exterior Light, External Power Supply, Fragrance, Fuel System, Head Up Display, Hybrid Drive Settings, Infotainment, Memory Seat Comfort, Mirror, Navigation, Parking Camera, Rear Entertainment Display, Seat Comfort, Seat Heater, Seat Ventilator, Steering Wheel, Sunroof, Sunshade, Tailgate, Trunk Door, Vehicle Security, Window, Wiper} \\
\midrule
Phone & \makecell[t{>{\raggedright\arraybackslash}p{\linewidth}}]{Accessibility, Alarm, Battery, Bluetooth, Connectivity, Dimmable Flashlight, Display, Flashlight, Focus, Hotspot, Location, Media Player, Notifications, Privacy, Rotation, Screen Capture, Sound, Stopwatch, System, System Settings, Timer, Wi Fi} \\
\bottomrule
\end{tabular}
\end{threeparttable}
\caption{Device classes by domain.}
\label{tab:ap_devices}
\end{table}

\clearpage
\subsection{System Tools and Device Examples}
\label[appendix]{app:tools}

\Cref{tab:ap_tools} lists the system and memory tools, device dispatcher, and Home device examples with their arguments and conditions.

\begin{table}[H]
\centering
\begin{threeparttable}

\scriptsize
\setlength{\tabcolsep}{4pt}
\begin{tabular}{@{}>{\raggedright\arraybackslash}p{0.32\linewidth}>{\raggedright\arraybackslash}p{0.26\linewidth}>{\raggedright\arraybackslash}p{\dimexpr0.42\linewidth-4\tabcolsep\relax}@{}}
\toprule
\textbf{Tool / device function} & \textbf{Arguments} & \textbf{Purpose and conditions} \\
\midrule
\multicolumn{3}{@{}l}{\textbf{Session tools}} \\
\rowcolor{black!5}
\quad\texttt{session.get\_capabilities} & \texttt{\{\}} & Return the domain, support for spaces, devices, and relations, and the simulation time unit. \\
\quad\texttt{session.get\_devices} & \texttt{space\_id} & List devices in a space with their IDs, classes, locations, and metadata. \\
\rowcolor{black!5}
\quad\texttt{session.get\_relations} & \texttt{\{\}} & Return the relations declared in the domain metadata. \\
\quad\texttt{session.get\_spaces} & \texttt{\{\}} & List space IDs, types, and whether each space is virtual. \\
\rowcolor{black!5}
\quad\texttt{session.get\_space\_state} & \texttt{space\_id} & Read a space's properties and metadata. \\
\quad\texttt{session.get\_space\_devices} & \texttt{space\_id} & List the IDs and classes of devices in a space. \\
\rowcolor{black!5}
\quad\texttt{session.get\_device\_tools} & \texttt{device\_id} & Retrieve device functions, parameter schemas, and read/write semantics. Required before invoking those functions in a later turn. \\
\quad\texttt{session.sleep\_time} & \texttt{[minutes]} & Advance simulation by a nonnegative integer number of minutes; omission uses the configured default. \\
\rowcolor{black!5}
\quad\texttt{session.submit\_answer} & \texttt{value} & Submit one typed JSON answer per query, in instruction order. \\
\quad\texttt{session.submit\_report} & \texttt{status}, \texttt{reason\_code}, \texttt{description}, \texttt{[affected\_properties]} & Report one infeasible or blocked requirement, with its reason and supporting explanation. \\
\rowcolor{black!5}
\quad\texttt{session.finish} & \texttt{\{\}} & End interaction; background simulation continues to the fixed evaluation horizon when one applies. \\
\midrule
\multicolumn{3}{@{}l}{\textbf{Memory tools}} \\
\rowcolor{black!5}
\quad\texttt{memory.read} & \texttt{\{\}} & Read memory IDs and contents in the current scope. \\
\quad\texttt{memory.create} & \texttt{content}, \texttt{[user\_id]} & Create a natural-language memory record, optionally associated with a user. \\
\rowcolor{black!5}
\quad\texttt{memory.update} & \texttt{memory\_id}, \texttt{content} & Replace a previously read record's content using its returned memory ID. \\
\midrule
\multicolumn{3}{@{}l}{\textbf{Device dispatcher}} \\
\rowcolor{black!5}
\quad\texttt{dcb\_device\_command} & \texttt{target}, \texttt{function}, \texttt{parameters} & Invoke a discovered device function with its exact target and parameter schema; \texttt{parameters} is a JSON object. \\
\midrule
\multicolumn{3}{@{}l}{\textbf{Home device examples}} \\
\rowcolor{black!5}
\quad\textit{Read (AirConditioner)}\newline\hspace*{1em}\texttt{get\_target\_temperature} & \texttt{\{\}} & Read the target temperature; requires power on and mode \texttt{cool}, \texttt{heat}, or \texttt{auto}. \\
\quad\textit{Write (AirConditioner)}\newline\hspace*{1em}\texttt{set\_target\_temperature} & \texttt{\{"value": 22\}} & Set the target to $22\,^{\circ}\mathrm{C}$ (integer range: $16$--$30$), under the same power and mode conditions. \\
\rowcolor{black!5}
\quad\textit{Complex control (WasherDryer)}\newline\hspace*{1em}\texttt{start\_cycle} & \texttt{\{"mode":}\newline\texttt{"wash\_and\_dry"\}} & Start a wash-and-dry cycle; requires power on and status \texttt{idle} or \texttt{completed}. Jointly update mode, status, phase, and timing; simulated time advances the cycle. \\
\bottomrule
\end{tabular}
\end{threeparttable}
\caption{System tools and device examples.}
\label{tab:ap_tools}
\end{table}

\clearpage
\subsection{Task Taxonomy and Coverage}

\Cref{tab:ap_taxonomy} reports validation and test task counts by domain, family, and subtype.

\begin{table}[H]
\centering
\begin{threeparttable}

\scriptsize
\begin{tabular}{lcccccc}
\toprule
\multirow{2.5}{*}{\textbf{Type / Subtype}} & \multicolumn{2}{c}{\textbf{Home}} & \multicolumn{2}{c}{\textbf{Car}} & \multicolumn{2}{c}{\textbf{Phone}} \\
\cmidrule(lr){2-3}\cmidrule(lr){4-5}\cmidrule(lr){6-7}
 & \textbf{Valid} & \textbf{Test} & \textbf{Valid} & \textbf{Test} & \textbf{Valid} & \textbf{Test} \\
\midrule
\textbf{Query} & \textbf{37} & \textbf{175} & \textbf{21} & \textbf{100} & \textbf{15} & \textbf{25} \\
\quad property.device & 25 & 10 & 11 & 10 & 15 & 25 \\
\quad property.space & 2 & 57 & 2 & 19 & 0 & 0 \\
\quad space.occupants & 1 & 10 & 0 & 0 & 0 & 0 \\
\quad space.person\_location & 1 & 10 & 0 & 0 & 0 & 0 \\
\quad target.comparison.top1 & 2 & 28 & 2 & 14 & 0 & 0 \\
\quad target.comparison.topk & 2 & 17 & 2 & 22 & 0 & 0 \\
\quad target.enum & 2 & 26 & 2 & 20 & 0 & 0 \\
\quad target.quantity & 2 & 17 & 2 & 15 & 0 & 0 \\
\midrule
\textbf{Control} & \textbf{19} & \textbf{261} & \textbf{17} & \textbf{156} & \textbf{11} & \textbf{105} \\
\quad complex & 5 & 97 & 3 & 77 & 3 & 43 \\
\quad multi\_property & 2 & 42 & 3 & 24 & 2 & 17 \\
\quad multi\_target & 4 & 24 & 5 & 17 & 2 & 10 \\
\quad single & 8 & 98 & 6 & 38 & 4 & 35 \\
\midrule
\textbf{Goal} & \textbf{11} & \textbf{94} & \textbf{11} & \textbf{65} & \textbf{0} & \textbf{0} \\
\quad complex & 2 & 10 & 3 & 10 & 0 & 0 \\
\quad multi\_property & 2 & 19 & 2 & 10 & 0 & 0 \\
\quad multi\_target & 5 & 11 & 4 & 10 & 0 & 0 \\
\quad single & 2 & 54 & 2 & 35 & 0 & 0 \\
\midrule
\textbf{Aggregate} & \textbf{71} & \textbf{669} & \textbf{44} & \textbf{359} & \textbf{23} & \textbf{326} \\
\quad control & 2 & 51 & 1 & 30 & 1 & 21 \\
\quad control\_goal & 2 & 10 & 1 & 10 & 0 & 0 \\
\quad control\_query & 53 & 521 & 33 & 248 & 20 & 273 \\
\quad control\_query\_goal & 4 & 10 & 2 & 10 & 0 & 0 \\
\quad goal & 4 & 10 & 1 & 10 & 0 & 0 \\
\quad query & 5 & 57 & 5 & 41 & 2 & 32 \\
\quad query\_goal & 1 & 10 & 1 & 10 & 0 & 0 \\
\midrule
\textbf{Workflow} & \textbf{12} & \textbf{82} & \textbf{12} & \textbf{88} & \textbf{10} & \textbf{55} \\
\quad at & 1 & 9 & 1 & 10 & 1 & 8 \\
\quad before & 1 & 9 & 1 & 9 & 1 & 9 \\
\quad condition\_delay & 1 & 10 & 1 & 10 & 1 & 10 \\
\quad condition\_follow & 2 & 9 & 1 & 10 & 2 & 10 \\
\quad condition\_window & 1 & 10 & 1 & 10 & 3 & 10 \\
\quad environment\_before\_process & 1 & 9 & 2 & 10 & 0 & 0 \\
\quad environment\_deadline & 1 & 6 & 1 & 9 & 0 & 0 \\
\quad single & 3 & 10 & 3 & 10 & 2 & 8 \\
\quad synchronize & 1 & 10 & 1 & 10 & 0 & 0 \\
\midrule
\textbf{Total tasks} & 150 & 1281 & 105 & 768 & 59 & 511 \\
\textbf{Total atoms} & 2,189 & 17,211 & 1,087 & 7,023 & 316 & 2,730 \\
\bottomrule
\end{tabular}
\begin{tablenotes}[flushleft]
\item[] All feasibility categories are included. Atoms include prerequisites and infeasible requirements.
\end{tablenotes}
\end{threeparttable}
\caption{Task distribution by domain, family, and subtype.}
\label{tab:ap_taxonomy}
\end{table}

\subsection{Feasibility and Memory Settings}

Negative tasks have explicit construction reasons: \texttt{SPACE\_TARGET\_NOT\_FOUND} or \texttt{DEVICE\_TARGET\_NOT\_FOUND}, a requested space or device is absent; \texttt{SPACE\_PROPERTY\_NOT\_SUPPORTED} or \texttt{DEVICE\_PROPERTY\_NOT\_SUPPORTED}, a property or capability is unsupported; \texttt{INVALID\_VALUE}, a value violates the declared domain; \texttt{UNREACHABLE\_GOAL}, a goal has a supported unreachable construction; or \texttt{BLOCKED\_BY\_DEPENDENCY}, a dependent requirement is blocked.
Partial tasks retain at least one independently executable requirement.
Their expected reports preserve multiplicity, so two requests failing for the same reason still require two reports.

Memory fixtures contain scoped synthetic natural-language records with stable IDs and optional user bindings.
Read-dependent targets specify the required records; write-dependent targets specify the observation and fact to preserve.
Read-before-write and observation-before-dependent-action checks use the audit sequence.
Memory-dependent tasks comprise 112 tasks with memory-conditioned Control/Goal atoms and 88 tasks with a memory-write atom.
Two of the former contain two memory-conditioned atoms, giving 114 read-dependent atoms; each write task contains exactly one write atom.
An update's prerequisite read is part of the write requirement.

\subsection{Instruction Variants and Alignment Checks}

The style ablation compares Template and seven rewritten instructions for each shared base task, effective world, private target, and reference evidence.
Direct states the requirements plainly; Concise compresses wording; Polite changes register; Conversational adds natural conversational phrasing; Folk uses colloquial expressions; Distracting adds irrelevant conversational material; and Implicit can rely on contextual information or indirect intent.
Automated alignment checks preserve required entities, literal values, quantities, temporal conditions, and other target references for six non-implicit rewrite styles.

\subsection{Complete Metric Definitions}
\label[appendix]{app:metrics}

\Cref{tab:ap_metrics} separates primary checks from diagnostic measures.
Object, space, answer, report, memory, and time component averages use only applicable attempts.
Missing evaluation evidence for a required object or space check receives zero.
The primary success computation treats inapplicable answer, report, memory, time, and workflow checks as satisfied.

\begin{table}[H]
\centering
\begin{threeparttable}

\scriptsize
\setlength{\tabcolsep}{4pt}
\begin{tabular}{>{\raggedright\arraybackslash}p{0.22\linewidth}>{\raggedright\arraybackslash}p{0.71\linewidth}}
\toprule
\textbf{Metric} & \textbf{Definition} \\
\midrule
\rowcolor{black!5}
Task success & Conjunction of the task's applicable state, answer, report, memory, time, and workflow checks. A failed background simulation prevents successful state evaluation. \\
State success & All specified device, space, and runtime assertions hold at the prescribed observation time. Operators include equality, ranges, inequalities, change, preservation, and structured collection change. \\
\rowcolor{black!5}
Property accuracy & Fraction of specified property assertions satisfied. Object and space means use only applicable attempts; the combined mean includes empty targets as correct. \\
Answer success / accuracy & All expected responses match after order-preserving alignment, with the required number of submissions. Accuracy is exact aligned matches divided by the larger of expected and submitted counts. \\
\rowcolor{black!5}
List answer recall & Mean recall for list-valued expected answers under the alignment; a missing list response receives zero. This is a soft diagnostic. \\
Report success / accuracy & Exact multiset agreement of reason codes; accuracy divides matched multiplicities by the larger expected/submitted count. Report count accuracy measures count-only agreement. \\
\rowcolor{black!5}
Report semantic score & A separate LLM judgment of request-specific status, reason, and explanation. Semantic task success additionally requires this judgment when applicable. \\
Memory correctness & Required record access and write ordering, plus expected mutations. When enabled, a semantic judge checks that written content preserves the required facts. ROUGE-L is a separate diagnostic. \\
\rowcolor{black!5}
Memory atom success & Read: target assertion and memory-access check for a memory-conditioned Control/Goal atom. Write: mutation, ordering, and semantic checks for a create/update atom. \\
Time / workflow success & Specified completion times, duration constraints, or relations among first qualifying state transitions satisfy the private target. \\
\rowcolor{black!5}
Tool-call success rate & Successful command records divided by all command records within an attempt, averaged across attempts. Empty and missing records follow the conventions in the text. \\
False completion & The agent explicitly finishes an attempt that fails task success. Budget termination is tracked separately. \\
\rowcolor{black!5}
Replay / side effects & Replay compares snapshots when reference evidence is supplied; unexpected state changes and time-plan validity provide additional diagnostics. They do not enforce an identical agent trajectory. \\
Cost & Agent tokens and API cost per attempt, with and without recorded cache discounts (\Cref{app:agent-cost}). \\
\rowcolor{black!5}
Eff-A / Eff-T & Mean per-attempt number of satisfied state assertions and exactly matched query answers per tool call / interaction turn. Missing reports and zero denominators contribute zero. \\
\bottomrule
\end{tabular}
\end{threeparttable}
\caption{Primary checks and diagnostic metrics.}
\label{tab:ap_metrics}
\end{table}

Answer alignment uses dynamic programming to preserve the order of questions while aligning missing or extra responses.
Scalar and structured exact-answer checks retain JSON value types.
The canonical ordering required for entity-selection answers is specified in the agent prompt.
Raw list-valued device properties retain their observed order.

Memory-write semantics, when enabled, are incorporated into memory correctness; transport or malformed judge responses are evaluation errors.
User-scoped memory resolves references such as ``me'' or ``the driver'' to the current user; a valid write preserves the requested observation and its user binding.

The saved tool-call success field, \texttt{action\_success\_rate}, counts \texttt{[PASS]} records across discovery, observation, and session commands.
An evaluated attempt with no command records receives one; an attempt with no evaluation report receives zero.

\para{Memory atom success.}
\Cref{tab:main} separates complete-task Pass@1 on the 200 memory-dependent tasks from atom success.
A memory-conditioned Control/Goal atom passes when its target state assertion and the saved memory-access check both pass.
The access check verifies retrieval of the required records before dependent execution; its shared ordering requirement applies to both atoms in the two tasks with two memory bindings.
The Read column pools 114 such atoms across Home/Car/Phone (84/5/25).
The Write column pools 88 create/update atoms (56/11/21), using the saved memory verdict, which includes the required mutation, observation-before-write and read-before-update ordering, and semantic content judgment.
The saved reports provide these memory and state checks, but do not assign success to every atom in the general task graph; we therefore report the directly recoverable memory atom scores separately from full-task Pass@1.

\para{Achievement efficiency.}
Let $\mathcal P_b$ be the scored state assertions and $\mathcal Q_b$ the expected query answers for task $b$.
Its hard achievement count is
\begin{equation}
 \begin{gathered}
 C_b=\sum_{p\in\mathcal P_b}\mathbf 1[p\text{ passes}]
     +\sum_{q\in\mathcal Q_b}\mathbf 1[q\text{ exactly matches}],\\
 \mathrm{Eff\mbox{-}A}=\frac{1}{N}\sum_{b=1}^{N}\frac{C_b}{A_b},
 \qquad
 \mathrm{Eff\mbox{-}T}=\frac{1}{N}\sum_{b=1}^{N}\frac{C_b}{T_b}.
 \end{gathered}
 \label{eq:efficiency}
\end{equation}
where $N=2{,}560$, $A_b$ counts recorded tool calls excluding \texttt{session.finish}, and $T_b$ counts interaction turns.
Calls include discovery, observations, device operations, answers, reports, memory operations, and waits.
Each zero-denominator ratio and each attempt without an evaluation report contributes zero.
Exact answer matches use the saved order-preserving alignment.
These metrics use hard state checks; they do not introduce numeric-distance partial credit.
Their numerator measures state/answer progress, while reports, memory, and workflow conditions are covered by their own success metrics.
Wholly infeasible tasks have no state/answer items and contribute zero achievement.
We average per-attempt ratios, rather than dividing aggregate achievement by aggregate cost.

\subsection{Interaction-Budget Protocol}
\label[appendix]{app:budget-method}

\Cref{fig:budget-analysis} evaluates 259 feasible Goal atoms in 180 tasks. We replay each trajectory through turn $k$, stop agent interaction, and simulate the world to the task's fixed evaluation horizon. Truncated counts Goal atoms attained after this simulation; Original counts those in trajectories voluntarily completed by $k$. Scores from earlier completed trajectories carry forward to later budgets.

The 83 truncated runs that reach the 20-turn limit continue from their saved world state and conversation, up to 50 turns. Replay verifies the restored state and command outcomes. Continuations use a parser that accepts multiline Thought; earlier parse failures remain unexecuted. The curves measure Goal attainment, including within mixed tasks.

\clearpage
\section{Complete Results}
\label[appendix]{app:results}

\subsection{Task Families}
\begin{table}[H]
\centering

\begin{threeparttable}

\scriptsize
\setlength{\tabcolsep}{4pt}
\begin{tabular}{lccccc}
\toprule
\textbf{Model} & \textbf{Query} $\uparrow$ & \textbf{Control} $\uparrow$ & \textbf{Goal} $\uparrow$ & \textbf{Aggregate} $\uparrow$ & \textbf{Workflow} $\uparrow$ \\
\midrule
Claude Haiku 4.5 & 58.3 & 66.7 & 57.9 & 43.1 & 54.2 \\
DeepSeek V4.1 Flash & 71.0 & 95.4 & 81.1 & 73.0 & 93.3 \\
Gemini 2.5 Flash-Lite & 69.7 & 93.5 & 73.0 & 66.8 & 80.0 \\
Gemini 3.8 Flash & 76.3 & 98.3 & 81.1 & 87.7 & 95.1 \\
GLM-5.3 Flash & 64.7 & 94.1 & 74.8 & 63.7 & 89.3 \\
GPT-5.6 Luna & 67.0 & 91.1 & 64.4 & 68.5 & 85.0 \\
GPT-5.6 Terra & 75.9 & 94.5 & 71.9 & 78.9 & 93.9 \\
GPT-5.6 Sol & 63.6 & 95.9 & 76.7 & 73.9 & 93.9 \\
GPT-6 Luna & 54.3 & 89.5 & 58.5 & 36.1 & 84.6 \\
GPT-6 Sol & 68.1 & 96.6 & 86.0 & 73.2 & 96.0 \\
GPT-6.1 Terra & 63.0 & 97.8 & 86.6 & 72.4 & 97.5 \\
GPT-6.1 Sol & 67.6 & 97.8 & 83.9 & 76.0 & 96.3 \\
GPT-6 Astra & 67.4 & 98.1 & 82.4 & 76.5 & 96.4 \\
\midrule
Qwen3.5-0.8B & 0.7 & 1.1 & 1.3 & 0.0 & 18.8 \\
Qwen3.5-2B & 0.0 & 0.0 & 0.0 & 0.0 & 18.7 \\
Qwen3.5-4B & 44.8 & 72.0 & 40.3 & 12.7 & 63.3 \\
Qwen3.5-9B & 29.8 & 79.2 & 46.1 & 22.3 & 74.7 \\
Qwen3.5-27B & 70.4 & 88.8 & 56.0 & 34.9 & 79.3 \\
MiniCPM5-1B & 0.0 & 0.0 & 0.0 & 0.0 & 18.7 \\
MiniCPM5-2B & 5.9 & 18.3 & 4.4 & 3.0 & 27.7 \\
Gemma-4-E2B-it & 0.4 & 0.2 & 0.8 & 0.0 & 17.2 \\
Gemma-4-E4B-it & 15.0 & 15.3 & 4.8 & 0.3 & 25.6 \\
\bottomrule
\end{tabular}
\end{threeparttable}
\caption{Test Pass@1 by task family (\%).}
\label{tab:ap_families}
\end{table}

\clearpage
\subsection{Task Subtypes}

\begin{figure}[H]
\centering
\rotatebox{90}{
    \includegraphics[width=0.9\textheight]{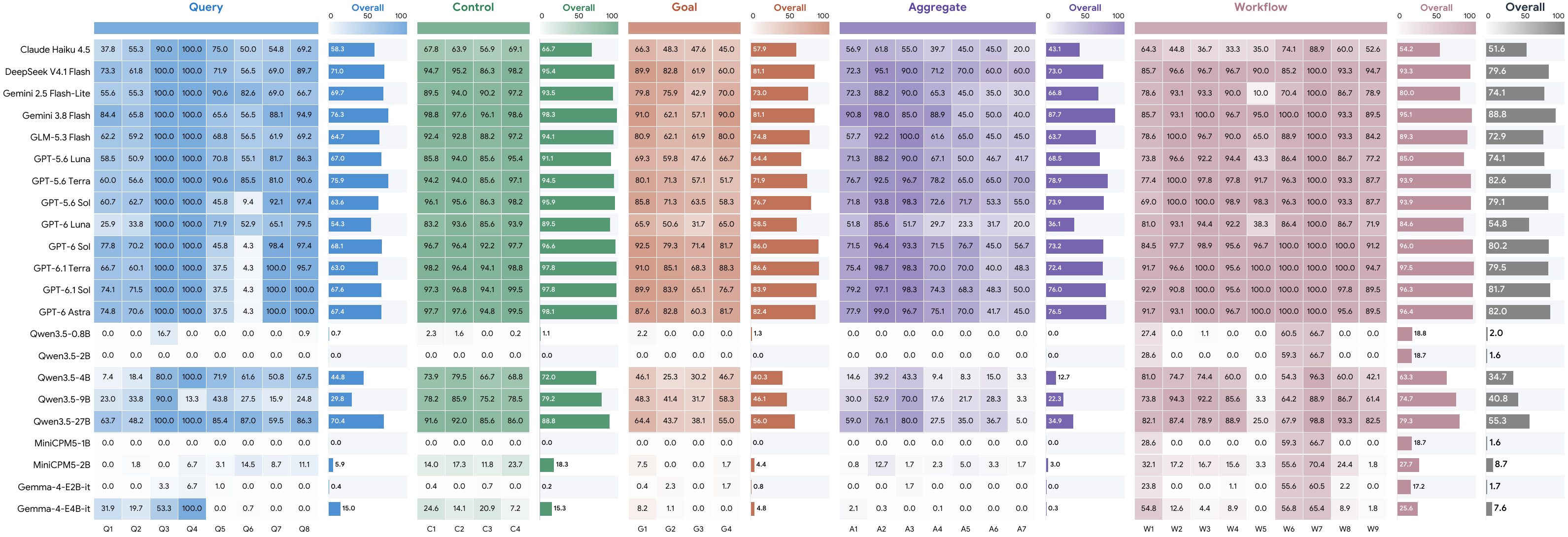}
}
\caption{Mean Pass@1 by task family and subtype.}
\label{fig:subtype_results}
\end{figure}

\subsection{Instruction Styles}
\begin{table}[H]

\centering
\begin{threeparttable}

\scriptsize
\setlength{\tabcolsep}{4pt}
\renewcommand{\arraystretch}{1}
\begin{tabular}{llccccccccc}
\toprule
\multirow{2.5}{*}{\textbf{Domain}} & \multirow{2.5}{*}{\textbf{Instruction style}} & \multicolumn{5}{c}{\textbf{Qwen3.5}} & \multicolumn{3}{c}{\textbf{GPT-5.6}} & \textbf{GPT-6} \\
\cmidrule(lr){3-7}\cmidrule(lr){8-10}\cmidrule(l){11-11}
& & \textbf{0.8B} & \textbf{2B} & \textbf{4B} & \textbf{9B} & \textbf{27B} & \textbf{Luna} & \textbf{Terra} & \textbf{Sol} & \textbf{Astra} \\
\midrule
\multirow{8}{*}{\textbf{Home}}
& Template (S1) & 1.3 & 1.3 & 23.3 & 34.0 & 44.0 & 65.3 & 68.0 & 72.7 & 76.0 \\
& Direct (S2) & 1.3 & 1.3 & 28.0 & 30.0 & 48.7 & 74.0 & 74.0 & 74.0 & 78.7 \\
& Concise (S3) & 1.3 & 1.3 & 27.3 & 30.7 & 50.0 & 64.0 & 74.0 & 74.0 & 78.7 \\
& Polite (S4) & 1.3 & 1.3 & 28.7 & 30.0 & 46.7 & 62.7 & 71.3 & 73.3 & 78.7 \\
& Conversational (S5) & 1.3 & 1.3 & 26.7 & 32.0 & 47.3 & 65.3 & 65.3 & 72.7 & 73.3 \\
& Folk (S6) & 1.3 & 1.3 & 29.3 & 27.3 & 40.7 & 60.7 & 70.0 & 61.3 & 67.3 \\
& Distracting (S7) & 2.0 & 1.3 & 30.7 & 33.3 & 44.0 & 60.7 & 66.0 & 66.0 & 72.0 \\
& Implicit (S8) & 1.3 & 1.3 & 20.0 & 20.0 & 30.0 & 32.7 & 35.3 & 40.7 & 42.0 \\
\midrule
\multirow{8}{*}{\textbf{Car}}
& Template (S1) & 3.8 & 2.9 & 24.8 & 36.2 & 36.2 & 61.9 & 67.6 & 72.4 & 81.0 \\
& Direct (S2) & 2.9 & 2.9 & 24.8 & 30.5 & 41.0 & 61.9 & 73.3 & 64.8 & 75.2 \\
& Concise (S3) & 3.8 & 2.9 & 25.7 & 29.5 & 40.0 & 60.0 & 69.5 & 60.0 & 73.3 \\
& Polite (S4) & 3.8 & 2.9 & 23.8 & 32.4 & 41.9 & 61.9 & 70.5 & 63.8 & 74.3 \\
& Conversational (S5) & 2.9 & 2.9 & 21.9 & 31.4 & 43.8 & 59.0 & 61.0 & 58.1 & 73.3 \\
& Folk (S6) & 2.9 & 2.9 & 25.7 & 33.3 & 37.1 & 54.3 & 71.4 & 55.2 & 66.7 \\
& Distracting (S7) & 2.9 & 2.9 & 25.7 & 30.5 & 41.0 & 53.3 & 66.7 & 60.0 & 66.7 \\
& Implicit (S8) & 2.9 & 2.9 & 19.0 & 17.1 & 28.6 & 30.5 & 31.4 & 35.2 & 42.9 \\
\midrule
\multirow{8}{*}{\textbf{Phone}}
& Template (S1) & 1.7 & 1.7 & 22.0 & 33.9 & 45.8 & 69.5 & 83.1 & 88.1 & 94.9 \\
& Direct (S2) & 1.7 & 1.7 & 18.6 & 28.8 & 50.8 & 84.7 & 84.7 & 86.4 & 100.0 \\
& Concise (S3) & 1.7 & 1.7 & 22.0 & 42.4 & 49.2 & 76.3 & 84.7 & 84.7 & 96.6 \\
& Polite (S4) & 1.7 & 1.7 & 16.9 & 27.1 & 54.2 & 79.7 & 79.7 & 86.4 & 96.6 \\
& Conversational (S5) & 1.7 & 1.7 & 18.6 & 30.5 & 49.2 & 69.5 & 78.0 & 81.4 & 94.9 \\
& Folk (S6) & 1.7 & 1.7 & 18.6 & 30.5 & 49.2 & 67.8 & 72.9 & 71.2 & 89.8 \\
& Distracting (S7) & 1.7 & 1.7 & 8.5 & 32.2 & 50.8 & 66.1 & 78.0 & 83.1 & 96.6 \\
& Implicit (S8) & 1.7 & 1.7 & 13.6 & 13.6 & 23.7 & 28.8 & 30.5 & 39.0 & 55.9 \\
\midrule
\multirow{8}{*}{\textbf{Overall}}
& Template (S1) & 2.2 & 1.9 & 23.6 & 34.7 & 41.7 & 65.0 & 70.7 & 75.5 & 81.2 \\
& Direct (S2) & 1.9 & 1.9 & 25.2 & 29.9 & 46.5 & 72.0 & 75.8 & 73.2 & 81.5 \\
& Concise (S3) & 2.2 & 1.9 & 25.8 & 32.5 & 46.5 & 65.0 & 74.5 & 71.3 & 80.3 \\
& Polite (S4) & 2.2 & 1.9 & 24.8 & 30.3 & 46.5 & 65.6 & 72.6 & 72.6 & 80.6 \\
& Conversational (S5) & 1.9 & 1.9 & 23.6 & 31.5 & 46.5 & 64.0 & 66.2 & 69.4 & 77.4 \\
& Folk (S6) & 1.9 & 1.9 & 26.1 & 29.9 & 41.1 & 59.9 & 71.0 & 61.1 & 71.3 \\
& Distracting (S7) & 2.2 & 1.9 & 24.8 & 32.2 & 44.3 & 59.2 & 68.5 & 67.2 & 74.8 \\
& Implicit (S8) & 1.9 & 1.9 & 18.5 & 17.8 & 28.3 & 31.2 & 33.1 & 38.5 & 44.9 \\
\bottomrule
\end{tabular}
\end{threeparttable}
\caption{Validation Pass@1 (\%) across all eight instruction styles.}
\label{tab:ap_styles}
\end{table}

\vspace{1.5cm}
\begin{table}[H]
\centering

\begin{threeparttable}

\scriptsize
\setlength{\tabcolsep}{4pt}
\begin{tabular}{l*{12}{>{\centering\arraybackslash}m{0.75cm}}}
\toprule
\multirow{2.5}{*}{\textbf{Model}} & \multicolumn{3}{c}{\textbf{Pass@1}} & \multicolumn{3}{c}{\textbf{Device-property acc.}} & \multicolumn{3}{c}{\textbf{Space-property acc.}} & \multicolumn{3}{c}{\textbf{Answer acc.}} \\
\cmidrule(lr){2-4}\cmidrule(lr){5-7}\cmidrule(lr){8-10}\cmidrule(lr){11-13}
& \textbf{Tpl.} & \textbf{Impl.} & {\boldmath$\Delta$} & \textbf{Tpl.} & \textbf{Impl.} & {\boldmath$\Delta$} & \textbf{Tpl.} & \textbf{Impl.} & {\boldmath$\Delta$} & \textbf{Tpl.} & \textbf{Impl.} & {\boldmath$\Delta$} \\
\midrule
GPT-6 Astra & 81.2 & 44.9 & -36.3 & 98.4 & 75.6 & -22.8 & 90.0 & 26.0 & -64.0 & 93.3 & 79.0 & -14.4 \\
GPT-5.6 Sol & 75.5 & 38.5 & -36.9 & 98.9 & 76.0 & -22.9 & 88.0 & 35.3 & -52.7 & 96.1 & 78.9 & -17.3 \\
GPT-5.6 Terra & 70.7 & 33.1 & -37.6 & 99.0 & 75.5 & -23.5 & 88.0 & 33.3 & -54.7 & 92.0 & 63.6 & -28.5 \\
GPT-5.6 Luna & 65.0 & 31.2 & -33.8 & 98.3 & 75.0 & -23.3 & 78.0 & 29.3 & -48.7 & 86.7 & 63.0 & -23.8 \\
\midrule
Qwen3.5-0.8B & 2.2 & 1.9 & -0.3 & 12.4 & 12.3 & -0.1 & 0.0 & 2.0 & +2.0 & 0.1 & 0.5 & +0.5 \\
Qwen3.5-2B & 1.9 & 1.9 & +0.0 & 12.3 & 12.3 & +0.0 & 0.0 & 0.0 & +0.0 & 0.0 & 0.0 & +0.0 \\
Qwen3.5-4B & 23.6 & 18.5 & -5.1 & 45.9 & 38.2 & -7.7 & 38.0 & 18.0 & -20.0 & 25.6 & 15.2 & -10.5 \\
Qwen3.5-9B & 34.7 & 17.8 & -16.9 & 57.0 & 45.0 & -12.0 & 50.0 & 20.0 & -30.0 & 32.0 & 16.6 & -15.4 \\
Qwen3.5-27B & 41.7 & 28.3 & -13.4 & 44.1 & 46.8 & +2.7 & 50.0 & 30.0 & -20.0 & 42.1 & 36.9 & -5.2 \\
\midrule
\textbf{Average} & 44.1 & 24.0 & -20.0 & 62.9 & 50.7 & -12.2 & 53.6 & 21.6 & -32.0 & 52.0 & 39.3 & -12.7 \\
\bottomrule
\end{tabular}
\medskip
\begin{tabular}{l*{12}{>{\centering\arraybackslash}m{0.75cm}}}
\toprule
\multirow{2.5}{*}{\textbf{Model}} & \multicolumn{3}{c}{\textbf{List-answer recall}} & \multicolumn{3}{c}{\textbf{Memory check success}} & \multicolumn{3}{c}{\textbf{Report reason-code acc.}} & \multicolumn{3}{c}{\textbf{Tool-call success}} \\
\cmidrule(lr){2-4}\cmidrule(lr){5-7}\cmidrule(lr){8-10}\cmidrule(lr){11-13}
& \textbf{Tpl.} & \textbf{Impl.} & {\boldmath$\Delta$} & \textbf{Tpl.} & \textbf{Impl.} & {\boldmath$\Delta$} & \textbf{Tpl.} & \textbf{Impl.} & {\boldmath$\Delta$} & \textbf{Tpl.} & \textbf{Impl.} & {\boldmath$\Delta$} \\
\midrule
GPT-6 Astra & 98.5 & 85.3 & -13.1 & 75.7 & 62.2 & -13.5 & 91.8 & 81.1 & -10.7 & 100.0 & 99.9 & -0.1 \\
GPT-5.6 Sol & 99.5 & 79.1 & -20.5 & 73.0 & 43.2 & -29.7 & 54.9 & 52.5 & -2.5 & 99.5 & 99.0 & -0.5 \\
GPT-5.6 Terra & 95.9 & 71.2 & -24.7 & 67.6 & 54.1 & -13.5 & 48.4 & 46.7 & -1.6 & 99.0 & 99.1 & +0.0 \\
GPT-5.6 Luna & 95.3 & 71.7 & -23.6 & 75.7 & 27.0 & -48.6 & 45.9 & 42.6 & -3.3 & 97.4 & 97.5 & +0.1 \\
\midrule
Qwen3.5-0.8B & 1.1 & 0.0 & -1.1 & 0.0 & 0.0 & +0.0 & 0.0 & 0.0 & +0.0 & 84.2 & 92.6 & +8.4 \\
Qwen3.5-2B & 0.0 & 0.0 & +0.0 & 0.0 & 0.0 & +0.0 & 0.0 & 0.0 & +0.0 & 95.5 & 100.0 & +4.5 \\
Qwen3.5-4B & 20.9 & 13.7 & -7.2 & 18.9 & 5.4 & -13.5 & 31.1 & 35.2 & +4.1 & 99.0 & 98.4 & -0.6 \\
Qwen3.5-9B & 18.4 & 9.1 & -9.3 & 62.2 & 10.8 & -51.4 & 49.2 & 35.2 & -13.9 & 98.0 & 97.8 & -0.2 \\
Qwen3.5-27B & 24.6 & 18.6 & -6.1 & 86.5 & 18.9 & -67.6 & 50.0 & 52.5 & +2.5 & 99.2 & 99.1 & -0.1 \\
\midrule
\textbf{Average} & 50.5 & 38.7 & -11.7 & 51.1 & 24.6 & -26.4 & 41.3 & 38.4 & -2.8 & 96.9 & 98.2 & +1.3 \\
\bottomrule
\end{tabular}
\end{threeparttable}
\caption{Effect of Implicit instructions on validation metrics.}
\label{tab:ap_implicit_metrics}
\end{table}

\clearpage

\subsection{Execution Diagnostics}
\begin{table}[H]
\centering

\begin{threeparttable}

\scriptsize
\setlength{\tabcolsep}{4pt}
\begin{tabular}{@{}lccccc@{}}
\toprule
\textbf{Model} & \textbf{State} $\uparrow$ & \textbf{Answer} $\uparrow$ & \textbf{Calls} $\uparrow$ & \textbf{False finish} $\downarrow$ & \textbf{Terminal} $\downarrow$ \\
\midrule
Claude Haiku 4.5 & 72.5 & 57.6 & 99.0 & 12.2 & 0 \\
DeepSeek V4.1 Flash & 96.8 & 87.3 & 98.6 & 18.1 & 0 \\
Gemini 2.5 Flash-Lite & 90.8 & 79.9 & 98.1 & 14.6 & 0 \\
Gemini 3.8 Flash & 96.3 & 93.4 & 99.6 & 7.5 & 0 \\
GLM-5.3 Flash & 96.0 & 79.0 & 93.7 & 21.3 & 0 \\
GPT-5.6 Luna & 95.2 & 84.3 & 97.8 & 24.6 & 0 \\
GPT-5.6 Terra & 97.4 & 90.4 & 99.1 & 16.4 & 0 \\
GPT-5.6 Sol & 98.1 & 90.0 & 99.6 & 19.7 & 0 \\
GPT-6 Luna & 92.7 & 54.5 & 98.2 & 19.8 & 0 \\
GPT-6 Sol & 98.2 & 89.7 & 99.0 & 19.2 & 0 \\
GPT-6.1 Terra & 98.2 & 89.4 & 99.8 & 20.1 & 0 \\
GPT-6.1 Sol & 98.0 & 90.7 & 100.0 & 17.7 & 0 \\
GPT-6 Astra & 98.0 & 90.7 & 100.0 & 17.5 & 0 \\
\midrule
Qwen3.5-0.8B & 26.4 & 0.2 & 92.6 & 0.7 & 94 \\
Qwen3.5-2B & 25.7 & 0.0 & 98.6 & 0.0 & 107 \\
Qwen3.5-4B & 63.3 & 26.0 & 98.6 & 8.7 & 5 \\
Qwen3.5-9B & 69.2 & 35.8 & 98.2 & 2.2 & 1 \\
Qwen3.5-27B & 70.9 & 51.1 & 99.3 & 11.0 & 0 \\
MiniCPM5-1B & 26.1 & 0.0 & 91.8 & 0.0 & 3 \\
MiniCPM5-2B & 39.9 & 5.7 & 93.3 & 0.9 & 1 \\
Gemma-4-E2B-it & 26.3 & 0.1 & 99.5 & 19.8 & 0 \\
Gemma-4-E4B-it & 29.8 & 4.4 & 99.5 & 7.3 & 0 \\
\bottomrule
\end{tabular}
\end{threeparttable}
\caption{Execution diagnostics.}
\label{tab:ap_diagnostics}
\end{table}

\subsection{Complete Test Results}
We present the complete results on test set in the following pages.
\clearpage
\begin{sidewaystable}[p]
\centering

\scriptsize
\setlength{\tabcolsep}{0.3pt}
\renewcommand{\arraystretch}{1}
\begin{tabular}{@{}>{\raggedright\arraybackslash}p{78.4pt}*{22}{>{\centering\arraybackslash}p{\dimexpr(\linewidth-78.4pt-44\tabcolsep)/22\relax}}@{}}
\toprule
\textbf{Metric} & \makecell[c]{Claude\\Haiku\\4.5} & \makecell[c]{Deep\\Seek\\V4.1\\Flash} & \makecell[c]{Gemini\\2.5\\Flash\\Lite} & \makecell[c]{Gemini\\3.8\\Flash} & \makecell[c]{GLM-\\5.3\\Flash} & \makecell[c]{GPT-\\5.6\\Luna} & \makecell[c]{GPT-\\5.6\\Terra} & \makecell[c]{GPT-\\5.6\\Sol} & \makecell[c]{GPT-\\6\\Luna} & \makecell[c]{GPT-\\6\\Sol} & \makecell[c]{GPT-\\6.1\\Terra} & \makecell[c]{GPT-\\6.1\\Sol} & \makecell[c]{GPT-\\6\\Astra} & \makecell[c]{Qwen\\3.5-\\0.8B} & \makecell[c]{Qwen\\3.5-\\2B} & \makecell[c]{Qwen\\3.5-\\4B} & \makecell[c]{Qwen\\3.5-\\9B} & \makecell[c]{Qwen\\3.5-\\27B} & \makecell[c]{Mini\\CPM5-\\1B} & \makecell[c]{Mini\\CPM5-\\2B} & \makecell[c]{Gemma\\4-E2B\\it} & \makecell[c]{Gemma\\4-E4B\\it} \\
\midrule
Pass@1 $\uparrow$ & \cellcolor{blue!10}63.5 & \cellcolor{blue!10}78.3 & \cellcolor{blue!10}72.7 & \cellcolor{blue!10}86.5 & \cellcolor{blue!10}71.3 & \cellcolor{blue!10}72.4 & \cellcolor{blue!10}81.4 & \cellcolor{blue!10}76.9 & \cellcolor{blue!10}56.0 & \cellcolor{blue!10}75.9 & \cellcolor{blue!10}75.6 & \cellcolor{blue!10}78.7 & \cellcolor{blue!10}79.2 & \cellcolor{blue!10}1.1 & \cellcolor{blue!10}0.6 & \cellcolor{blue!10}35.4 & \cellcolor{blue!10}38.1 & \cellcolor{blue!10}53.8 & \cellcolor{blue!10}0.6 & \cellcolor{blue!10}5.4 & \cellcolor{blue!10}0.8 & \cellcolor{blue!10}3.9 \\
State success $\uparrow$ & 85.2 & 96.9 & 90.1 & 95.9 & 95.5 & 95.4 & 97.4 & 98.5 & 92.1 & 98.6 & 99.0 & 98.6 & 98.8 & 26.0 & 25.4 & 60.5 & 65.9 & 68.2 & 25.7 & 33.3 & 26.1 & 27.1 \\
Property acc. $\uparrow$ & 89.3 & 98.9 & 94.5 & 98.0 & 98.4 & 98.2 & 99.3 & 99.6 & 96.4 & 99.5 & 99.6 & 99.5 & 99.6 & 28.9 & 28.4 & 72.7 & 78.3 & 72.7 & 28.5 & 38.8 & 28.9 & 30.9 \\
Device-property acc. $\uparrow$ & 85.7 & 98.6 & 93.4 & 98.6 & 98.5 & 98.9 & 99.2 & 99.6 & 98.2 & 99.7 & 99.8 & 99.6 & 99.7 & 6.0 & 5.4 & 64.2 & 72.0 & 62.0 & 5.5 & 19.9 & 6.1 & 8.7 \\
Space-property acc. $\uparrow$ & 86.6 & 97.7 & 85.1 & 84.1 & 92.2 & 87.2 & 97.6 & 97.5 & 67.6 & 95.8 & 96.5 & 97.3 & 97.0 & 0.5 & 0.0 & 56.1 & 61.0 & 75.4 & 0.0 & 7.1 & 0.0 & 2.8 \\
Action success $\uparrow$ & 98.5 & 98.4 & 97.9 & 99.4 & 95.8 & 96.4 & 98.3 & 99.4 & 97.0 & 98.3 & 99.7 & 100.0 & 100.0 & 94.8 & 98.4 & 98.4 & 97.7 & 99.1 & 95.1 & 97.3 & 99.2 & 99.5 \\
Answer acc. $\uparrow$ & 69.6 & 85.2 & 79.2 & 91.5 & 77.4 & 82.2 & 88.2 & 87.7 & 60.1 & 87.0 & 87.2 & 88.8 & 88.9 & 0.4 & 0.0 & 26.8 & 33.1 & 47.4 & 0.0 & 4.1 & 0.1 & 4.4 \\
List answer recall $\uparrow$ & 72.0 & 88.9 & 81.9 & 93.9 & 81.2 & 90.1 & 93.6 & 98.4 & 78.0 & 98.1 & 99.6 & 99.6 & 99.8 & 0.7 & 0.0 & 32.6 & 29.7 & 38.4 & 0.0 & 6.1 & 0.1 & 3.7 \\
Report acc. $\uparrow$ & 79.2 & 89.2 & 89.2 & 98.5 & 88.5 & 81.5 & 84.6 & 89.1 & 84.1 & 92.4 & 95.4 & 93.6 & 94.7 & 0.0 & 0.0 & 55.1 & 64.6 & 81.3 & 0.0 & 5.4 & 1.7 & 3.8 \\
Report count acc. $\uparrow$ & 82.7 & 94.2 & 92.7 & 98.5 & 90.4 & 87.1 & 88.8 & 91.8 & 90.6 & 97.7 & 98.3 & 97.1 & 98.5 & 0.0 & 0.0 & 58.8 & 69.9 & 85.0 & 0.0 & 5.4 & 4.8 & 8.7 \\
Report semantic acc. $\uparrow$ & 79.2 & 89.2 & 89.2 & 98.5 & 86.2 & 81.3 & 84.6 & 89.1 & 83.8 & 92.4 & 95.4 & 93.6 & 94.7 & 0.0 & 0.0 & 53.3 & 63.8 & 81.3 & 0.0 & 5.2 & 1.0 & 2.8 \\
Semantic success $\uparrow$ & 63.5 & 78.3 & 72.7 & 86.5 & 71.0 & 72.3 & 81.4 & 76.9 & 55.9 & 75.9 & 75.6 & 78.7 & 79.2 & 1.1 & 0.6 & 35.2 & 38.0 & 53.8 & 0.6 & 5.4 & 0.7 & 3.8 \\
Memory success $\uparrow$ & 75.4 & 81.2 & 73.2 & 81.9 & 83.3 & 78.7 & 84.5 & 83.1 & 80.7 & 85.5 & 81.4 & 86.2 & 86.7 & 1.0 & 0.0 & 37.0 & 49.5 & 53.1 & 0.0 & 8.9 & 0.0 & 2.4 \\
Memory ROUGE-L F1 $\uparrow$ & 0.477 & 0.510 & 0.578 & 0.526 & 0.432 & 0.545 & 0.610 & 0.528 & 0.610 & 0.490 & 0.526 & 0.454 & 0.460 & 0.549 & -- & 0.553 & 0.592 & 0.622 & -- & 0.513 & -- & -- \\
Effect replay pass $\uparrow$ & 74.9 & 86.0 & 79.5 & 87.2 & 82.8 & 83.5 & 84.7 & 82.1 & 80.1 & 83.4 & 84.0 & 81.7 & 82.6 & 25.8 & 25.5 & 53.6 & 58.5 & 61.0 & 25.5 & 31.9 & 25.7 & 26.3 \\
Time-plan validity $\uparrow$ & 100.0 & 100.0 & 100.0 & 100.0 & 100.0 & 100.0 & 100.0 & 100.0 & 99.8 & 100.0 & 100.0 & 100.0 & 100.0 & 100.0 & 100.0 & 99.9 & 100.0 & 99.9 & 100.0 & 100.0 & 100.0 & 100.0 \\
False completion $\downarrow$ & 17.0 & 19.5 & 15.6 & 10.1 & 24.4 & 26.4 & 17.5 & 22.1 & 22.3 & 23.4 & 24.0 & 20.4 & 20.1 & 1.0 & 0.0 & 9.1 & 2.3 & 10.1 & 0.0 & 0.5 & 23.6 & 4.0 \\
Unexpected side effects $\downarrow$ & 0.9 & 1.0 & 0.9 & 0.9 & 1.0 & 1.0 & 0.9 & 0.8 & 1.1 & 0.8 & 0.8 & 0.8 & 0.8 & 0.0 & 0.0 & 0.7 & 0.7 & 0.5 & 0.0 & 0.2 & 0.0 & 0.1 \\
Turns $\downarrow$ & 12.4 & 9.1 & 11.8 & 10.3 & 8.0 & 7.0 & 7.5 & 7.6 & 7.0 & 7.8 & 6.3 & 8.0 & 7.9 & 19.4 & 19.7 & 14.7 & 15.4 & 14.5 & 20.0 & 17.8 & 17.0 & 19.0 \\
Tool calls $\downarrow$ & 24.8 & 25.4 & 22.1 & 24.3 & 24.7 & 24.9 & 23.9 & 24.6 & 24.5 & 24.8 & 23.8 & 23.2 & 22.8 & 16.3 & 0.0 & 15.4 & 16.1 & 12.6 & 5.3 & 8.2 & 0.7 & 1.5 \\
Failed calls $\downarrow$ & 0.5 & 0.5 & 0.6 & 0.2 & 1.3 & 1.1 & 0.5 & 0.2 & 0.8 & 0.5 & 0.1 & 0.0 & 0.0 & 1.0 & 0.0 & 0.3 & 0.4 & 0.2 & 1.2 & 0.8 & 0.1 & 0.1 \\
Model calls $\downarrow$ & 12.4 & 9.1 & 11.8 & 10.3 & 8.0 & 7.0 & 7.5 & 7.6 & 7.0 & 7.8 & 6.3 & 8.0 & 7.9 & 19.6 & 19.9 & 14.7 & 15.4 & 14.5 & 20.0 & 17.8 & 17.0 & 19.0 \\
\midrule
Eff-A $\uparrow$ & 0.15 & 0.19 & 0.20 & 0.20 & 0.20 & 0.20 & 0.21 & 0.21 & 0.19 & 0.21 & 0.22 & 0.22 & 0.23 & 0.01 & 0.00 & 0.14 & 0.16 & 0.16 & 0.02 & 0.03 & 0.03 & 0.02 \\
Eff-T $\uparrow$ & 0.39 & 0.53 & 0.41 & 0.49 & 0.61 & 0.73 & 0.68 & 0.70 & 0.66 & 0.63 & 0.81 & 0.60 & 0.60 & 0.02 & 0.01 & 0.17 & 0.19 & 0.18 & 0.01 & 0.06 & 0.02 & 0.03 \\
Infeasible success $\uparrow$ & 73.8 & 86.2 & 87.5 & 97.5 & 86.2 & 78.3 & 82.1 & 85.8 & 79.6 & 88.3 & 92.1 & 91.7 & 92.1 & 0.0 & 0.0 & 60.8 & 72.1 & 80.8 & 0.0 & 5.8 & 1.7 & 6.2 \\
Partial success $\uparrow$ & 86.0 & 94.0 & 92.0 & 100.0 & 92.0 & 82.0 & 88.7 & 94.0 & 46.0 & 98.0 & 100.0 & 96.7 & 98.0 & 0.0 & 0.0 & 24.7 & 47.3 & 82.0 & 0.0 & 0.7 & 0.0 & 0.0 \\
Memory task success $\uparrow$ & 56.5 & 65.2 & 55.1 & 70.3 & 59.4 & 54.6 & 60.4 & 66.2 & 29.5 & 67.6 & 63.5 & 69.3 & 71.0 & 0.0 & 0.0 & 10.4 & 16.2 & 36.5 & 0.0 & 1.0 & 0.0 & 0.7 \\
Memory read $\uparrow$ & 88.1 & 95.2 & 92.9 & 96.4 & 97.6 & 97.2 & 99.6 & 100.0 & 94.0 & 99.2 & 100.0 & 99.6 & 100.0 & 0.0 & 0.0 & 35.3 & 47.6 & 49.6 & 0.0 & 8.7 & 0.0 & 4.0 \\
Memory write $\uparrow$ & 57.1 & 60.7 & 41.1 & 58.9 & 62.5 & 51.8 & 62.5 & 58.3 & 58.9 & 64.9 & 54.2 & 66.7 & 67.3 & 2.4 & 0.0 & 38.1 & 50.0 & 56.0 & 0.0 & 8.3 & 0.0 & 0.0 \\
\bottomrule
\end{tabular}
\caption{Complete Home test results.}
\label{tab:ap_complete_home}
\label{tab:ap_complete_results}
\end{sidewaystable}
\clearpage
\begin{sidewaystable}[p]
\centering

\scriptsize
\setlength{\tabcolsep}{0.3pt}
\renewcommand{\arraystretch}{1}
\begin{tabular}{@{}>{\raggedright\arraybackslash}p{78.4pt}*{22}{>{\centering\arraybackslash}p{\dimexpr(\linewidth-78.4pt-44\tabcolsep)/22\relax}}@{}}
\toprule
\textbf{Metric} & \makecell[c]{Claude\\Haiku\\4.5} & \makecell[c]{Deep\\Seek\\V4.1\\Flash} & \makecell[c]{Gemini\\2.5\\Flash\\Lite} & \makecell[c]{Gemini\\3.8\\Flash} & \makecell[c]{GLM-\\5.3\\Flash} & \makecell[c]{GPT-\\5.6\\Luna} & \makecell[c]{GPT-\\5.6\\Terra} & \makecell[c]{GPT-\\5.6\\Sol} & \makecell[c]{GPT-\\6\\Luna} & \makecell[c]{GPT-\\6\\Sol} & \makecell[c]{GPT-\\6.1\\Terra} & \makecell[c]{GPT-\\6.1\\Sol} & \makecell[c]{GPT-\\6\\Astra} & \makecell[c]{Qwen\\3.5-\\0.8B} & \makecell[c]{Qwen\\3.5-\\2B} & \makecell[c]{Qwen\\3.5-\\4B} & \makecell[c]{Qwen\\3.5-\\9B} & \makecell[c]{Qwen\\3.5-\\27B} & \makecell[c]{Mini\\CPM5-\\1B} & \makecell[c]{Mini\\CPM5-\\2B} & \makecell[c]{Gemma\\4-E2B\\it} & \makecell[c]{Gemma\\4-E4B\\it} \\
\midrule
Pass@1 $\uparrow$ & \cellcolor{blue!10}43.6 & \cellcolor{blue!10}74.1 & \cellcolor{blue!10}69.0 & \cellcolor{blue!10}87.0 & \cellcolor{blue!10}65.8 & \cellcolor{blue!10}70.4 & \cellcolor{blue!10}78.6 & \cellcolor{blue!10}73.0 & \cellcolor{blue!10}57.1 & \cellcolor{blue!10}76.1 & \cellcolor{blue!10}74.3 & \cellcolor{blue!10}76.0 & \cellcolor{blue!10}75.8 & \cellcolor{blue!10}2.5 & \cellcolor{blue!10}2.1 & \cellcolor{blue!10}36.7 & \cellcolor{blue!10}41.2 & \cellcolor{blue!10}51.1 & \cellcolor{blue!10}2.1 & \cellcolor{blue!10}11.1 & \cellcolor{blue!10}2.2 & \cellcolor{blue!10}10.9 \\
State success $\uparrow$ & 67.4 & 95.2 & 88.9 & 95.2 & 94.8 & 94.1 & 96.7 & 96.5 & 92.1 & 96.5 & 96.0 & 95.7 & 95.3 & 32.2 & 31.3 & 68.1 & 72.0 & 71.7 & 31.7 & 46.1 & 31.9 & 37.4 \\
Property acc. $\uparrow$ & 72.1 & 97.4 & 92.2 & 96.5 & 97.3 & 96.8 & 98.3 & 98.2 & 95.8 & 98.3 & 98.1 & 97.8 & 97.5 & 36.5 & 36.0 & 78.7 & 81.9 & 76.0 & 36.0 & 54.0 & 36.2 & 42.0 \\
Device-property acc. $\uparrow$ & 64.4 & 98.8 & 92.5 & 97.7 & 98.9 & 99.4 & 99.7 & 99.2 & 98.6 & 99.2 & 99.1 & 99.1 & 99.2 & 11.8 & 11.3 & 73.2 & 77.9 & 68.1 & 11.3 & 38.4 & 11.4 & 19.2 \\
Space-property acc. $\uparrow$ & 26.1 & 72.5 & 52.9 & 66.7 & 71.0 & 57.7 & 75.8 & 80.9 & 50.2 & 81.4 & 80.9 & 75.8 & 71.7 & 2.4 & 0.0 & 33.8 & 39.1 & 44.4 & 0.0 & 10.9 & 1.0 & 9.9 \\
Action success $\uparrow$ & 99.3 & 98.4 & 98.0 & 99.5 & 91.1 & 98.8 & 99.7 & 99.7 & 99.1 & 99.7 & 99.9 & 100.0 & 100.0 & 91.3 & 98.9 & 98.6 & 98.5 & 99.0 & 88.3 & 92.3 & 99.8 & 99.4 \\
Answer acc. $\uparrow$ & 53.3 & 83.6 & 74.3 & 92.6 & 71.7 & 84.2 & 89.4 & 86.9 & 63.6 & 87.1 & 85.2 & 87.0 & 86.6 & 0.0 & 0.0 & 31.4 & 37.6 & 47.0 & 0.0 & 7.9 & 0.1 & 4.4 \\
List answer recall $\uparrow$ & 53.1 & 86.6 & 76.9 & 94.5 & 75.0 & 91.5 & 93.8 & 98.6 & 76.8 & 98.5 & 98.8 & 98.8 & 98.7 & 0.3 & 0.0 & 35.4 & 35.9 & 42.6 & 0.0 & 10.5 & 0.1 & 0.5 \\
Report acc. $\uparrow$ & 32.7 & 64.4 & 69.3 & 86.6 & 62.4 & 54.1 & 47.9 & 55.0 & 57.4 & 82.0 & 89.3 & 82.3 & 86.6 & 0.1 & 0.0 & 29.9 & 47.5 & 46.5 & 0.0 & 2.3 & 0.8 & 11.2 \\
Report count acc. $\uparrow$ & 32.7 & 71.8 & 74.8 & 90.1 & 69.3 & 66.8 & 59.6 & 66.0 & 75.1 & 85.8 & 92.1 & 91.4 & 92.4 & 0.1 & 0.0 & 50.0 & 62.4 & 70.6 & 0.0 & 2.6 & 4.7 & 34.3 \\
Report semantic acc. $\uparrow$ & 31.7 & 59.4 & 69.3 & 85.6 & 59.4 & 53.1 & 47.2 & 55.0 & 55.1 & 82.0 & 89.3 & 82.3 & 86.6 & 0.1 & 0.0 & 29.5 & 45.9 & 45.2 & 0.0 & 2.0 & 0.0 & 10.6 \\
Semantic success $\uparrow$ & 43.5 & 73.4 & 69.0 & 86.8 & 65.4 & 70.2 & 78.5 & 73.0 & 56.9 & 76.1 & 74.3 & 76.0 & 75.8 & 2.5 & 2.1 & 36.7 & 41.0 & 51.0 & 2.1 & 11.0 & 2.2 & 10.8 \\
Memory success $\uparrow$ & 12.5 & 43.8 & 62.5 & 75.0 & 37.5 & 37.5 & 56.2 & 37.5 & 47.9 & 56.2 & 45.8 & 56.2 & 54.2 & 0.0 & 0.0 & 50.0 & 31.2 & 22.9 & 0.0 & 8.3 & 0.0 & 16.7 \\
Memory ROUGE-L F1 $\uparrow$ & 0.374 & 0.410 & 0.487 & 0.466 & 0.324 & 0.455 & 0.487 & 0.431 & 0.458 & 0.467 & 0.415 & 0.385 & 0.369 & -- & -- & 0.481 & 0.470 & 0.404 & -- & 0.479 & -- & 0.431 \\
Effect replay pass $\uparrow$ & 58.3 & 77.8 & 73.6 & 84.5 & 77.8 & 77.6 & 78.6 & 77.7 & 76.0 & 80.2 & 81.4 & 79.6 & 80.2 & 29.6 & 29.4 & 57.0 & 59.1 & 58.8 & 29.5 & 39.4 & 29.4 & 31.6 \\
Time-plan validity $\uparrow$ & 100.0 & 100.0 & 100.0 & 100.0 & 100.0 & 100.0 & 100.0 & 100.0 & 100.0 & 100.0 & 100.0 & 100.0 & 100.0 & 100.0 & 100.0 & 100.0 & 100.0 & 100.0 & 100.0 & 100.0 & 100.0 & 100.0 \\
False completion $\downarrow$ & 10.3 & 22.5 & 16.5 & 6.9 & 24.5 & 28.2 & 19.9 & 25.0 & 18.8 & 23.3 & 24.9 & 23.5 & 23.5 & 0.7 & 0.0 & 9.5 & 2.3 & 14.7 & 0.0 & 0.8 & 17.3 & 11.5 \\
Unexpected side effects $\downarrow$ & 0.6 & 1.2 & 0.6 & 0.6 & 0.6 & 0.9 & 0.9 & 1.0 & 0.7 & 1.0 & 1.0 & 1.1 & 1.0 & 0.0 & 0.0 & 0.4 & 0.5 & 0.4 & 0.0 & 0.3 & 0.0 & 0.1 \\
Turns $\downarrow$ & 14.6 & 9.5 & 12.9 & 10.9 & 9.0 & 7.6 & 7.6 & 7.7 & 7.0 & 7.7 & 6.2 & 7.2 & 7.2 & 19.6 & 19.8 & 14.8 & 15.3 & 14.7 & 20.0 & 17.4 & 17.8 & 17.5 \\
Tool calls $\downarrow$ & 17.5 & 25.8 & 21.3 & 22.5 & 25.5 & 22.2 & 21.9 & 22.9 & 19.4 & 21.9 & 20.7 & 19.6 & 18.9 & 20.5 & 0.0 & 15.6 & 16.5 & 13.1 & 3.1 & 13.2 & 0.4 & 1.9 \\
Failed calls $\downarrow$ & 0.2 & 0.6 & 0.6 & 0.2 & 2.8 & 0.4 & 0.1 & 0.1 & 0.3 & 0.1 & 0.0 & 0.0 & 0.0 & 2.8 & 0.0 & 0.3 & 0.3 & 0.2 & 2.3 & 2.6 & 0.0 & 0.1 \\
Model calls $\downarrow$ & 14.6 & 9.5 & 12.9 & 10.9 & 9.0 & 7.6 & 7.6 & 7.7 & 7.0 & 7.7 & 6.2 & 7.2 & 7.2 & 19.7 & 20.0 & 14.8 & 15.3 & 14.7 & 20.0 & 17.4 & 17.8 & 17.5 \\
\midrule
Eff-A $\uparrow$ & 0.12 & 0.17 & 0.19 & 0.20 & 0.17 & 0.20 & 0.21 & 0.21 & 0.21 & 0.21 & 0.22 & 0.24 & 0.24 & 0.02 & 0.00 & 0.15 & 0.17 & 0.17 & 0.01 & 0.05 & 0.03 & 0.05 \\
Eff-T $\uparrow$ & 0.26 & 0.48 & 0.33 & 0.43 & 0.51 & 0.60 & 0.62 & 0.61 & 0.60 & 0.58 & 0.72 & 0.61 & 0.61 & 0.02 & 0.02 & 0.17 & 0.20 & 0.18 & 0.02 & 0.11 & 0.03 & 0.05 \\
Infeasible success $\uparrow$ & 24.3 & 64.3 & 68.6 & 84.3 & 62.9 & 53.3 & 42.4 & 51.4 & 55.7 & 80.0 & 85.2 & 79.0 & 82.9 & 0.0 & 0.0 & 32.4 & 48.6 & 43.3 & 0.0 & 1.4 & 0.5 & 16.2 \\
Partial success $\uparrow$ & 51.6 & 64.5 & 71.0 & 90.3 & 61.3 & 55.9 & 60.2 & 61.3 & 32.3 & 84.9 & 96.8 & 88.2 & 93.5 & 0.0 & 0.0 & 15.1 & 43.0 & 53.8 & 0.0 & 2.2 & 0.0 & 0.0 \\
Memory task success $\uparrow$ & 12.5 & 25.0 & 56.2 & 62.5 & 31.2 & 33.3 & 37.5 & 27.1 & 25.0 & 45.8 & 37.5 & 47.9 & 50.0 & 0.0 & 0.0 & 8.3 & 16.7 & 12.5 & 0.0 & 0.0 & 0.0 & 2.1 \\
Memory read $\uparrow$ & 0.0 & 80.0 & 80.0 & 80.0 & 80.0 & 26.7 & 80.0 & 80.0 & 6.7 & 80.0 & 86.7 & 80.0 & 86.7 & 0.0 & 0.0 & 0.0 & 0.0 & 20.0 & 0.0 & 0.0 & 0.0 & 0.0 \\
Memory write $\uparrow$ & 18.2 & 18.2 & 54.5 & 72.7 & 9.1 & 36.4 & 36.4 & 12.1 & 57.6 & 36.4 & 21.2 & 36.4 & 33.3 & 0.0 & 0.0 & 72.7 & 39.4 & 15.2 & 0.0 & 12.1 & 0.0 & 24.2 \\
\bottomrule
\end{tabular}
\caption{Complete Car test results.}
\label{tab:ap_complete_car}
\end{sidewaystable}
\clearpage
\begin{sidewaystable}[p]
\centering

\scriptsize
\setlength{\tabcolsep}{0.3pt}
\renewcommand{\arraystretch}{1}
\begin{tabular}{@{}>{\raggedright\arraybackslash}p{78.4pt}*{22}{>{\centering\arraybackslash}p{\dimexpr(\linewidth-78.4pt-44\tabcolsep)/22\relax}}@{}}
\toprule
\textbf{Metric} & \makecell[c]{Claude\\Haiku\\4.5} & \makecell[c]{Deep\\Seek\\V4.1\\Flash} & \makecell[c]{Gemini\\2.5\\Flash\\Lite} & \makecell[c]{Gemini\\3.8\\Flash} & \makecell[c]{GLM-\\5.3\\Flash} & \makecell[c]{GPT-\\5.6\\Luna} & \makecell[c]{GPT-\\5.6\\Terra} & \makecell[c]{GPT-\\5.6\\Sol} & \makecell[c]{GPT-\\6\\Luna} & \makecell[c]{GPT-\\6\\Sol} & \makecell[c]{GPT-\\6.1\\Terra} & \makecell[c]{GPT-\\6.1\\Sol} & \makecell[c]{GPT-\\6\\Astra} & \makecell[c]{Qwen\\3.5-\\0.8B} & \makecell[c]{Qwen\\3.5-\\2B} & \makecell[c]{Qwen\\3.5-\\4B} & \makecell[c]{Qwen\\3.5-\\9B} & \makecell[c]{Qwen\\3.5-\\27B} & \makecell[c]{Mini\\CPM5-\\1B} & \makecell[c]{Mini\\CPM5-\\2B} & \makecell[c]{Gemma\\4-E2B\\it} & \makecell[c]{Gemma\\4-E4B\\it} \\
\midrule
Pass@1 $\uparrow$ & \cellcolor{blue!10}33.7 & \cellcolor{blue!10}91.4 & \cellcolor{blue!10}85.3 & \cellcolor{blue!10}97.3 & \cellcolor{blue!10}87.9 & \cellcolor{blue!10}84.1 & \cellcolor{blue!10}91.7 & \cellcolor{blue!10}93.7 & \cellcolor{blue!10}48.3 & \cellcolor{blue!10}96.9 & \cellcolor{blue!10}97.1 & \cellcolor{blue!10}98.0 & \cellcolor{blue!10}98.2 & \cellcolor{blue!10}3.7 & \cellcolor{blue!10}3.5 & \cellcolor{blue!10}29.7 & \cellcolor{blue!10}47.2 & \cellcolor{blue!10}65.0 & \cellcolor{blue!10}3.5 & \cellcolor{blue!10}13.5 & \cellcolor{blue!10}3.1 & \cellcolor{blue!10}11.9 \\
State success $\uparrow$ & 48.3 & 99.0 & 95.5 & 99.2 & 99.2 & 96.6 & 98.6 & 99.3 & 95.2 & 99.6 & 99.4 & 99.8 & 99.8 & 18.7 & 17.9 & 63.0 & 73.3 & 76.5 & 18.4 & 47.4 & 18.4 & 25.1 \\
Property acc. $\uparrow$ & 54.2 & 99.8 & 98.1 & 99.7 & 99.8 & 99.1 & 99.7 & 99.9 & 98.5 & 99.9 & 99.9 & 100.0 & 100.0 & 23.7 & 23.4 & 76.3 & 85.7 & 81.8 & 23.5 & 56.5 & 23.5 & 31.1 \\
Device-property acc. $\uparrow$ & 46.2 & 99.8 & 97.7 & 99.7 & 99.8 & 99.0 & 99.7 & 99.8 & 98.3 & 99.9 & 99.8 & 99.9 & 99.9 & 10.4 & 10.1 & 72.2 & 83.2 & 78.6 & 10.1 & 48.9 & 10.1 & 19.0 \\
Space-property acc. $\uparrow$ & -- & -- & -- & -- & -- & -- & -- & -- & -- & -- & -- & -- & -- & -- & -- & -- & -- & -- & -- & -- & -- & -- \\
Action success $\uparrow$ & 99.9 & 99.2 & 98.7 & 99.9 & 92.4 & 99.9 & 99.9 & 99.9 & 99.9 & 100.0 & 100.0 & 100.0 & 100.0 & 89.1 & 98.4 & 99.0 & 99.0 & 100.0 & 88.8 & 84.8 & 99.7 & 99.6 \\
Answer acc. $\uparrow$ & 34.8 & 96.7 & 88.5 & 98.7 & 92.0 & 89.1 & 96.5 & 99.4 & 30.0 & 99.6 & 99.9 & 99.9 & 99.9 & 0.0 & 0.0 & 17.3 & 39.9 & 64.9 & 0.0 & 6.5 & 0.0 & 4.1 \\
List answer recall $\uparrow$ & -- & -- & -- & -- & -- & -- & -- & -- & -- & -- & -- & -- & -- & -- & -- & -- & -- & -- & -- & -- & -- & -- \\
Report acc. $\uparrow$ & 5.0 & 75.0 & 80.0 & 97.5 & 67.5 & 60.8 & 65.0 & 72.5 & 62.4 & 91.7 & 100.0 & 100.0 & 100.0 & 1.9 & 0.0 & 26.7 & 40.4 & 68.3 & 0.0 & 3.9 & 0.4 & 21.7 \\
Report count acc. $\uparrow$ & 5.0 & 75.0 & 80.0 & 97.5 & 70.0 & 61.7 & 65.8 & 74.2 & 64.0 & 91.7 & 100.0 & 100.0 & 100.0 & 1.9 & 0.0 & 26.7 & 40.4 & 68.3 & 0.0 & 3.9 & 5.1 & 24.2 \\
Report semantic acc. $\uparrow$ & 5.0 & 75.0 & 80.0 & 97.5 & 67.5 & 60.8 & 65.0 & 72.5 & 62.4 & 91.7 & 100.0 & 100.0 & 100.0 & 1.7 & 0.0 & 25.8 & 40.4 & 68.3 & 0.0 & 3.1 & 0.0 & 21.7 \\
Semantic success $\uparrow$ & 33.7 & 91.4 & 85.3 & 97.3 & 87.9 & 84.1 & 91.7 & 93.7 & 48.3 & 96.9 & 97.1 & 98.0 & 98.2 & 3.7 & 3.5 & 29.7 & 47.2 & 65.0 & 3.5 & 13.4 & 3.1 & 11.9 \\
Memory success $\uparrow$ & 15.2 & 82.6 & 87.0 & 95.7 & 80.4 & 59.4 & 80.4 & 67.4 & 50.0 & 84.8 & 74.6 & 83.3 & 84.1 & 0.0 & 0.0 & 1.4 & 15.9 & 42.8 & 0.0 & 8.0 & 0.0 & 12.3 \\
Memory ROUGE-L F1 $\uparrow$ & 0.292 & 0.525 & 0.551 & 0.554 & 0.484 & 0.527 & 0.510 & 0.485 & 0.477 & 0.452 & 0.459 & 0.448 & 0.412 & -- & -- & 0.429 & 0.487 & 0.492 & -- & 0.615 & -- & 0.478 \\
Effect replay pass $\uparrow$ & 48.6 & 96.3 & 93.8 & 98.4 & 96.5 & 95.5 & 96.4 & 97.5 & 94.9 & 98.6 & 99.4 & 99.1 & 99.5 & 19.4 & 19.1 & 62.3 & 73.3 & 76.2 & 19.3 & 47.4 & 19.3 & 24.7 \\
Time-plan validity $\uparrow$ & 100.0 & 100.0 & 100.0 & 100.0 & 100.0 & 100.0 & 100.0 & 100.0 & 100.0 & 100.0 & 100.0 & 100.0 & 100.0 & 100.0 & 100.0 & 100.0 & 100.0 & 100.0 & 100.0 & 100.0 & 100.0 & 100.0 \\
False completion $\downarrow$ & 2.9 & 7.8 & 9.0 & 2.0 & 8.6 & 14.8 & 8.1 & 5.7 & 14.9 & 2.9 & 2.9 & 2.0 & 1.8 & 0.1 & 0.0 & 6.7 & 2.2 & 7.8 & 0.0 & 2.3 & 14.3 & 9.3 \\
Unexpected side effects $\downarrow$ & 0.0 & 0.1 & 0.0 & 0.0 & 0.1 & 0.1 & 0.1 & 0.1 & 0.1 & 0.0 & 0.0 & 0.0 & 0.0 & 0.0 & 0.0 & 0.1 & 0.1 & 0.1 & 0.0 & 0.1 & 0.0 & 0.0 \\
Turns $\downarrow$ & 16.1 & 8.6 & 11.6 & 9.2 & 8.2 & 7.8 & 7.7 & 7.5 & 6.8 & 7.8 & 6.7 & 7.3 & 7.2 & 19.4 & 19.7 & 17.0 & 15.7 & 14.8 & 20.0 & 16.2 & 18.0 & 17.8 \\
Tool calls $\downarrow$ & 11.4 & 21.4 & 16.2 & 17.4 & 20.0 & 17.3 & 17.0 & 17.1 & 15.5 & 17.3 & 16.8 & 15.7 & 15.7 & 17.5 & 0.0 & 11.9 & 13.0 & 11.5 & 3.1 & 15.8 & 0.3 & 2.1 \\
Failed calls $\downarrow$ & 0.0 & 0.3 & 0.3 & 0.0 & 2.0 & 0.0 & 0.0 & 0.0 & 0.0 & 0.0 & 0.0 & 0.0 & 0.0 & 2.6 & 0.0 & 0.2 & 0.1 & 0.0 & 1.9 & 5.3 & 0.0 & 0.0 \\
Model calls $\downarrow$ & 16.1 & 8.6 & 11.6 & 9.2 & 8.2 & 7.8 & 7.7 & 7.5 & 6.8 & 7.8 & 6.7 & 7.3 & 7.2 & 19.5 & 19.9 & 17.0 & 15.7 & 14.8 & 20.0 & 16.2 & 18.0 & 17.8 \\
\midrule
Eff-A $\uparrow$ & 0.12 & 0.25 & 0.31 & 0.30 & 0.27 & 0.29 & 0.30 & 0.30 & 0.27 & 0.30 & 0.32 & 0.34 & 0.34 & 0.09 & 0.00 & 0.19 & 0.25 & 0.26 & 0.01 & 0.09 & 0.02 & 0.11 \\
Eff-T $\uparrow$ & 0.25 & 0.63 & 0.46 & 0.59 & 0.68 & 0.69 & 0.70 & 0.73 & 0.63 & 0.69 & 0.82 & 0.74 & 0.75 & 0.06 & 0.02 & 0.18 & 0.24 & 0.26 & 0.02 & 0.19 & 0.03 & 0.06 \\
Infeasible success $\uparrow$ & 5.0 & 70.0 & 80.0 & 100.0 & 70.0 & 63.3 & 63.3 & 66.7 & 68.3 & 90.0 & 100.0 & 100.0 & 100.0 & 3.3 & 0.0 & 30.0 & 56.7 & 76.7 & 0.0 & 3.3 & 0.0 & 38.3 \\
Partial success $\uparrow$ & 5.0 & 80.0 & 80.0 & 95.0 & 60.0 & 58.3 & 66.7 & 78.3 & 36.7 & 93.3 & 100.0 & 100.0 & 100.0 & 0.0 & 0.0 & 10.0 & 23.3 & 60.0 & 0.0 & 1.7 & 0.0 & 3.3 \\
Memory task success $\uparrow$ & 13.0 & 80.4 & 65.2 & 95.7 & 71.7 & 47.8 & 69.6 & 67.4 & 10.9 & 84.8 & 74.6 & 83.3 & 84.1 & 0.0 & 0.0 & 0.0 & 2.9 & 37.7 & 0.0 & 1.4 & 0.0 & 5.8 \\
Memory read $\uparrow$ & 20.0 & 92.0 & 88.0 & 100.0 & 96.0 & 49.3 & 81.3 & 90.7 & 30.7 & 96.0 & 96.0 & 100.0 & 100.0 & 0.0 & 0.0 & 0.0 & 1.3 & 28.0 & 0.0 & 2.7 & 0.0 & 1.3 \\
Memory write $\uparrow$ & 9.5 & 71.4 & 85.7 & 90.5 & 61.9 & 71.4 & 79.4 & 39.7 & 73.0 & 71.4 & 49.2 & 63.5 & 65.1 & 0.0 & 0.0 & 3.2 & 33.3 & 60.3 & 0.0 & 14.3 & 0.0 & 25.4 \\
\bottomrule
\end{tabular}
\caption{Complete Phone test results.}
\label{tab:ap_complete_phone}
\end{sidewaystable}
\clearpage
\begin{sidewaystable}[p]
\centering

\scriptsize
\setlength{\tabcolsep}{0.3pt}
\renewcommand{\arraystretch}{1}
\begin{tabular}{@{}>{\raggedright\arraybackslash}p{78.4pt}*{22}{>{\centering\arraybackslash}p{\dimexpr(\linewidth-78.4pt-44\tabcolsep)/22\relax}}@{}}
\toprule
\textbf{Metric} & \makecell[c]{Claude\\Haiku\\4.5} & \makecell[c]{Deep\\Seek\\V4.1\\Flash} & \makecell[c]{Gemini\\2.5\\Flash\\Lite} & \makecell[c]{Gemini\\3.8\\Flash} & \makecell[c]{GLM-\\5.3\\Flash} & \makecell[c]{GPT-\\5.6\\Luna} & \makecell[c]{GPT-\\5.6\\Terra} & \makecell[c]{GPT-\\5.6\\Sol} & \makecell[c]{GPT-\\6\\Luna} & \makecell[c]{GPT-\\6\\Sol} & \makecell[c]{GPT-\\6.1\\Terra} & \makecell[c]{GPT-\\6.1\\Sol} & \makecell[c]{GPT-\\6\\Astra} & \makecell[c]{Qwen\\3.5-\\0.8B} & \makecell[c]{Qwen\\3.5-\\2B} & \makecell[c]{Qwen\\3.5-\\4B} & \makecell[c]{Qwen\\3.5-\\9B} & \makecell[c]{Qwen\\3.5-\\27B} & \makecell[c]{Mini\\CPM5-\\1B} & \makecell[c]{Mini\\CPM5-\\2B} & \makecell[c]{Gemma\\4-E2B\\it} & \makecell[c]{Gemma\\4-E4B\\it} \\
\midrule
Pass@1 $\uparrow$ & \cellcolor{blue!10}51.6 & \cellcolor{blue!10}79.6 & \cellcolor{blue!10}74.1 & \cellcolor{blue!10}88.8 & \cellcolor{blue!10}72.9 & \cellcolor{blue!10}74.1 & \cellcolor{blue!10}82.6 & \cellcolor{blue!10}79.1 & \cellcolor{blue!10}54.8 & \cellcolor{blue!10}80.2 & \cellcolor{blue!10}79.5 & \cellcolor{blue!10}81.7 & \cellcolor{blue!10}82.0 & \cellcolor{blue!10}2.0 & \cellcolor{blue!10}1.6 & \cellcolor{blue!10}34.7 & \cellcolor{blue!10}40.8 & \cellcolor{blue!10}55.3 & \cellcolor{blue!10}1.6 & \cellcolor{blue!10}8.7 & \cellcolor{blue!10}1.7 & \cellcolor{blue!10}7.6 \\
State success $\uparrow$ & 72.5 & 96.8 & 90.8 & 96.3 & 96.0 & 95.2 & 97.4 & 98.1 & 92.7 & 98.2 & 98.2 & 98.0 & 98.0 & 26.4 & 25.7 & 63.3 & 69.2 & 70.9 & 26.1 & 39.9 & 26.3 & 29.8 \\
Property acc. $\uparrow$ & 77.1 & 98.6 & 94.5 & 97.9 & 98.4 & 98.0 & 99.1 & 99.2 & 96.6 & 99.2 & 99.2 & 99.1 & 99.0 & 30.1 & 29.7 & 75.2 & 80.8 & 75.5 & 29.7 & 46.9 & 30.0 & 34.3 \\
Device-property acc. $\uparrow$ & 70.3 & 98.9 & 94.2 & 98.6 & 98.9 & 99.0 & 99.5 & 99.6 & 98.3 & 99.6 & 99.6 & 99.5 & 99.6 & 8.7 & 8.2 & 68.6 & 76.3 & 67.7 & 8.2 & 32.0 & 8.5 & 14.1 \\
Space-property acc. $\uparrow$ & 63.4 & 88.1 & 72.8 & 77.4 & 84.1 & 75.9 & 89.3 & 91.1 & 60.9 & 90.3 & 90.6 & 89.1 & 87.3 & 1.2 & 0.0 & 47.6 & 52.6 & 63.5 & 0.0 & 8.5 & 0.4 & 5.5 \\
Action success $\uparrow$ & 99.0 & 98.6 & 98.1 & 99.6 & 93.7 & 97.8 & 99.1 & 99.6 & 98.2 & 99.0 & 99.8 & 100.0 & 100.0 & 92.6 & 98.6 & 98.6 & 98.2 & 99.3 & 91.8 & 93.3 & 99.5 & 99.5 \\
Answer acc. $\uparrow$ & 57.6 & 87.3 & 79.9 & 93.4 & 79.0 & 84.3 & 90.4 & 90.0 & 54.5 & 89.7 & 89.4 & 90.7 & 90.7 & 0.2 & 0.0 & 26.0 & 35.8 & 51.1 & 0.0 & 5.7 & 0.1 & 4.4 \\
List answer recall $\uparrow$ & 65.4 & 88.1 & 80.1 & 94.1 & 79.0 & 90.6 & 93.7 & 98.5 & 77.6 & 98.2 & 99.3 & 99.3 & 99.4 & 0.5 & 0.0 & 33.6 & 31.9 & 39.9 & 0.0 & 7.7 & 0.1 & 2.6 \\
Report acc. $\uparrow$ & 50.9 & 77.9 & 80.4 & 93.9 & 75.6 & 68.3 & 68.0 & 73.9 & 71.0 & 88.4 & 93.8 & 90.3 & 92.5 & 0.3 & 0.0 & 41.5 & 54.7 & 66.4 & 0.0 & 4.0 & 1.2 & 9.2 \\
Report count acc. $\uparrow$ & 52.6 & 83.0 & 84.1 & 95.2 & 79.5 & 75.8 & 74.5 & 79.6 & 80.9 & 92.4 & 96.2 & 95.4 & 96.4 & 0.3 & 0.0 & 50.8 & 62.7 & 77.2 & 0.0 & 4.2 & 4.8 & 20.5 \\
Report semantic acc. $\uparrow$ & 50.6 & 76.0 & 80.4 & 93.5 & 73.4 & 67.8 & 67.8 & 73.9 & 70.0 & 88.4 & 93.8 & 90.3 & 92.5 & 0.3 & 0.0 & 40.4 & 53.7 & 65.9 & 0.0 & 3.7 & 0.5 & 8.5 \\
Semantic success $\uparrow$ & 51.6 & 79.5 & 74.1 & 88.8 & 72.7 & 74.1 & 82.6 & 79.1 & 54.7 & 80.2 & 79.5 & 81.7 & 82.0 & 2.0 & 1.6 & 34.6 & 40.8 & 55.2 & 1.6 & 8.7 & 1.6 & 7.5 \\
Memory success $\uparrow$ & 56.5 & 78.5 & 75.5 & 84.5 & 79.0 & 71.0 & 81.3 & 75.8 & 71.0 & 83.0 & 77.0 & 83.2 & 83.5 & 0.7 & 0.0 & 29.8 & 40.3 & 48.3 & 0.0 & 8.7 & 0.0 & 5.8 \\
Memory ROUGE-L F1 $\uparrow$ & 0.447 & 0.498 & 0.557 & 0.525 & 0.428 & 0.526 & 0.564 & 0.502 & 0.552 & 0.476 & 0.489 & 0.442 & 0.434 & 0.549 & -- & 0.528 & 0.551 & 0.560 & -- & 0.532 & -- & 0.457 \\
Effect replay pass $\uparrow$ & 64.6 & 85.6 & 80.6 & 88.6 & 84.0 & 84.1 & 85.2 & 83.9 & 81.8 & 85.5 & 86.3 & 84.6 & 85.3 & 25.7 & 25.4 & 56.4 & 61.7 & 63.4 & 25.5 & 37.3 & 25.6 & 27.6 \\
Time-plan validity $\uparrow$ & 100.0 & 100.0 & 100.0 & 100.0 & 100.0 & 100.0 & 100.0 & 100.0 & 99.9 & 100.0 & 100.0 & 100.0 & 100.0 & 100.0 & 100.0 & 100.0 & 100.0 & 100.0 & 100.0 & 100.0 & 100.0 & 100.0 \\
False completion $\downarrow$ & 12.2 & 18.1 & 14.6 & 7.5 & 21.3 & 24.6 & 16.4 & 19.7 & 19.8 & 19.2 & 20.1 & 17.7 & 17.5 & 0.7 & 0.0 & 8.7 & 2.2 & 11.0 & 0.0 & 0.9 & 19.8 & 7.3 \\
Unexpected side effects $\downarrow$ & 0.6 & 0.9 & 0.6 & 0.6 & 0.7 & 0.8 & 0.7 & 0.7 & 0.8 & 0.7 & 0.7 & 0.7 & 0.7 & 0.0 & 0.0 & 0.5 & 0.5 & 0.4 & 0.0 & 0.2 & 0.0 & 0.1 \\
Turns $\downarrow$ & 13.8 & 9.1 & 12.1 & 10.3 & 8.4 & 7.3 & 7.6 & 7.6 & 7.0 & 7.7 & 6.4 & 7.6 & 7.6 & 19.4 & 19.7 & 15.2 & 15.4 & 14.6 & 20.0 & 17.4 & 17.4 & 18.3 \\
Tool calls $\downarrow$ & 19.9 & 24.7 & 20.7 & 22.4 & 24.0 & 22.6 & 21.9 & 22.6 & 21.1 & 22.4 & 21.5 & 20.6 & 20.2 & 17.8 & 0.0 & 14.8 & 15.6 & 12.5 & 4.2 & 11.2 & 0.5 & 1.7 \\
Failed calls $\downarrow$ & 0.3 & 0.5 & 0.5 & 0.1 & 1.9 & 0.7 & 0.3 & 0.1 & 0.5 & 0.3 & 0.0 & 0.0 & 0.0 & 1.9 & 0.0 & 0.3 & 0.3 & 0.1 & 1.7 & 2.2 & 0.0 & 0.1 \\
Model calls $\downarrow$ & 13.8 & 9.1 & 12.1 & 10.3 & 8.4 & 7.3 & 7.6 & 7.6 & 7.0 & 7.7 & 6.4 & 7.6 & 7.6 & 19.6 & 19.9 & 15.2 & 15.4 & 14.6 & 20.0 & 17.4 & 17.4 & 18.3 \\
\midrule
Eff-A $\uparrow$ & 0.13 & 0.19 & 0.22 & 0.22 & 0.21 & 0.22 & 0.23 & 0.23 & 0.21 & 0.23 & 0.24 & 0.25 & 0.26 & 0.03 & 0.00 & 0.15 & 0.18 & 0.18 & 0.01 & 0.05 & 0.03 & 0.05 \\
Eff-T $\uparrow$ & 0.32 & 0.54 & 0.40 & 0.50 & 0.59 & 0.69 & 0.66 & 0.68 & 0.64 & 0.63 & 0.78 & 0.63 & 0.63 & 0.03 & 0.02 & 0.17 & 0.20 & 0.20 & 0.02 & 0.10 & 0.02 & 0.04 \\
Infeasible success $\uparrow$ & 45.3 & 75.3 & 78.8 & 92.4 & 74.7 & 66.3 & 63.5 & 69.4 & 68.4 & 85.1 & 90.2 & 87.5 & 89.2 & 0.4 & 0.0 & 45.5 & 60.6 & 64.9 & 0.0 & 3.7 & 1.0 & 14.1 \\
Partial success $\uparrow$ & 59.4 & 82.2 & 83.2 & 96.0 & 76.2 & 69.3 & 75.6 & 80.9 & 39.9 & 93.1 & 99.0 & 94.7 & 97.0 & 0.0 & 0.0 & 18.8 & 41.3 & 69.0 & 0.0 & 1.3 & 0.0 & 0.7 \\
Memory task success $\uparrow$ & 43.0 & 65.5 & 57.5 & 75.5 & 60.0 & 51.3 & 60.7 & 63.3 & 24.8 & 69.8 & 64.0 & 70.8 & 72.3 & 0.0 & 0.0 & 7.8 & 13.2 & 34.8 & 0.0 & 1.0 & 0.0 & 2.0 \\
Memory read $\uparrow$ & 69.3 & 93.9 & 91.2 & 96.5 & 96.5 & 83.6 & 94.7 & 97.1 & 76.3 & 97.7 & 98.5 & 98.8 & 99.4 & 0.0 & 0.0 & 26.0 & 35.4 & 43.6 & 0.0 & 7.0 & 0.0 & 3.2 \\
Memory write $\uparrow$ & 40.9 & 58.0 & 53.4 & 68.2 & 55.7 & 54.5 & 63.3 & 48.1 & 62.1 & 62.9 & 48.9 & 62.1 & 62.5 & 1.5 & 0.0 & 34.1 & 44.7 & 51.9 & 0.0 & 10.2 & 0.0 & 9.1 \\
\bottomrule
\end{tabular}
\caption{Complete overall test results.}
\label{tab:ap_complete_overall}
\end{sidewaystable}

\clearpage
\section{Error Analysis}
\label[appendix]{app:failure-method}

\subsection{Failure Taxonomy and Attribution}
We analyze all 27,252 failed first attempts across 22 models and 2,560 test tasks. \Cref{tab:failure-taxonomy} groups twelve failure mechanisms into four categories. GPT-6 Luna assigns ordered causes from each task's instruction, observations, tool trace, and evaluation results. GPT-6.1 Sol then reviews all 399 cases initially assigned E12, retaining E12 in 212 cases.

\para{Analysis unit and evidence.}
The attribution set contains every accepted failure from the first attempt of each model-task pair, drawn from 56,320 attempts across 66 model--domain evaluations. Each case preserves the public instruction and interaction protocol, agent-visible observations, submitted actions, execution trace, private evaluation targets, and final evaluation report. Accepted reruns supply the replacement trajectories; accepted rescoring supplies the updated evaluation reports. The judge returns an ordered list of contributing mechanisms and an explanation grounded in this evidence. Sol's review replaces the attribution of the 399 E12 candidates using the same case context.

\begin{table}[H]
\centering

\scriptsize
\setlength{\tabcolsep}{1pt}
\renewcommand{\arraystretch}{1}
\begin{tabular}{@{}>{\centering\arraybackslash}m{0.1\linewidth}>{\raggedright\arraybackslash}m{0.23\linewidth}>{\raggedright\arraybackslash}m{0.66\linewidth}@{}}
\toprule
\textbf{Group} & \textbf{Failure category} & \textbf{Failure mechanism} \\
\midrule
\multirow{3}{*}{\makecell[c]{Basic\\interaction}}
& Information extraction (E1) & Misreading observations or binding values to the wrong object or user. \\
& Protocol adherence (E2) & Malformed interaction structure or failure to submit the required answer. \\
& Tool-call construction (E3) & Incorrect interfaces, argument names, types, or submission structures. \\
\midrule
\multirow{4}{*}{\makecell[c]{Complex\\reasoning}}
& Math \& logic reasoning (E4) & Incorrect calculation, counting, comparison, ranking, or deduction. \\
& Constraint reasoning (E5) & Missing or misapplying conditions, exclusions, or joint constraints. \\
& Task-wise coordination (E6) & Omitting a subtask or misunderstanding dependencies across subtasks. \\
& Subtask planning (E7) & Choosing incorrect actions or missing prerequisite steps within a subtask. \\
\midrule
\multirow{3}{*}{\makecell[c]{Integrated\\reasoning}}
& Environment \& state (E8) & Failing to identify information gaps or combine observations into a consistent state. \\
& Feasibility \& conflicts (E9) & Misjudging reachability, resource limits, or incompatible requirements. \\
& Dynamics \& timing (E10) & Incorrect monitoring, durations, scheduling, synchronization, or responses to state changes. \\
\midrule
\multirow{2}{*}{\makecell[c]{External\\errors}}
& Runtime \& service (E11) & Simulator, parser, tool, or model-service faults interrupting valid execution. \\
& Task \& evaluation (E12) & Inconsistent instructions, targets, references, scoring rules, or judge decisions. \\
\bottomrule
\end{tabular}
\caption{Four groups of failure mechanisms and twelve attribution labels.}
\label{tab:failure-taxonomy}
\end{table}

Each failed attempt contributes total weight one. For $K_i$ ordered causes, cause $r$ receives
\[
 w_{i,r}=\frac{2^{-(r-1)}}{\sum_{j=1}^{K_i}2^{-(j-1)}}.
\]
Model bars in \Cref{fig:errors_combined} show weighted failure counts; task-family and domain bars normalize these counts within failures. Task success and failure attribution are scored separately.

For example, two ordered causes receive weights $2/3$ and $1/3$. Each model bar therefore sums to its number of failed first attempts, while each task-family or domain bar sums to 100\%. A case can contribute to several categories, but its total contribution remains one. The 212 retained E12 cases are an incidence count: each case bearing E12 is counted once, regardless of its position in the ordered list. Since all accepted first-attempt failures are included, these aggregates require no sampling or inverse-inclusion correction.

\subsection{Human Annotation Calibration}
Three annotators independently label the same 200 cases, yielding 600 annotations.
Each annotator assigns labels using the same definitions, with optional feedback and an ``other'' category. GPT-6 Luna and GPT-6.1 Sol are compared with the humans on these same cases.

For label sets $L_{a,i}$ and $L_{b,i}$, agreement is the mean per-case Jaccard, including ``other'' labels:
\[
 J(a,b)=\frac{1}{200}\sum_{i=1}^{200}\frac{|L_{a,i}\cap L_{b,i}|}{|L_{a,i}\cup L_{b,i}|}.
\]
The mean across human pairs is 65.2\%; GPT-6 Luna and GPT-6.1 Sol average 70.9\% and 69.1\% against the three annotators. \Cref{fig:errors_combined} includes the pairwise agreement matrix.

\subsection{Representative Failure Traces}
\label[appendix]{app:failure-cases}

\newcounter{failurecase}
\refstepcounter{failurecase}
\para{Case \thefailurecase: correct observation, incorrect answer type (E3).}
\label{case:answer-type}
Qwen3.5-0.8B, Home, \texttt{task\_cb215ba7}.
The instruction asks, ``How noisy is the garden right now?'' The agent obtains the correct value but submits a string where the protocol requires a raw numeric answer. All four tool calls execute successfully; the task fails at typed answer comparison.
\begin{center}
\small
\begin{tabular}{@{}>{\raggedright\arraybackslash}p{7em}>{\raggedright\arraybackslash}p{\dimexpr\linewidth-7em-2\tabcolsep\relax}@{}}
\toprule
Stage & Saved evidence \\
\midrule
Turn 3 & \makecell[t{>{\raggedright\arraybackslash}p{\linewidth}}]{\texttt{session\_get\_space\_state(garden)} returns \texttt{noise: 45}.} \\
Turn 4 & \makecell[t{>{\raggedright\arraybackslash}p{\linewidth}}]{\texttt{session\_submit\_answer(value="45")} is accepted as a submission.} \\
Evaluation & \makecell[t{>{\raggedright\arraybackslash}p{\linewidth}}]{Expected \texttt{[45]}, submitted \texttt{["45"]}; answer success is false and action success is 100\%.} \\
\bottomrule
\end{tabular}
\end{center}

\refstepcounter{failurecase}
\para{Case \thefailurecase: repeated response-format errors prevent execution (E2).}
\label{case:protocol}
Qwen3.5-2B, Car, \texttt{task\_1f8e0b0b}.
The task combines powering on seat comfort and parking cameras with a request for an unavailable third exterior mirror. The agent repeatedly places its response in multiple answer blocks. Parser feedback states that the actions were not executed, but the same structural error persists until the turn limit.
\begin{center}
\small
\begin{tabular}{@{}>{\raggedright\arraybackslash}p{7em}>{\raggedright\arraybackslash}p{\dimexpr\linewidth-7em-2\tabcolsep\relax}@{}}
\toprule
Stage & Saved evidence \\
\midrule
Turns 1--20 & \makecell[t{>{\raggedright\arraybackslash}p{\linewidth}}]{Every response records \texttt{extra\_answer\_blocks}; no requested action reaches tool execution.} \\
Termination & \makecell[t{>{\raggedright\arraybackslash}p{\linewidth}}]{\texttt{max\_turns} at turn 20; \texttt{total\_tool\_calls=0}.} \\
Evaluation & \makecell[t{>{\raggedright\arraybackslash}p{\linewidth}}]{Both requested power states remain false, and the required \texttt{DEVICE\_TARGET\_NOT\_FOUND} report is absent.} \\
\bottomrule
\end{tabular}
\end{center}

\refstepcounter{failurecase}
\para{Case \thefailurecase: successful execution on the wrong spatial target (E5, E1).}
\label{case:wrong-space}
Claude Haiku 4.5, Home, \texttt{task\_2b3532a6}.
The request is ``Could you turn Bedroom Washer Dryer's power on?'' The bedroom inventory has no washer dryer; the available \texttt{washer\_dryer\_1} belongs to the laundry room. The agent binds that device to the bedroom request and powers it on. Luna identifies the missed spatial constraint as E5 and the incorrect object binding as E1.
\begin{center}
\small
\begin{tabular}{@{}>{\raggedright\arraybackslash}p{7em}>{\raggedright\arraybackslash}p{\dimexpr\linewidth-7em-2\tabcolsep\relax}@{}}
\toprule
Stage & Saved evidence \\
\midrule
Inventory & \makecell[t{>{\raggedright\arraybackslash}p{\linewidth}}]{The returned device inventory places \texttt{washer\_dryer\_1} in \texttt{laundry\_room}, not \texttt{bedroom}.} \\
Turns 5--6 & \makecell[t{>{\raggedright\arraybackslash}p{\linewidth}}]{\texttt{washer\_dryer\_1.set\_power(true)} succeeds; the next query confirms \texttt{power=true}.} \\
Evaluation & \makecell[t{>{\raggedright\arraybackslash}p{\linewidth}}]{The expected \texttt{DEVICE\_TARGET\_NOT\_FOUND} report is missing; one unexpected side effect is recorded. All 16 tool calls succeed.} \\
\bottomrule
\end{tabular}
\end{center}

\refstepcounter{failurecase}
\para{Case \thefailurecase: intermediate observations become extra answers (E6).}
\label{case:answer-coordination}
GPT-6.1 Sol, Car, \texttt{task\_ef1eb0e7}.
This aggregate task combines device controls with six requested query answers. Some queries first ask the agent to read a value and then use it as a comparison baseline. The agent computes all six answers correctly but also submits two intermediate readings, producing eight answers. The order-preserving alignment matches the six requested values and flags the two extras.
\begin{center}
\small
\begin{tabular}{@{}>{\raggedright\arraybackslash}p{7em}>{\raggedright\arraybackslash}p{\dimexpr\linewidth-7em-2\tabcolsep\relax}@{}}
\toprule
Stage & Saved evidence \\
\midrule
Turn 7 & \makecell[t{>{\raggedright\arraybackslash}p{\linewidth}}]{The trunk's illuminance is 50 and multimedia's \texttt{audio\_level} is 60; both serve as baselines for later comparisons.} \\
Turn 8 & \makecell[t{>{\raggedright\arraybackslash}p{\linewidth}}]{The agent submits 50 and 60 as additional answers, at zero-based submission indices 3 and 5.} \\
Evaluation & \makecell[t{>{\raggedright\arraybackslash}p{\linewidth}}]{All six expected answers match and all control assertions pass, but \texttt{extra\_submitted\_indices=[3,5]} makes answer success false.} \\
\bottomrule
\end{tabular}
\end{center}

\refstepcounter{failurecase}
\para{Case \thefailurecase: final states pass, but the follow-up is too early (E10).}
\label{case:time-window}
GPT-6 Luna, Phone, \texttt{task\_e60cde2b}.
The instruction permits enabling battery power saving only 1--5 minutes after a running timer finishes. The agent waits for timer completion, then enables power saving in the same simulated minute. Its eight-turn interaction meets all final-state assertions but violates the required positive delay.
\begin{center}
\small
\begin{tabular}{@{}>{\raggedright\arraybackslash}p{7em}>{\raggedright\arraybackslash}p{\dimexpr\linewidth-7em-2\tabcolsep\relax}@{}}
\toprule
Stage & Saved evidence \\
\midrule
Turn 4 & \makecell[t{>{\raggedright\arraybackslash}p{\linewidth}}]{Timer: \texttt{status=running}, \texttt{remaining\_minutes=2}, and \texttt{duration\_minutes=3}.} \\
Turns 5--6 & \makecell[t{>{\raggedright\arraybackslash}p{\linewidth}}]{Sleeping for 2 minutes returns \texttt{12:03}, elapsed minute 3; a status query then returns \texttt{completed}.} \\
Turn 7 & \makecell[t{>{\raggedright\arraybackslash}p{\linewidth}}]{\texttt{battery\_1.set\_power\_saving\_mode(true)} executes at minute 3.} \\
Evaluation & \makecell[t{>{\raggedright\arraybackslash}p{\linewidth}}]{Timer completion and follow-up both occur at minute 3. All six final-state assertions pass; the required 1--5 minute workflow relation fails.} \\
\bottomrule
\end{tabular}
\end{center}

\refstepcounter{failurecase}
\para{Case \thefailurecase: an extra wait changes the required process state (E10).}
\label{case:state-timing}
GPT-5.6 Luna, Car, \texttt{task\_c8508f9c}.
The request includes selecting the propulsion air-suspension height. The reference checks the running adjustment state immediately after the requested controls. After completing the controls, the agent adds a one-minute wait, advancing the suspension beyond that target state. This trace shows why a successful command and a matching process state are distinct checks.
\begin{center}
\small
\begin{tabular}{@{}>{\raggedright\arraybackslash}p{7em}>{\raggedright\arraybackslash}p{\dimexpr\linewidth-7em-2\tabcolsep\relax}@{}}
\toprule
Stage & Saved evidence \\
\midrule
Turn 4 & \makecell[t{>{\raggedright\arraybackslash}p{\linewidth}}]{\texttt{air\_suspension\_1.select\_height(height="normal")} executes successfully.} \\
Turn 6 & \makecell[t{>{\raggedright\arraybackslash}p{\linewidth}}]{\texttt{session\_sleep\_time(minutes=1)} advances the simulated clock by one minute.} \\
Evaluation & \makecell[t{>{\raggedright\arraybackslash}p{\linewidth}}]{Expected \texttt{status=running}, \texttt{phase=adjusting}, and \texttt{remaining\_minutes=1}; the final values are \texttt{completed}, \texttt{completed}, and 0. These three assertions fail; the other assertions pass.} \\
\bottomrule
\end{tabular}
\end{center}

\refstepcounter{failurecase}
\para{Case \thefailurecase: a combined control plan triggers a simulation conflict (E8, E7, E11).}
\label{case:runtime-anomaly}
GPT-5.6 Sol, Home, \texttt{task\_433f5c74}.
The task requires bedroom 2 to reach 16--18 degrees strictly before 14:06. The agent treats the air exchanger's outdoor-temperature reading as evidence that opening the window will also cool the room, then combines air conditioning, ventilation, and an open window. All calls succeed during interaction, but the ensuing background simulation rejects the concurrent temperature effects.
\begin{center}
\small
\begin{tabular}{@{}>{\raggedright\arraybackslash}p{7em}>{\raggedright\arraybackslash}p{\dimexpr\linewidth-7em-2\tabcolsep\relax}@{}}
\toprule
Stage & Saved evidence \\
\midrule
Turns 2 and 6 & \makecell[t{>{\raggedright\arraybackslash}p{\linewidth}}]{Bedroom 2 temperature is 20; \texttt{air\_exchanger\_1} reports \texttt{outdoor\_temperature=10.0}.} \\
Turns 6--7 & \makecell[t{>{\raggedright\arraybackslash}p{\linewidth}}]{The agent sets cooling to 16 with high fan speed, enables ventilation, and opens \texttt{window\_opener\_1}.} \\
Post-simulation & \makecell[t{>{\raggedright\arraybackslash}p{\linewidth}}]{\texttt{effect\_conflict} on bedroom 2 temperature: \texttt{['approach', 'approach', 'delta']}. The \texttt{entered\_range} transition is absent.} \\
\bottomrule
\end{tabular}
\end{center}
The ordered labels distinguish the unsupported state inference (E8), the resulting control plan (E7), and the simulator's failure to compose the concurrent effects (E11).

\refstepcounter{failurecase}
\para{Case \thefailurecase: the public comparison and private reference disagree (E12, E3).}
\label{case:reference-anomaly}
Claude Haiku 4.5, Car, \texttt{task\_8de76e0d}.
The first public question asks which of cabin and multimedia has the lower volume. The agent observes 20 and 60 and selects cabin. The private reference instead filters for values strictly below the cabin's volume, yielding an empty list. Sol retains E12 for this mismatch and adds E3 because the agent submits a string where target selection requires an ID list.
\begin{center}
\small
\begin{tabular}{@{}>{\raggedright\arraybackslash}p{7em}>{\raggedright\arraybackslash}p{\dimexpr\linewidth-7em-2\tabcolsep\relax}@{}}
\toprule
Stage & Saved evidence \\
\midrule
Turn 2 & \makecell[t{>{\raggedright\arraybackslash}p{\linewidth}}]{\texttt{audio\_level} is 20 in cabin and 60 in multimedia.} \\
Turn 3 & \makecell[t{>{\raggedright\arraybackslash}p{\linewidth}}]{\texttt{session\_submit\_answer(value="cabin")} submits a string; the selection protocol calls for \texttt{["cabin"]}.} \\
Evaluation & \makecell[t{>{\raggedright\arraybackslash}p{\linewidth}}]{The first expected answer is \texttt{[]}. The other four answers match and all control assertions pass. Sol's reviewed labels are E12 followed by E3.} \\
\bottomrule
\end{tabular}
\end{center}

\refstepcounter{failurecase}
\para{Case \thefailurecase: a correct enumeration fails because the baseline is also submitted (E6).}
\label{case:astra-enumeration}
GPT-6 Astra, Car, \texttt{task\_4aba80f9} (Q6).
The instruction asks, ``Check the current noise in the trunk, then tell me which of the cabin, exterior, and trunk have a different noise level from it.'' The agent reads all three values and identifies the correct targets, but treats the baseline lookup as a separate query requiring an answer. It submits both the trunk reading and the target list, whereas the task's single-query reference requires only the list.
\begin{center}
\small
\begin{tabular}{@{}>{\raggedright\arraybackslash}p{7em}>{\raggedright\arraybackslash}p{\dimexpr\linewidth-7em-2\tabcolsep\relax}@{}}
\toprule
Stage & Saved evidence \\
\midrule
Turn 1 & \makecell[t{>{\raggedright\arraybackslash}p{\linewidth}}]{The agent plans to ``answer both queries in order,'' treating the baseline lookup as an additional output.} \\
Turn 2 & \makecell[t{>{\raggedright\arraybackslash}p{\linewidth}}]{Readings: \texttt{noise=40} for \texttt{trunk}, 30 for \texttt{cabin}, and 120 for \texttt{exterior}.} \\
Turn 3 & \makecell[t{>{\raggedright\arraybackslash}p{\linewidth}}]{The agent submits \texttt{40}, then \texttt{["cabin","exterior"]}, and finishes.} \\
Evaluation & \makecell[t{>{\raggedright\arraybackslash}p{\linewidth}}]{The target list matches exactly, but \texttt{extra\_submitted\_indices=[0]} makes answer success false. All six recorded tool calls succeed.} \\
\bottomrule
\end{tabular}
\end{center}
Across Astra's 46 Q6 tasks, 44 first attempts fail; 43 of these contain the exact target list plus an extra baseline answer. The current attribution assigns E6 to 42 of those cases and E3 to one. This pattern localizes the dominant failure to distinguishing intermediate observations from required answer submissions. The low Q6 task-success rate therefore measures sensitivity to task decomposition and the answer contract, while these traces demonstrate correct target selection.

\clearpage
\section{Agent Cost Analysis}
\label[appendix]{app:agent-cost}

\Cref{fig:api-cost-performance} compares mean Pass@1 with inference cost per task attempt. \Cref{tab:api-prices} gives the fixed October 28, 2026 prices, and \Cref{tab:agent-costs} summarizes token usage and costs across the full test set. Raw cost prices all input tokens at the ordinary input rate; cached cost applies cache-read and cache-write rates. Both count the first request of each agent turn and exclude transport retries and judge calls.

\begin{table}[H]
\centering

\begin{threeparttable}

\scriptsize
\begin{tabular}{lccccc}
\toprule
\textbf{Model} & \textbf{Input} & \textbf{Cache read} & \textbf{Output} & \textbf{Cache write}& \textbf{Official pricing} \\
\midrule
Claude Haiku 4.5 & 1 & 0.1 & 5 & 1.25 & \href{https://platform.claude.com/docs/en/about-claude/pricing}{Pricing} \\
DeepSeek V4.1 Flash & 0.3 & 0.006 & 1.2 & -- & \href{https://api-docs.deepseek.com/quick_start/pricing/}{Pricing} \\
Gemini 2.5 Flash-Lite & 0.1 & 0.01 & 0.4 & -- & \href{https://ai.google.dev/gemini-api/docs/pricing#gemini-2.5-flash-lite}{Pricing} \\
Gemini 3.8 Flash & 0.75 & 0.075 & 3.75 & -- & \href{https://ai.google.dev/gemini-api/docs/pricing#gemini-3.8-flash}{Pricing} \\
GLM-5.3 Flash$^\dagger$ & 0.120 & 0.034 & 0.419 & -- & \href{https://docs.bigmodel.cn/cn/guide/start/pricing}{Pricing} \\
GPT-5.6 Luna & 0.2 & 0.02 & 1.2 & 0.25 & \href{https://developers.openai.com/api/docs/pricing}{Pricing} \\
GPT-5.6 Terra & 2 & 0.2 & 12 & 2.5 & \href{https://developers.openai.com/api/docs/pricing}{Pricing} \\
GPT-5.6 Sol & 4 & 0.4 & 20 & 5 & \href{https://developers.openai.com/api/docs/pricing}{Pricing} \\
GPT-6 Luna & 0.1 & 0.01 & 0.5 & 0.125 & \href{https://developers.openai.com/api/docs/pricing}{Pricing} \\
GPT-6 Sol & 2 & 0.2 & 10 & 2.5 & \href{https://developers.openai.com/api/docs/pricing}{Pricing} \\
GPT-6.1 Terra & 1 & 0.05 & 5 & 1.25 & \href{https://developers.openai.com/api/docs/pricing}{Pricing} \\
GPT-6.1 Sol & 2 & 0.1 & 10 & 2.5 & \href{https://developers.openai.com/api/docs/pricing}{Pricing} \\
GPT-6 Astra & 10 & 1 & 50 & 12.5 & \href{https://developers.openai.com/api/docs/models/gpt-6-astra}{Pricing} \\
\bottomrule
\end{tabular}
\begin{tablenotes}[flushleft]
\item[] Standard API prices collected from official websites on October 28, 2026. We report prices for the short-context tier, as nearly all API calls fall within the corresponding input-length limit.
\item[$\dagger$] We use the currency of 1 USD = 6.68 CNY.
\end{tablenotes}
\end{threeparttable}
\caption{API prices (USD per 1M tokens) used for cost estimates.}
\label{tab:api-prices}
\end{table}

\begin{table}[H]
\centering

\begin{threeparttable}

\scriptsize
\setlength{\tabcolsep}{4pt}
\begin{tabular}{lccccccrr}
\toprule
\textbf{Model} & {\boldmath$S$} & \textbf{Input} & \textbf{Output} & \textbf{Total} & \makecell[c]{\textbf{Cache}\\\textbf{read}} & \makecell[c]{\textbf{Cache}\\\textbf{write}$^\dagger$} & \makecell[c]{\textbf{Raw total}\\\textbf{(per attempt)}} & \makecell[c]{\textbf{Cached total}\\\textbf{(per attempt)}} \\
\midrule
Claude Haiku 4.5 & 1 & 304.930 & 3.403 & 308.333 & 97.330 & 20.006 & 321.95 (0.1258) & 239.35 (0.0935) \\
DeepSeek V4.1 Flash & 1 & 215.737 & 6.615 & 222.352 & 161.952 & 0.000 & 72.66 (0.0284) & 25.05 (0.0098) \\
Gemini 2.5 Flash-Lite & 1 & 297.477 & 6.627 & 304.104 & 185.163 & 0.000 & 32.40 (0.0127) & 15.73 (0.0061) \\
Gemini 3.8 Flash & 1 & 257.052 & 9.218 & 266.271 & 83.669 & 0.000 & 227.36 (0.0888) & 170.88 (0.0668) \\
GLM-5.3 Flash & 1 & 180.235 & 6.196 & 186.430 & 104.169 & 0.000 & 24.18 (0.0094) & 15.29 (0.0060) \\
GPT-5.6 Luna & 3 & 467.338 & 13.789 & 481.127 & 373.580 & 93.758 & 110.01 (0.0143) & 47.46 (0.0062) \\
GPT-5.6 Terra & 3 & 486.210 & 9.878 & 496.088 & 376.757 & 109.453 & 1,090.96 (0.1421) & 467.52 (0.0609) \\
GPT-5.6 Sol & 3 & 483.845 & 10.314 & 494.159 & 383.116 & 100.729 & 2,141.65 (0.2789) & 863.16 (0.1124) \\
GPT-6 Luna & 3 & 411.031 & 6.977 & 418.007 & 318.379 & 92.651 & 44.59 (0.0058) & 18.25 (0.0024) \\
GPT-6 Sol & 3 & 499.327 & 7.349 & 506.676 & 415.988 & 83.338 & 1,072.14 (0.1396) & 365.03 (0.0475) \\
GPT-6.1 Terra & 3 & 344.092 & 4.982 & 349.074 & 154.248 & 189.844 & 369.00 (0.0480) & 269.93 (0.0351) \\
GPT-6.1 Sol & 3 & 491.725 & 5.497 & 497.222 & 390.989 & 100.736 & 1,038.42 (0.1352) & 345.91 (0.0450) \\
GPT-6 Astra & 3 & 455.528 & 5.023 & 460.551 & 379.829 & 75.699 & 4,806.41 (0.6258) & 1,577.19 (0.2054) \\
\midrule
Qwen3.5-0.8B & 3 & 1051.367 & 30.143 & 1081.510 & 0.000 & 0.000 & -- & -- \\
Qwen3.5-2B & 3 & 881.936 & 29.232 & 911.168 & 0.000 & 0.000 & -- & -- \\
Qwen3.5-4B & 3 & 940.926 & 12.530 & 953.456 & 0.000 & 0.000 & -- & -- \\
Qwen3.5-9B & 3 & 982.853 & 15.358 & 998.211 & 0.000 & 0.000 & -- & -- \\
Qwen3.5-27B & 3 & 862.170 & 19.137 & 881.307 & 0.000 & 0.000 & -- & -- \\
MiniCPM5-1B & 3 & 1182.593 & 57.070 & 1239.663 & 0.000 & 0.000 & -- & -- \\
MiniCPM5-2B & 3 & 940.823 & 17.041 & 957.863 & 0.000 & 0.000 & -- & -- \\
Gemma-4-E2B-it & 3 & 751.790 & 19.708 & 771.498 & 0.000 & 0.000 & -- & -- \\
Gemma-4-E4B-it & 3 & 790.875 & 14.996 & 805.871 & 0.000 & 0.000 & -- & -- \\
\bottomrule
\end{tabular}
\begin{tablenotes}[flushleft]
\item[] Tokens are in millions; costs are in USD. $S$ is the number of samples per task.
\item[$\dagger$] For GPT models, we estimate cache-write tokens as total input minus recorded cache-read tokens.
\end{tablenotes}
\end{threeparttable}
\caption{Agent token usage and cost across the test set.}
\label{tab:agent-costs}
\end{table}

\clearpage
\section{Prompts}
\label[appendix]{app:prompts}

The following templates specify the prompts used for construction, instruction rewriting, agent evaluation, and analysis. Braced fields denote runtime inputs. Shared text is printed once and inserted at the named fields.

\begin{CJK*}{UTF8}{gbsn}

\subsection{World Distillation}

\subsubsection{Shared Project and Model Context}

The project context prefixes construction and judge requests. Model-generation system prompts also insert the shared model rules below in the order shown by their named fields. ReAct uses its own system prompt.

\inputprompt{themeAcolor}{GENERAL_SYSTEM_PROMPT}{prompts/pipeline/general_system_prompt.txt}

\inputprompt{themeAcolor}{MODEL_GENERAL_SYSTEM_PROMPT}{prompts/pipeline/model_general_system_prompt.txt}

\inputprompt{themeAcolor}{HOME_ENVIRONMENT_EFFECT_RULES}{prompts/pipeline/home_environment_effect_rules.txt}

\inputprompt{themeAcolor}{PROPERTY_RUNTIME_CONTRACT}{prompts/pipeline/property_runtime_contract.txt}

\inputprompt{themeAcolor}{GENERATED_CODE_SECURITY_RULES}{prompts/pipeline/generated_code_security_rules.txt}

\inputprompt{themeAcolor}{SPECIFICATION_GUIDELINES}{prompts/pipeline/specification_guidelines.txt}

\inputprompt{themeAcolor}{CAPABILITY_RULES}{prompts/pipeline/capability_rules.txt}

\subsubsection{Device List Creation}

\inputprompt{themeAcolor}{OBJECT_LIST_CREATOR_SYSTEM_PROMPT}{prompts/pipeline/object_list_creator_system_prompt.txt}

\inputprompt{themeBcolor}{OBJECT_LIST_CREATOR_USER_PROMPT}{prompts/pipeline/object_list_creator_user_prompt.txt}

\subsubsection{Device Placement}

\inputprompt{themeAcolor}{PLACEMENT_SYSTEM_PROMPT}{prompts/pipeline/placement_system_prompt.txt}

\inputprompt{themeBcolor}{PLACEMENT_USER_PROMPT}{prompts/pipeline/placement_user_prompt.txt}

\subsubsection{Device Specification}

\inputprompt{themeAcolor}{BASE_SYSTEM_PROMPT}{prompts/pipeline/base_system_prompt.txt}

\inputprompt{themeBcolor}{BASE_USER_PROMPT}{prompts/pipeline/base_user_prompt.txt}

\inputprompt{themeBcolor}{INHERIT_USER_PROMPT}{prompts/pipeline/inherit_user_prompt.txt}

The reference-knowledge field combines the schema, domain property catalog, and retrieved specification references. Inherited specifications are supplied when applicable.

\inputprompt{themeBcolor}{META_SCHEMA_REFERENCE_PROMPT}{prompts/pipeline/meta_schema_reference_prompt.txt}

\inputprompt{themeBcolor}{META_PROPERTY_CATALOG_REFERENCE_PROMPT}{prompts/pipeline/meta_property_catalog_reference_prompt.txt}

\inputprompt{themeBcolor}{META_PROPERTY_CATALOG_CONTENT_PROMPT}{prompts/pipeline/meta_property_catalog_content_prompt.txt}

\inputprompt{themeBcolor}{META_RETRIEVED_EXAMPLE_PROMPT}{prompts/pipeline/meta_retrieved_example_prompt.txt}

\inputprompt{themeBcolor}{META_BASE_REFERENCE_PROMPT}{prompts/pipeline/meta_base_reference_prompt.txt}

\subsubsection{Specification Review and Repair}

\inputprompt{themeAcolor}{META_REVIEW_SYSTEM_PROMPT}{prompts/pipeline/meta_review_system_prompt.txt}

\inputprompt{themeBcolor}{META_REVIEW_USER_PROMPT}{prompts/pipeline/meta_review_user_prompt.txt}

\inputprompt{themeBcolor}{META_REVIEW_DOMAIN_CONTEXT_PROMPT}{prompts/pipeline/meta_review_domain_context_prompt.txt}

\inputprompt{themeBcolor}{META_REPAIR_USER_PROMPT}{prompts/pipeline/meta_repair_user_prompt.txt}

\subsubsection{Behavior Implementation}

\inputprompt{themeAcolor}{CODE_COMPLETION_SYSTEM_PROMPT}{prompts/pipeline/code_completion_system_prompt.txt}

\inputprompt{themeBcolor}{CODE_COMPLETION_BASE_USER_PROMPT}{prompts/pipeline/code_completion_base_user_prompt.txt}

\inputprompt{themeBcolor}{CODE_COMPLETION_INHERIT_USER_PROMPT}{prompts/pipeline/code_completion_inherit_user_prompt.txt}

\inputprompt{themeBcolor}{CODE_RETRIEVED_EXAMPLE_PROMPT}{prompts/pipeline/code_retrieved_example_prompt.txt}

Specification-only implementation was used for the Connectivity component; generated method bodies were assembled into the deterministic class template.

\inputprompt{themeAcolor}{CODE_SPECIFICATION_METHODS_SYSTEM_PROMPT}{prompts/pipeline/code_specification_methods_system_prompt.txt}

\inputprompt{themeBcolor}{CODE_SPECIFICATION_METHODS_USER_PROMPT}{prompts/pipeline/code_specification_methods_user_prompt.txt}

\subsubsection{Code Repair}

\inputprompt{themeAcolor}{CODE_REPAIR_SYSTEM_PROMPT}{prompts/pipeline/code_repair_system_prompt.txt}

\inputprompt{themeBcolor}{CODE_FIX_BASE_USER_PROMPT}{prompts/pipeline/code_fix_base_user_prompt.txt}

\inputprompt{themeBcolor}{CODE_FIX_INHERIT_USER_PROMPT}{prompts/pipeline/code_fix_inherit_user_prompt.txt}

\subsubsection{Test Generation and Repair}

\inputprompt{themeAcolor}{OBJECT_CLASS_VERIFIER_SYSTEM_PROMPT}{prompts/pipeline/object_class_verifier_system_prompt.txt}

\inputprompt{themeBcolor}{OBJECT_CLASS_VERIFIER_USER_PROMPT}{prompts/pipeline/object_class_verifier_user_prompt.txt}

\inputprompt{themeBcolor}{VERIFIER_PARENT_IMPLEMENTATIONS_PROMPT}{prompts/pipeline/verifier_parent_implementations_prompt.txt}

\inputprompt{themeBcolor}{VERIFIER_REPAIR_CONTEXT_PROMPT}{prompts/pipeline/verifier_repair_context_prompt.txt}

\inputprompt{themeBcolor}{VERIFIER_SPECIFICATION_ONLY_TEXT}{prompts/pipeline/verifier_specification_only_text.txt}

\subsubsection{Verification Analysis}

\inputprompt{themeAcolor}{OBJECT_CLASS_VERIFIER_ANALYZER_SYSTEM_PROMPT}{prompts/pipeline/object_class_verifier_analyzer_system_prompt.txt}

\inputprompt{themeBcolor}{VERIFIER_ANALYZER_SPECIFICATION_PROMPT}{prompts/pipeline/verifier_analyzer_specification_prompt.txt}

\inputprompt{themeBcolor}{OBJECT_CLASS_VERIFIER_ANALYZER_USER_PROMPT}{prompts/pipeline/object_class_verifier_analyzer_user_prompt.txt}

\subsection{Instruction Rewriting}

The shared system template receives one of the seven style definitions below. Template instructions are generated deterministically. The user message contains the task fields and applicable context fields shown here.

\inputprompt{themeAcolor}{TASK_REWRITE_SYSTEM_PROMPT}{prompts/pipeline/task_rewrite_system_prompt.txt}

\inputprompt{themeBcolor}{TASK_REWRITE_USER_PROMPT}{prompts/pipeline/task_rewrite_user_prompt.txt}

\inputprompt{themeBcolor}{TASK_REWRITE_CONTEXT}{prompts/pipeline/task_rewrite_context.txt}

Failed-request and dependency fields appear for the corresponding tasks. Direct and infeasible requests use the protected-literal field; other explicit styles use the required-values field. Validation feedback is included for retries.

\subsubsection{Direct Style}

\inputprompt{themeBcolor}{REWRITE_STYLE_DIRECT}{prompts/pipeline/rewrite_style_direct.txt}

\subsubsection{Concise Style}

\inputprompt{themeBcolor}{REWRITE_STYLE_CONCISE}{prompts/pipeline/rewrite_style_concise.txt}

\subsubsection{Polite Style}

\inputprompt{themeBcolor}{REWRITE_STYLE_POLITE}{prompts/pipeline/rewrite_style_polite.txt}

\subsubsection{Conversational Style}

\inputprompt{themeBcolor}{REWRITE_STYLE_CONVERSATIONAL}{prompts/pipeline/rewrite_style_conversational.txt}

\subsubsection{Distracting Style}

\inputprompt{themeBcolor}{REWRITE_STYLE_DISTRACTING}{prompts/pipeline/rewrite_style_distracting.txt}

\subsubsection{Folk Style}

\inputprompt{themeBcolor}{REWRITE_STYLE_FOLK}{prompts/pipeline/rewrite_style_folk.txt}

\subsubsection{Implicit Style}

\inputprompt{themeBcolor}{REWRITE_STYLE_IMPLICIT}{prompts/pipeline/rewrite_style_implicit.txt}

\subsection{ReAct Agent}

\inputprompt{themeAcolor}{AGENT_SYSTEM_PROMPT}{prompts/pipeline/agent_system_prompt.txt}

\inputprompt{themeBcolor}{AGENT_USER_PROMPT}{prompts/pipeline/agent_user_prompt.txt}

\inputprompt{themeBcolor}{AGENT_OBSERVATION_PROMPT}{prompts/pipeline/agent_observation_prompt.txt}

\subsubsection{Parser Feedback}

The observation message reports parser status and actual tool outputs. The following feedback templates describe accepted or rejected response structure.

\inputprompt{themeBcolor}{AGENT_PARSER_FEEDBACK}{prompts/pipeline/agent_parser_feedback.txt}

\subsection{LLM Judges}

\subsubsection{Failure Report Evaluation}

\inputprompt{themeAcolor}{REPORT_JUDGE_SYSTEM_PROMPT}{prompts/pipeline/report_judge_system_prompt.txt}

\inputprompt{themeBcolor}{REPORT_JUDGE_USER_PROMPT}{prompts/pipeline/report_judge_user_prompt.txt}

\subsubsection{Memory Evaluation}

\inputprompt{themeAcolor}{MEMORY_JUDGE_SYSTEM_PROMPT}{prompts/pipeline/memory_judge_system_prompt.txt}

\inputprompt{themeBcolor}{MEMORY_JUDGE_USER_PROMPT}{prompts/pipeline/memory_judge_user_prompt.txt}

\subsubsection{Failure Attribution}

\inputprompt{themeAcolor}{FAILURE_JUDGE_SYSTEM_PROMPT}{prompts/pipeline/failure_judge_system_prompt.txt}

\inputprompt{themeBcolor}{FAILURE_JUDGE_USER_PROMPT}{prompts/pipeline/failure_judge_user_prompt.txt}

\subsection{Construction Comparison}

\subsubsection{Codex Device Construction}

Codex receives the same task request at each knowledge level. The corresponding external knowledge is supplied in DEVICE.md, with the implementation framework and expert references available in the checkout.

\inputprompt{themeBcolor}{CODEX_CONSTRUCTION_REQUEST}{prompts/pipeline/codex_construction_request.txt}

\subsubsection{Generated-Task Coverage}

\inputprompt{themeAcolor}{COVERAGE_JUDGE_SYSTEM_PROMPT}{prompts/pipeline/coverage_judge_system_prompt.txt}

\inputprompt{themeBcolor}{COVERAGE_JUDGE_USER_PROMPT}{prompts/pipeline/coverage_judge_user_prompt.txt}

\end{CJK*}

\end{document}